# DECISIVE Test Methods Handbook

## Test Methods for Evaluating sUAS in Subterranean and Constrained Indoor Environments, Version 1.1

Adam Norton, Reza Ahmadzadeh, Kshitij Jerath, Paul Robinette,
Jay Weitzen, Thanuka Wickramarathne, Holly Yanco, Minseop Choi,
Ryan Donald, Brendan Donoghue, Christian Dumas, Peter Gavriel,
Alden Giedraitis, Brendan Hertel, Jack Houle, Nathan Letteri,
Edwin Meriaux, Zahra Rezaei Khavas, Rakshith Singh, Gregg Willcox, Naye Yoni

University of Massachusetts Lowell
U.S. Army Combat Capabilities Development Command Soldier Center (DEVCOM-SC)

**Abstract**: This handbook outlines all test methods developed under the Development and Execution of Comprehensive and Integrated Subterranean Intelligent Vehicle Evaluations (DECISIVE) project by the University of Massachusetts Lowell for evaluating small unmanned aerial systems (sUAS) performance in subterranean and constrained indoor environments, spanning communications, field readiness, interface, obstacle avoidance, navigation, mapping, autonomy, trust, and situation awareness.

# Table of Contents

# Revision History

Changes in each release of the document are captured here:

| Version | Release Date | Updates |
|---|---|---|
| 1.1 | October 2022 | Various updates made throughout due to recent test method developments across all categories |
| 1.0 | April 2022 | Document released |

# Scope

The test methods specified in this handbook are scoped for evaluating sUAS intended for deployment in subterranean and constrained indoor environments, which puts forth two assumptions about all sUAS to be evaluated using these test methods: (1) able to operate without access to GPS signal, and (2) width from prop top to prop tip does not exceed 91 cm (36 in) wide (i.e., can physically fit through a typical doorway, although successful navigation through is not guaranteed). These two capabilities sufficiently narrow the size of sUAS of which these test methods are applicable to.

This handbook consists of nine categories of test methods that each represent different sets of sUAS capabilities. Each category section contains at least one test method specification, which follow a common format:

- Purpose: A brief description of the objectives of the test method.
- Summary of Test Method: A review of all pertinent components of the test method, conditions that can be varied, and a definition of each test contained in the test method (if multiple test options exist).
- Apparatus and Artifacts: A description of any dimensional and material requirements of the environments where the tests are performed and any elements that need to be fabricated in order to run the test (e.g., wall panels, obstacles).
- Equipment: Tools and electronic devices used to support data collection (e.g., timers, sensors).
- Metrics: A definition of each metric evaluated in the test method.
- Procedure: Steps for the test administrator and operator to follow in order to conduct the test and analyze the metrics (if additional analysis description is needed beyond what is in the metrics section).
- Example Data: A table of example data from one or more sUAS to serve as a template for others to report data using the test methods. Note: all example data presented is anonymized.

All test methods are designed to be run in real-world environments (e.g., MOUT sites) or using fabricated apparatuses (e.g., test bays built from wood, or contained inside of one or more shipping containers). All depictions of test method apparatuses contained throughout are examples.

Throughout the document, the term “sUAS” and “drone” are used interchangeably.

Two types of test artifacts are first specified on the next page, followed by the test method specifications. These test artifacts, particularly the visual acuity targets, are utilized in multiple test methods, so their details are not repeated throughout.

The test methods specified in this handbook were used to conduct benchmarking of 8 sUAS platforms: Cleo Robotics Dronut X1P (P = prototype), FLIR Black Hornet PRS, Flyability Elios 2 GOV, Lumenier Nighthawk V3, Parrot ANAFI USA GOV, Skydio X2D, Teal Golden Eagle, and Vantage Robotics Vesper. The resulting performance data and analysis can be found in the DECISIVE Benchmarking Data Report [Norton et al., 2023].

# Test Artifacts

Several test artifacts are utilized in the test methods reviewed in this handbook. Each leverages designs from the National Institute of Standards and Technology (NIST) response robots program, elements of which are standardized through the ASTM E54.09 Committee on Homeland Security Applications; Subcommittee on Response Robots.

## Visual Acuity Targets

Visual acuity targets come in three types: nested Landolt Cs, hazmat labels, and real world object photos (see Figure 1). The targets are used to evaluate the level of visual acuity able to be achieved by a sUAS based on its camera and OCU display characteristics. Each size of Landolt C corresponds to a different detail level able to be resolved; from the largest to the smallest C: 20, 8, 3, 1.3, 0.5 mm (0.8, 0.3, 0.125, 0.05, and 0.02 in). There are also colored rings around each target that can be used for color acuity identification. Visual acuity targets are printed on 20 cm (8 in) circle stickers.

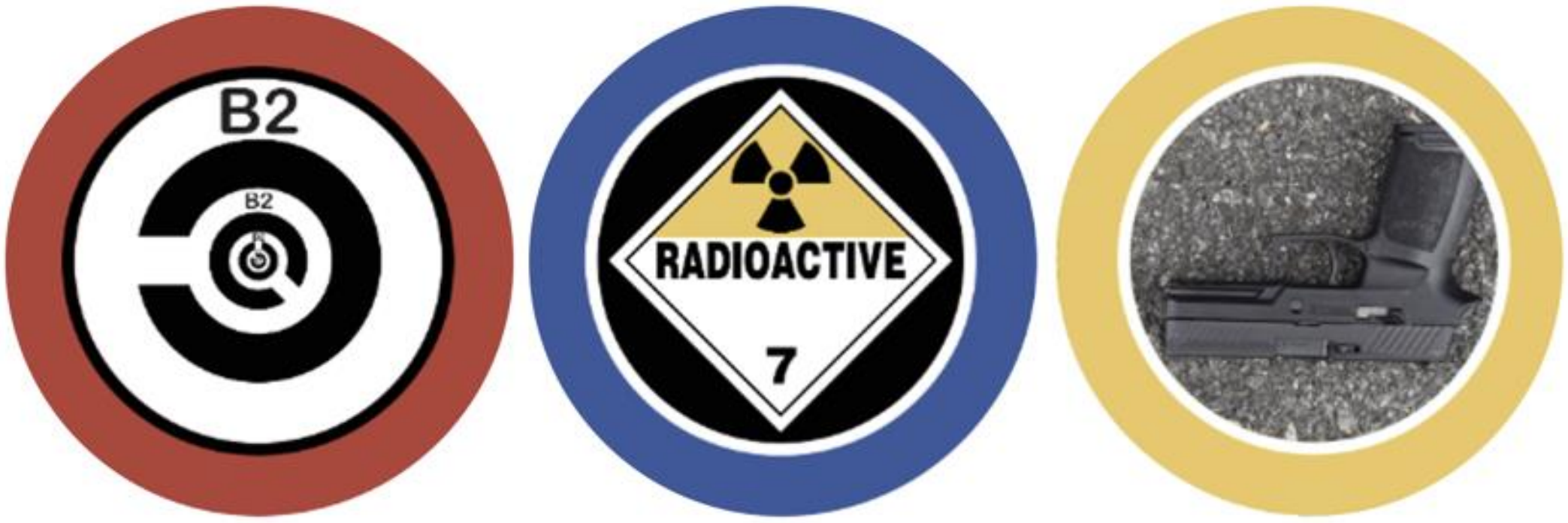

*Figure 1. Examples of NIST visual acuity targets. Left to right: nested Landolt Cs, hazmat labels, real world object photos.*

## Split-cylinder Fiducial

The split-cylinder fiducials are used for evaluating mapping capabilities. The fiducials are fabricated out of 60 cm (24 in) diameter cardboard tube forms, cut to 120 cm (48 in) length and split in half. Five visual acuity targets are attached to the outside of each half-cylinder in a 5-dice pattern. A wooden structure is used to hold the shape of each half cylinder and so it can be mounted in the environment at various heights. See Figure 2. Each pair of fiducials is assigned a level of mapping difficulty based on how they are positioned in an environment. The difficulty rating consists of two metrics:

- <u>Minimum traversal distance</u>: the shortest distance required to traverse from one side of the fiducial to the other, reported in meters.
- <u>Minimum orientation changes</u>: the lowest number of approximately 90 degree turns required when traversing the path defined by the minimum traversal distance.

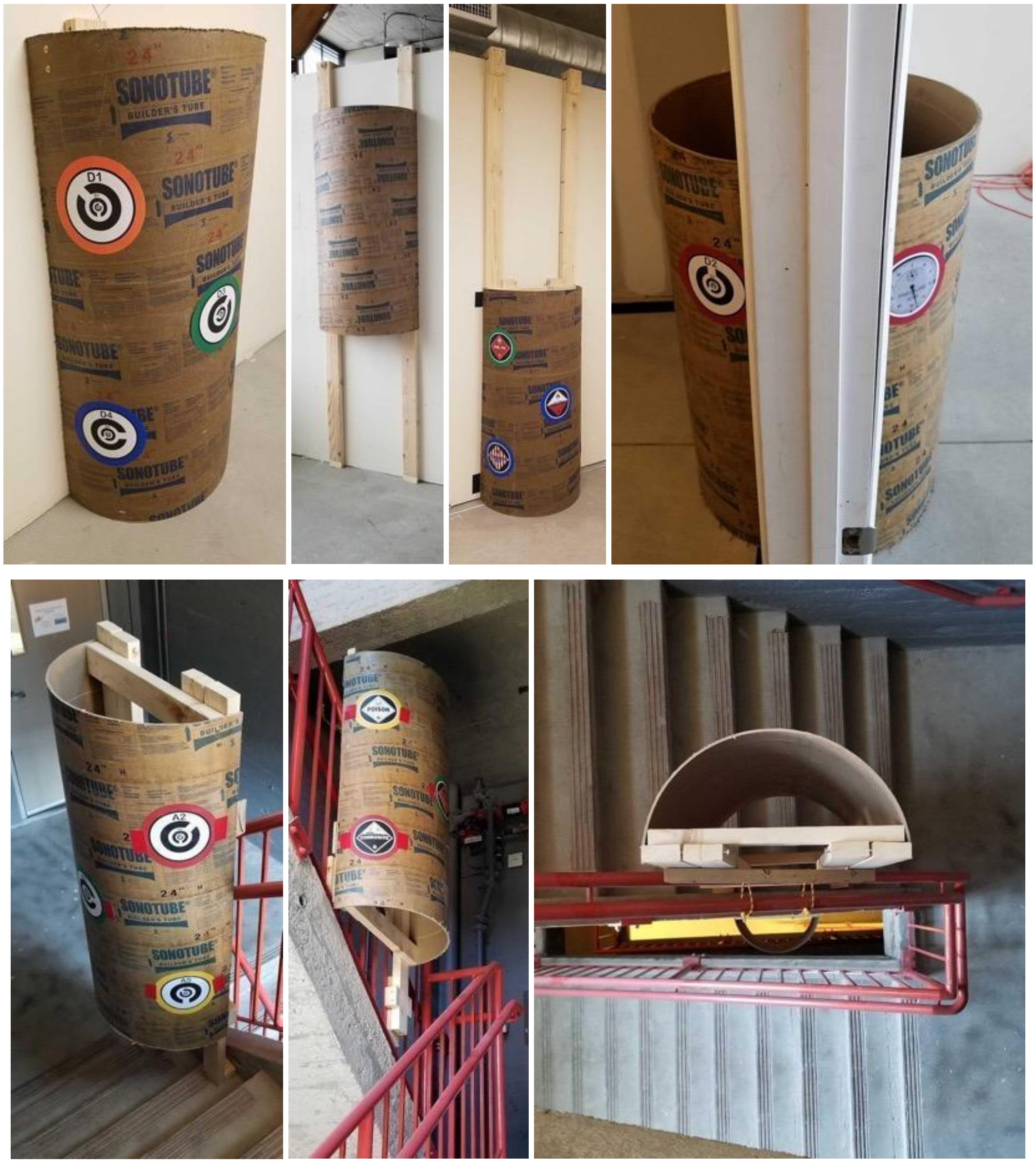

*Figure 2. Split-cylinder fiducials. <u>Top</u>: Horizontally split fiducials on either side of a wall. <u>Bottom</u>: Horizontally and vertically split fiducials on either side of a railing across different floors.*

# Communications

These test methods investigate the quality of the communication links of sUAS operating indoors and in subterranean environments. Indoors operations involve non-line-of-sight (NLOS) signal propagation where there is no direct path between the transmitter and receiver antennas and the operator is solely dependent on the video link for operation. Any obstructions like walls or floors can impact the quality of the communication channel. The test methods consist of measuring the NLOS radio range between the sUAS and the ground control station (GCS) or operator control unit (OCU), observing the sUAS behavior with partial or total communication failure, any induced issues with video latency when in NLOS conditions, and in the presence of induced radio interference.

## Non-Line-of-Sight (NLOS) Communications

### Purpose

This test evaluates the NLOS range of communication indoors by attempting to transmit data through walls and floors. The ability to transmit video and control signals, command the sUAS to perform basic tasks, and any signal indications provided on the OCU are evaluated.

### Summary of Test Method

This test method consists of connecting the sUAS and the OCU at an initial position and then moving the sUAS to other positions that are in NLOS with the initial point due to a wall or floor obstruction. The NLOS communication range for each position is recorded, measured as a straight path through one or more floors and walls between the sUAS and the OCU. The position at which communication fails is indicated by a lack of ability to transmit video, control signal, or command the sUAS to perform tasks. This measure provides an approximate scenario at which the sUAS would be expected to lose communications signal in a real-world deployment.

For each OCU position in the test, the sUAS starts on the ground while the operator attempts to make initial connection to confirm video and control signals (i.e., static connection test). Once confirmed, the operator attempts the following tasks: takeoff, hovering in place, yawing, pitching forward and back, rolling left and right, ascending and descending, camera movement, and landing.

This test method can be run concurrently with the NLOS Video Latency test method.

### Apparatus and Artifacts

A real-world indoor environment deemed relevant for the use case that contains multiple rooms, floors, and/or passageways that are separated by solid wall and/or floor boundaries (e.g., office spaces, multi-level buildings, underground tunnel systems). Minimal cellular coverage and WiFi signals are desirable to limit potential communications interference. Positions in the environment should be identified that progressively add new obstructions for horizontal NLOS (e.g., walls that separate rooms and hallways) and vertical NLOS (e.g., floors of a building, starting from the bottom floor and moving upwards). A single position (X) is used for the sUAS location and should provided sufficient space so as to not impose restrictions on sUAS operations (i.e., takeoff, movement, landing) and the other positions (1, 2, 3, etc.) should be accessible by the operator with the OCU.

Example environments are shown in Figure 1. The horizontal NLOS environments consist of a multi-room office space (the walls are made of drywall and wood) and a series of underground tunnels with connected passageways (the walls are made of concrete, with building infrastructure and earth outside of the walls). The vertical NLOS environment is evaluated across multiple floors of an office building (the ceilings/floors are made out of concrete). The straight line distance through obstructions between the sUAS position (X) and the OCU positions (1, 2, 3, etc.) is shown and the number of obstructions counted.

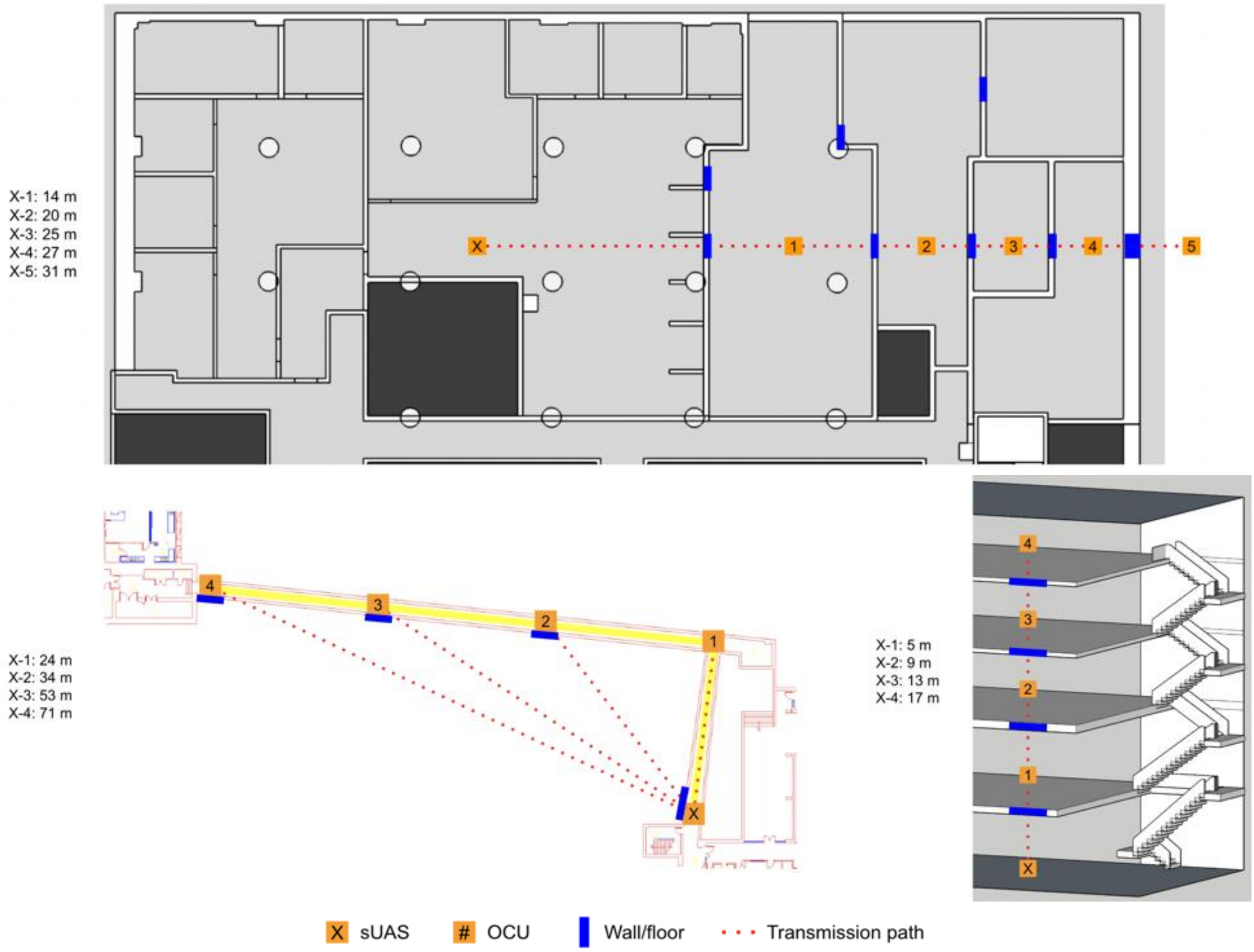

*Figure 1. Example layouts for horizontal NLOS (top and bottom left) and vertical NLOS (bottom right) positions in a real-world environment.*

## Equipment

No additional equipment is required to conduct this test.

## Metrics

- <u>Connection quality</u>: Ability of the operator to send control commands through the OCU to the sUAS and stream video of the sUAS camera to the OCU. Qualitatively evaluated as either good (✓), bad (//), or none (X). Indicators of "bad" video link quality include screen tearing, pixelation, and other artifacts not present when video link quality is "good."
- <u>Video link quality</u>: Visual appearance of the video stream of the sUAS camera on the OCU. Qualitatively evaluated as either good (✓), bad (//), or none (X).
- <u>OCU signal indication</u>: Any signal indication provided on the OCU such as a bar chart or numerical readout for the control and/or video signal level.
- <u>Fly performance</u>: Ability of the operator to control the sUAS to takeoff, hover, yaw, pitch forward and backward, roll left and right, ascend and descend, move the camera, and land, while the OCU is at each position. Qualitatively evaluated as either possible (✓) or not possible (X).
- <u>Position distance</u>: Distance between the sUAS position and the OCU position
- <u>Position obstructions</u>: Number of wall and/or floor boundaries between the sUAS position and the OCU position and the type of material they are made out of.
- <u>Maximum NLOS performance</u>: The maximum position distance and number of position obstructions that the sUAS is able to maintain communications link either horizontally (i.e., through walls) or vertically (i.e., through floors) for both successful static performance (i.e., video and control while grounded) and flying performance (i.e., takeoff, hover, yaw, etc.).

## Procedure

1. Mark the desired NLOS testing positions throughout the environment and record each position's distance and obstructions.
2. Connect the sUAS and the GCS/OCU at the initial position.
3. Leave sUAS stationary on the ground at the initial position.
4. Attempt to connect to view the sUAS video feed and move the sUAS camera.
5. Record control link quality, video link quality, and any available OCU signal indications.
6. If video and control link quality is viable, attempt the following tasks: takeoff, hovering in place, yawing in place, pitching forward and back, rolling left and right, ascending and descending, camera movement, and landing.
7. Record the outcome of each task attempt.
8. Reposition the sUAS at the initial position if needed (i.e., if it did not land accurately).
9. Move OCU to the next position.
10. Repeat steps 4 through 9 until both the control link quality and video link quality fail.

## Example Data

- Environment characterization: Using the example layouts in Figure 1

| Metrics | Horizontal, through walls | | | | | Vertical, through floors | | | |
|---|---|---|---|---|---|---|---|---|---|
| | X-1 | X-2 | X-3 | X-4 | X-5 | X-1 | X-2 | X-3 | X-4 |
| Position distance (m) | 14 | 20 | 25 | 27 | 31 | 5 | 9 | 13 | 17 |
| Position obstructions | 1 drywall | 2 drywall | 3 drywall | 4 drywall | 4 drywall<br>1 concrete | 1 concrete | 2 concrete | 3 concrete | 4 concrete |

- Performance data:

| sUAS | Metrics | Horizontal, through walls | | | | | | Vertical, through floors | | | | |
|---|---|---|---|---|---|---|---|---|---|---|---|---|
| | | X | 1 | 2 | 3 | 4 | 5 | X | 1 | 2 | 3 | 4 |
| sUAS A | Connect | ✓ | ✓ | ✓ | X | X | X | ✓ | X | X | X | X |
| | Fly | ✓ | ✓ | ✓ | X | X | X | ✓ | X | X | X | X |
| | Maximum NLOS performance | 20 m, 2 walls | | | | | | 0 | | | | |
| sUAS B | Connect | ✓ | ✓ | ✓ | ✓ | ✓ | ✓ | ✓ | ✓ | ✓ | ✓ | ✓ |
| | Fly | ✓ | ✓ | ✓ | ✓ | ✓ | ✓ | ✓ | ✓ | ✓ | ✓ | ✓ |
| | Maximum NLOS performance | 31 m, 5 walls | | | | | | 17 m, 4 floors | | | | |

# Non-Line-of-Sight (NLOS) Video Latency

## Purpose

This test method measures the latency of video transmitted from the sUAS to the OCU when indoors in NLOS range transmitting data through walls and floors.

## Summary of Test Method

This test method is an expansion of an existing test method currently under development by NIST for standardization through the ASTM E54.09 Committee on Homeland Security; Subcommittee on Response Robots. In that test method, a flashing light is placed within view of the sUAS and an external camera is used to record the flashing light and the OCU display of the flashing light as seen by the sUAS camera in the same view. The sUAS and light are positioned further and further apart from the OCU while still maintaining that the light and OCU screen are visible in the external camera view to evaluate the impact of range on video latency. The external camera records while the light flashes several times. The video is then exported and the amount of delay between when the light actually flashes compared to when it is seen flashing on the OCU screen is calculated by counting video frames and converting to milliseconds (based on the frames per second of the recorded video).

This test method adapts the existing method for NLOS operations by instead using two synchronized stopwatches (with millisecond displays) rather than camera flashing lights, which will move between the different rooms and floors that separate the sUAS and the OCU. An external video camera captures one of the stopwatches and the OCU display in a single frame. Once the watches are synchronized, the sUAS and other stopwatch are moved into position, pointing the sUAS camera at the stopwatch such that both stopwatches can be seen in the external camera view back at the starting point. After all positions are completed, the video from that camera is then exported and evaluated the same as previously described (i.e., counting time difference between the two).

This test method can be run concurrently with the NLOS Communications test method.

## Apparatus and Artifacts

A real-world indoor environment deemed relevant for the use case that contains multiple rooms, floors, and/or passageways that are separated by solid wall and/or floor boundaries (e.g., office spaces, multi-level buildings, underground tunnel systems). Minimal cellular coverage and WiFi signals are desirable to limit potential communications interference. Positions in the environment should be identified that progressively add new obstructions for horizontal NLOS (e.g., walls that separate rooms and hallways) and vertical NLOS (e.g., floors of a building, starting from the bottom floor and moving upwards). A single position (X) is used for the sUAS location and the other positions (1, 2, 3, etc.) should be accessible by the operator with the OCU.

Example environments are shown in Figure 1. The horizontal NLOS environments consist of a multi-room office space (the walls are made of drywall and wood) and a series of underground tunnels with connected passageways (the walls are made of concrete, with building infrastructure and earth outside of the walls). The vertical NLOS environment is evaluated across multiple floors of an office building (the ceilings/floors are made out of concrete). The straight line distance through obstructions between the sUAS position (X) and the OCU positions (1, 2, 3, etc.) is shown and the number of obstructions counted.

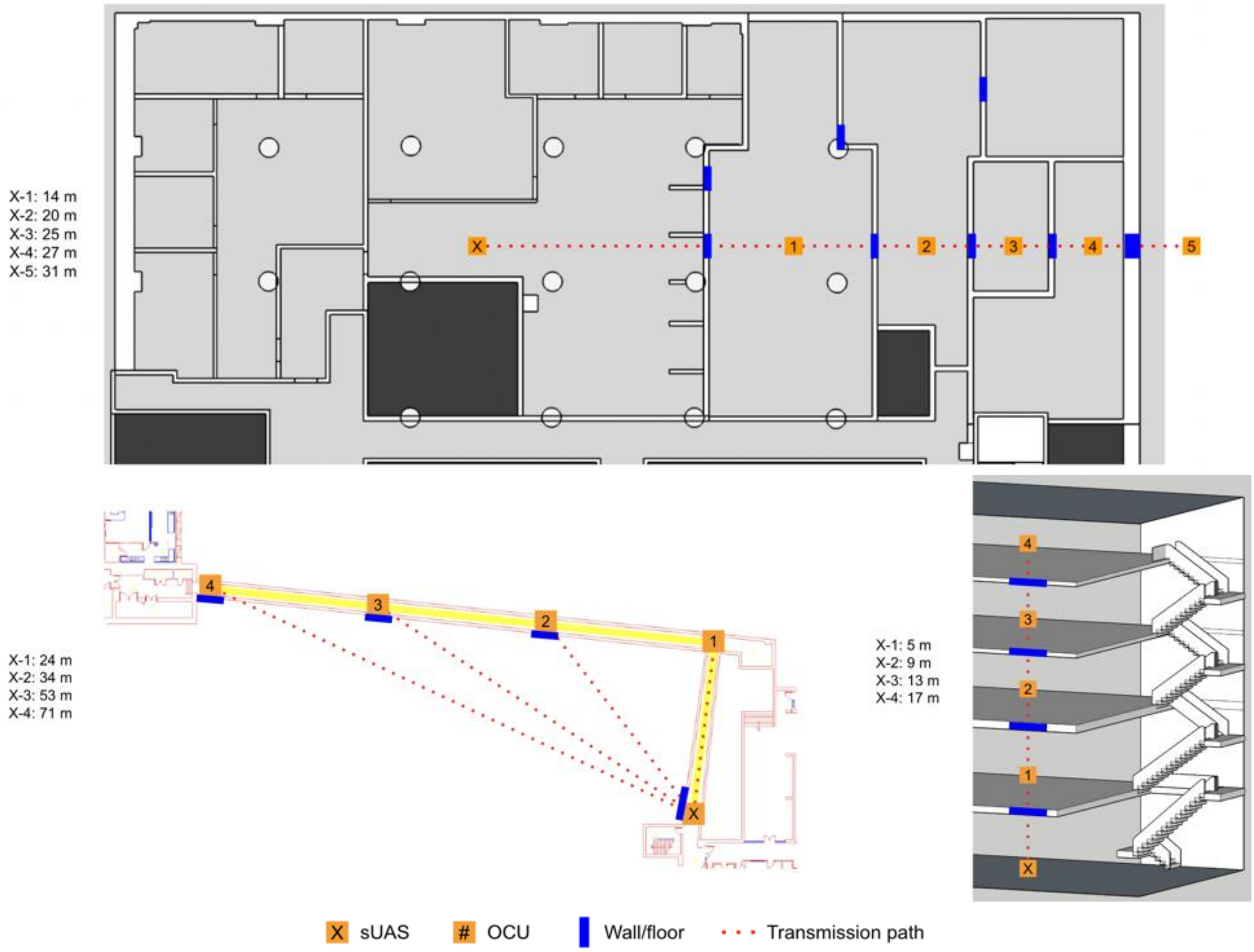

*Figure 1. Example layouts for horizontal NLOS (top and bottom left) and vertical NLOS (bottom right) positions in a real-world environment.*

## Equipment

A video camera that can record video at 30 frames per second or higher is needed.

Two stopwatches that display milliseconds must be used.

## Metrics

- Average video latency: For each position, calculate the average video latency with standard deviation measured across at least 10 trials, reported in milliseconds.
    - Inter-camera video latency: The average video latency measured between the two external cameras when streamed to the video multiviewer and recorded.
- Position distance: Distance between the sUAS position and the OCU position
- Position obstructions: Number of wall and/or floor boundaries between the sUAS position and the OCU position and the type of material they are made out of.
- Maximum NLOS latency: The average video latency metric measured at the maximum achievable NLOS operational range, i.e., where video and control link quality are good (✓) and task performance is possible (✓), derived from the NLOS Communications test method. Report this metric alongside corresponding maximum NLOS performance metric from that test method (i.e., the maximum position distance and number of position obstructions).

## Procedure

1. Mark the desired NLOS testing positions throughout the environment and record each position's distance and obstructions.
2. Synchronize both stopwatches.
3. Connect the sUAS and the GCS/OCU.
4. Position one stopwatch at the initial position along with the OCU and point the external camera such that both are visible in the video frame and begin recording.
5. At the first position, record at least 10 seconds of the stopwatch counting
6. Move OCU and external camera to the next position.
7. Repeat steps 6 through 9 until connection to the sUAS is no longer possible.
8. Export the videos and calculate the metrics.

## Example Data

- Environment characterization: Using the example layouts in Figure 1

| | **Horizontal, through walls** | | | | | **Vertical, through floors** | | | |
|---|---|---|---|---|---|---|---|---|---|
| **Metrics** | **X-1** | **X-2** | **X-3** | **X-4** | **X-5** | **X-1** | **X-2** | **X-3** | **X-4** |
| Position distance (m) | 14 | 20 | 25 | 27 | 31 | 5 | 9 | 13 | 17 |
| Position obstructions | 1 drywall | 2 drywall | 3 drywall | 4 drywall | 4 drywall<br>1 concrete | 1 concrete | 2 concrete | 3 concrete | 4 concrete |

- Performance data:

| | | **Horizontal, through walls** | | | | | | **Vertical, through floors** | | | | |
|---|---|---|---|---|---|---|---|---|---|---|---|---|
| **sUAS** | **Metrics (ms)** | **X** | **1** | **2** | **3** | **4** | **5** | **X** | **1** | **2** | **3** | **4** |
| sUAS A | Latency | 200 | 200 | 200 | X | X | X | 200 | X | X | X | X |
| | Maximum NLOS latency | 200 ms<br>20 m, 2 walls | | | | | | 200 ms<br>0 | | | | |
| sUAS B | Latency | 200 | 200 | 200 | 300 | 300 | 500 | 200 | 300 | 300 | 2500 | X |
| | Maximum NLOS latency | 500 ms<br>31 m, 5 walls | | | | | | 2500 ms<br>13 m, 3 floors | | | | |

# Interference Reaction

## Purpose

When sUAS that operate indoors, especially those that communicate in the 2.4 GHz ISM band, they can be susceptible to interference from WiFi and other transmitters that routinely operate indoors. Interference can potentially degrade both the video channel used to control the vehicle beyond line of sight and can degrade the communication channel used to control the unit. The objective of this test is to add signal interference to the radio wave channels used to communicate between the sUAS and the OCU for camera data and control, helping to understand the vulnerability of individual sUAS to potential interferences to be encountered in the field including counter-UAS (cUAS) attacks. It will help understand the robustness of the individual waveforms used by the sUAS and the interference mitigation strategies implemented by the different units.

## Summary of Test Method

The test consists of generating an interfering radio-frequency signal whose frequency will fall within the sUAS communication camera or control channels (i.e., jamming its communication channel). The possible outcomes of sUAS behavior once jammed include exhibiting lost or degraded communication functionality (e.g., landing, return to home), automatic channel hopping to deconflict with the interfering signal, or inability to reconnect after interference has ceased (i.e., sUAS needs to be restarted before connection is regained). There are multiple types of interference tests that are performed (note that each test serves as a prerequisite for running the subsequent tests; e.g., run the Hovering test before running the Command Input test):

<u>Frequency Characterization</u>: Before the sUAS signal can be interfered with, a receiver antenna connected to a spectrum analyzer can be used to determine exactly which frequencies are being used by the sUAS to operate.

<u>Grounded Interference</u>: While the sUAS is grounded, transmit the interfering signal and attempt to take off.

<u>Hovering Interference</u>: While the sUAS is hovering, transmit the interfering signal and attempt to continue hovering in place, yaw, pitch forward and back, roll left and right, ascend and descend, move the camera, and then land.

<u>Command Input Interference</u>: Command the sUAS to continuously yaw in place while hovering, then proceed to transmit the interfering signal. Note whether the sUAS either continues to turn or stops if and when control is lost.

Note: running these tests may severely degrade existing WiFi networks in the area where testing is conducted.

## Apparatus and Artifacts

No particular apparatus is needed for running this test aside from an environment big enough to fit the distance criteria noted for each test in the summary section.

The distances between the sUAS, OCU, receiver antenna, and transmitting antenna should be measured, recorded, and fixed for all tests.

## Equipment

A spectrum analyzer with a low noise preamplifier between the antenna and the analyzer as shown in Figure 1 is used to determine the communication frequencies and bandwidths used by the sUAS and OCU to communicate and the presence and strengths of background interference sources.

Interfering signals can be generated in several ways: (1) using a software defined radio (SDR) that is programmed to transmit an interfering waveform such as WiFi or Bluetooth; (2) a signal generator set to transmit a relatively wideband waveform of 10-20 MHz (a fast chirp waveform would meet this requirement), or (3) using a standard WiFi modem and a long ethernet cable that can be moved into the proximity of the sUAS.

An example set up (option 1 described above) is shown in Figure 1, using a low cost “Pluto” software defined radio (SDR) connected to an amplifier capable of outputting 100-200 mW and connected to standard commercial WiFi/cellular antenna, configured so the modem is continuously transmitting. Components of the example set up shown in Figure 1 include the SDR, several different antennas, the spectrum analyzer with pre-amplifier, and power amplifier connected to the SDR.

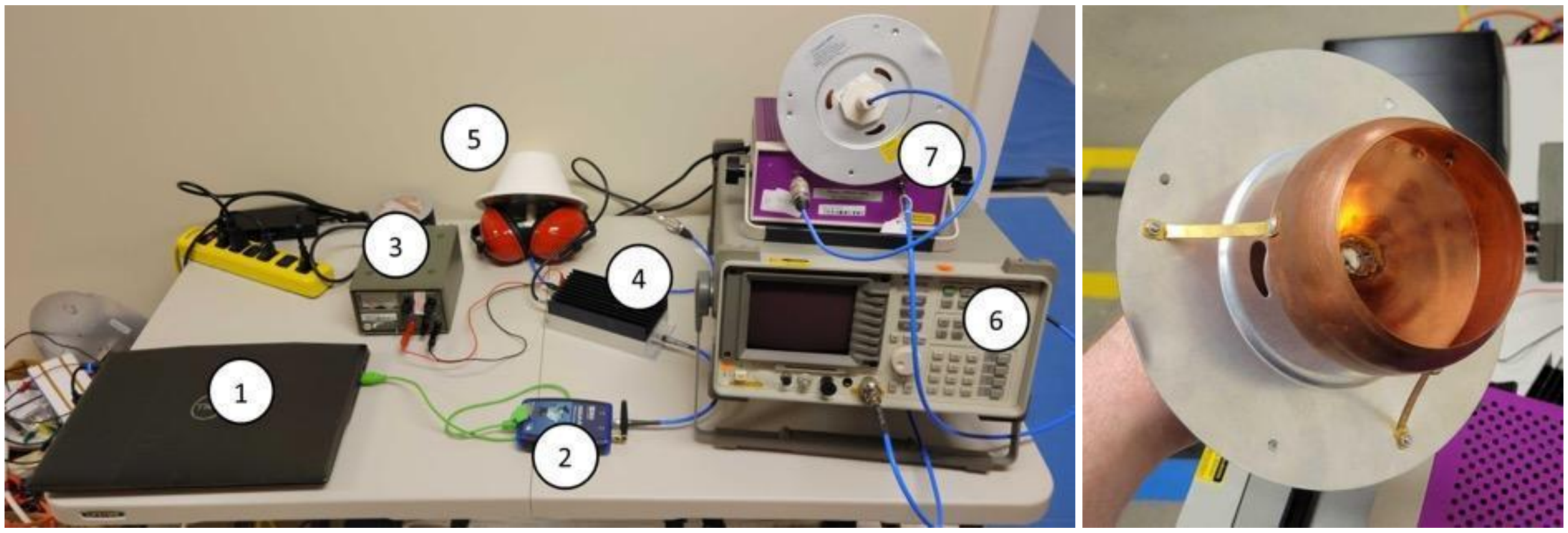

*Figure 1. Left: Example jammer equipment set up: (1) laptop for SDR running Matlab, (2) Pluto SDR used to generate WiFi signal, (3) power supply, (4) RF power amplifier (20-30 dB), (5) transmit antenna connected to power amplifier, (6) spectrum analyzer, (7, and right image) Receiver antenna and preamplifier.*

## Metrics

- Channel hopping: Evaluate whether or not the sUAS automatically changes its radio frequency channel to deconflict with the interfering signal while being interfered. If the unit does change channel, report the frequency it hopped to.
- Flight control: Ability of the operator to send control commands through the OCU to the sUAS while interference is active. Qualitatively evaluated as either good (✓), bad (//), or none (X). For the ground interference test, this can be evaluated by attempting to move the sUAS camera gimbal, zoom, etc. For the other tests, this can be evaluated by attempting the specified tasks.
- Video link: Visual appearance of the video stream of the sUAS camera on the OCU while interference is active. Qualitatively evaluated as either good (✓), bad (//), or none (X). Indicators of “bad” video link quality include screen tearing, pixelation, and other artifacts not present when video link quality is “good.”
- Takeoff: Applies only to the grounded test condition. Denotes whether the sUAS is able to takeoff while the interference signal is being transmitted.
- Auto land: Applies to the hover and yaw conditions. Denotes whether the sUAS automatically lands itself when the interference signal is transmitted.
- Command input persistence: Whether or not the sUAS continues to yaw in place while being jammed during the yaw test condition.
- Interference Power Level: Each test condition should be conducted using the same power level of interference signal at the same distances between sUAS, OCU, and antennas. Multiple power levels can be tested. The strength of the signal should be characterized in a manner similar to characterizing the sUAS signal frequency and strength.

## Procedure

Frequency Characterization:

1. Using a spectrum analyzer with preamplifier and external antenna, take ambient readings of existing radio signals present in the test environment (e.g., nearby WiFi access points) and record the measurements for environment characterization purposes. Set the Trace mode of the spectrum analyzer to “Maxhold” because interfering signals tend to be intermittent. Identify any extraneous interference sources before turning on the sUAS or jammer signal. The frequency range being inspected can be based on the manufacturer specifications, or can be progressively narrowed down, by starting at a very wide range, and zooming in on spikes that appear when the sUAS is powered on.
2. Power on the sUAS and OCU and establish connection.

3. Use the spectrum analyzer to determine the sUAS radio channel frequencies being utilized and record the measurements. Set the trace mode of the spectrum analyzer to “maxhold”, note the peak frequencies.
4. Tune the signal generator (i.e., jammer) to emit a frequency that falls within the sUAS radio channel.
5. Power down the sUAS and OCU.
6. Turn on the jammer to begin emitting an interfering signal and use the spectrum analyzer to verify the signal is being emitted at the intended frequency and power level. Tune as needed to match the intended frequency and power level, then turn off the jammer to cease transmitting the interfering signal.

Grounded Interference:

7. Position the sUAS and OCU at the specified distance away from the jammer antenna.
8. Power on the sUAS and OCU and establish connection.
9. Begin transmitting the interfering signal from the jammer.
10. Evaluate the quality of the video link and whether controls are still responsive.
11. If connection is maintained, attempt commanding the sUAS to takeoff.
12. Cease transmitting the interfering signal from the jammer.
13. Record the metrics.

Hovering Interference:

7. Position the sUAS and OCU at the specified distance away from the jammer antenna.
8. Power on the sUAS and OCU and establish connection.
9. Command the sUAS to takeoff and hover in place.
10. Begin transmitting the interfering signal from the jammer.
11. If connection is maintained, attempt the following tasks: hovering in place, yawing in place, pitching forward and back, rolling left and right, ascending and descending, camera movement, and landing.
12. Cease transmitting the interfering signal from the jammer.
13. Record the metrics.

Command Input Interference:

7. Position the sUAS and OCU at the specified distance away from the jammer antenna.
8. Power on the sUAS and OCU and establish connection.
9. Command the sUAS to takeoff and yaw in place.
10. Begin transmitting the interfering signal from the jammer.
11. If connection is maintained, attempt the following tasks: hovering in place, yawing in place, pitching forward and back, rolling left and right, ascending and descending, camera movement, and landing.
12. If connection is lost, denote whether the sUAS continues to yaw in place.
13. Cease transmitting the interfering signal from the jammer.
14. Record the metrics.

## Example Data

- Environment characterization: 2.4 GHz signal present in space due to WiFi access points
- Interference characterization:

| Low Power Signal | High Power Signal |
|---|---|
| -25.61 dBm 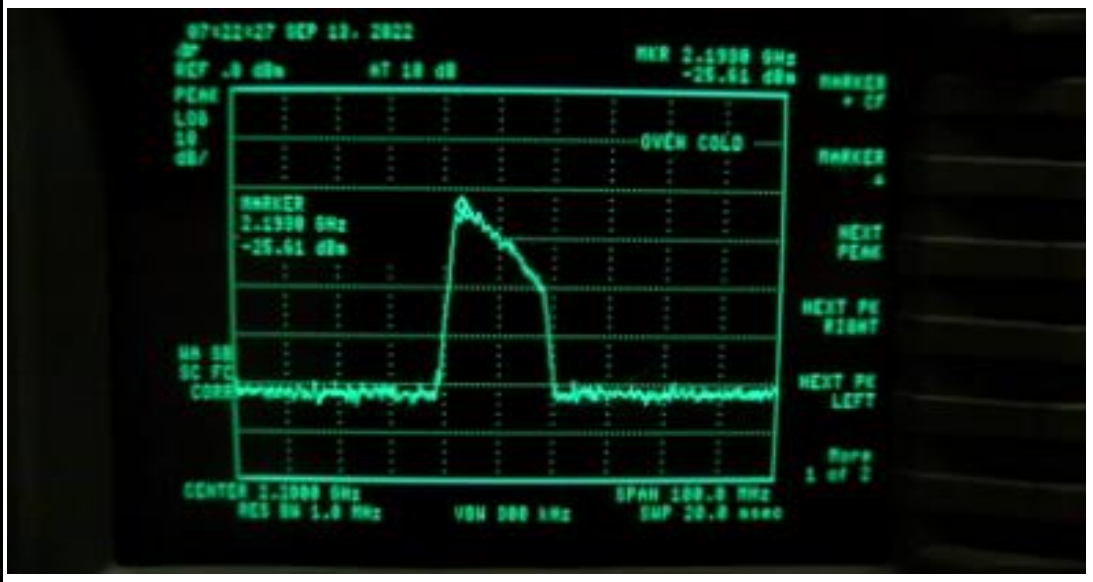 | 13.00 dBm (85x stronger) 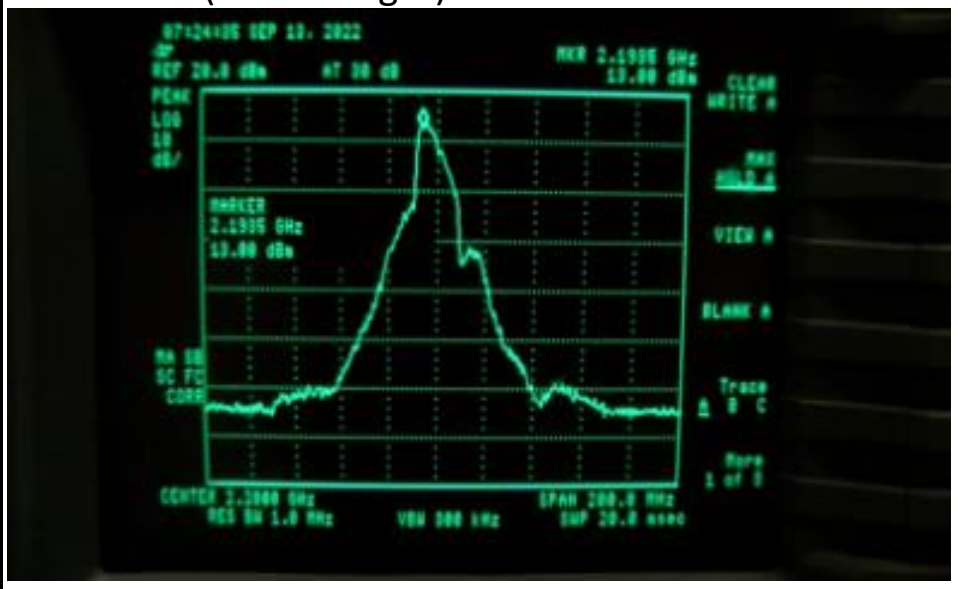 (note: frequency span is twice that of the low power image) |

- Performance data:

| sUAS | | Ambient Signal | | | sUAS Signal | | |
|---|---|---|---|---|---|---|---|
| sUAS A | Image | 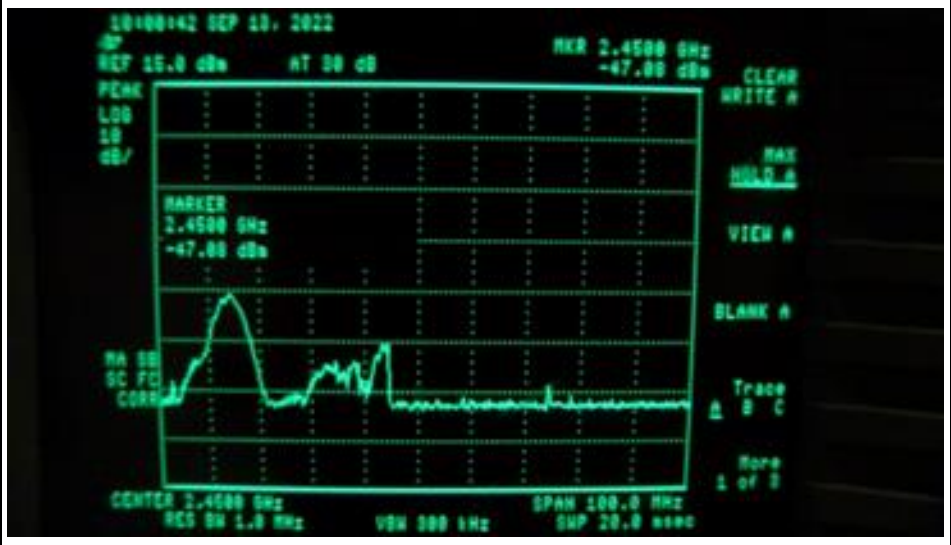 | | | 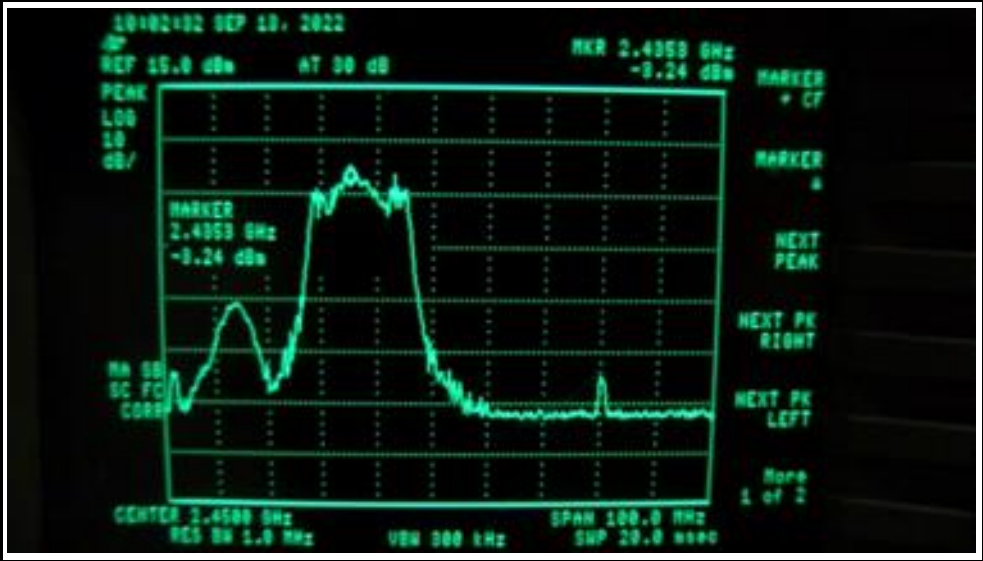 | | |
| | Freq Start (GHz) | 2.4 | | | 2.4 | | |
| | Freq End (GHz) | 2.5 | | | 2.5 | | |
| | Peaks (GHz) | | | | 2.435 | | |
| | | sUAS with Low Power Interference | | | sUAS with High Power Interference | | |
| | Image | 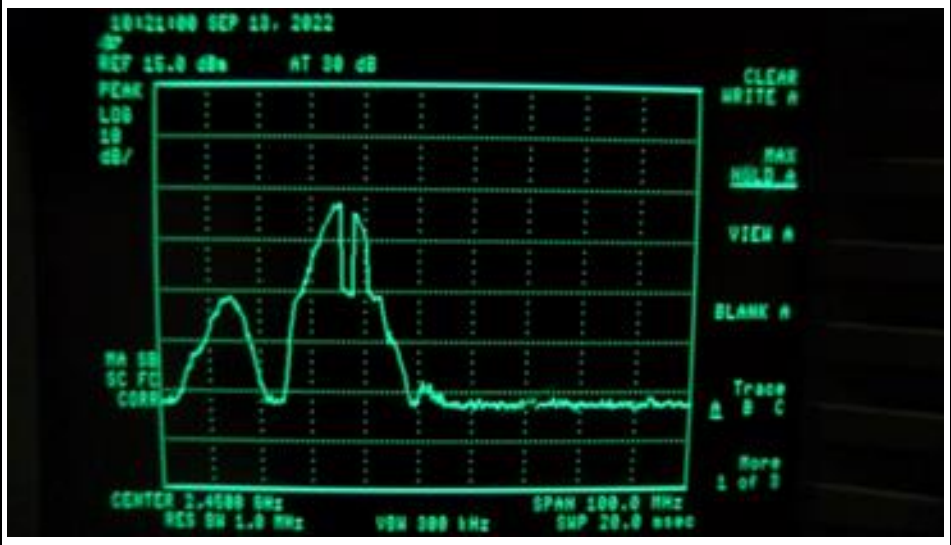 | | | 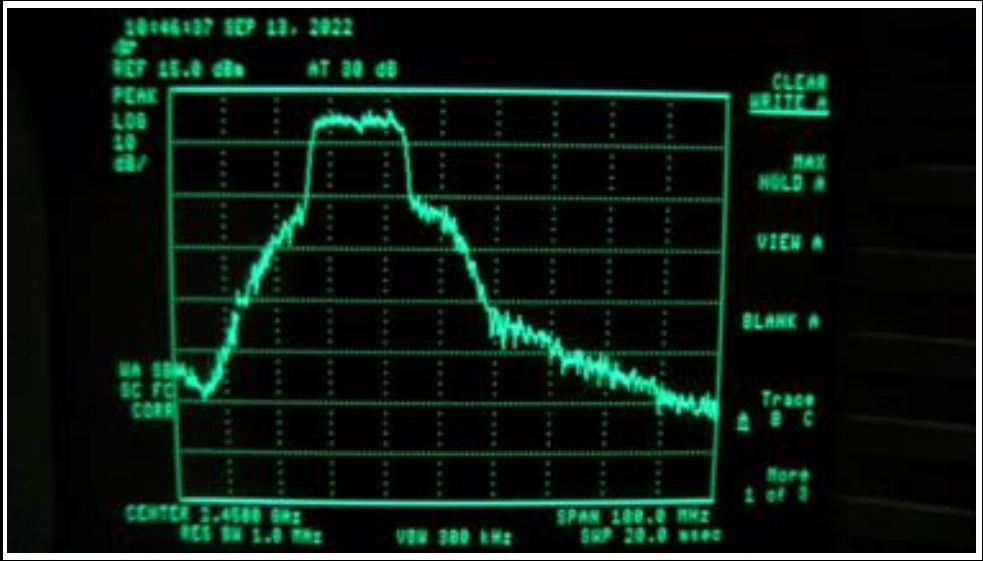 | | |
| | Freq Start (GHz) | 2.4 | | | 2.4 | | |
| | Freq End (GHz) | 2.5 | | | 2.5 | | |
| | Peaks (GHz) | | | | | | |
| | Metrics | Grounded | Hover | Yaw in Place | Grounded | Hover | Yaw in Place |
| | Takeoff | X | - | - | X | - | - |
| | Video Link | X | ✓ | // | X | X | X |
| | Flight Control | X | ✓ | ✓ | X | X | X |
| | Auto Land | - | X | X | - | ✓ | ✓ |
| | Channel Hop | X | X | X | X | X | X |
| | Input Persist | - | - | N/A | - | - | ✓ |

| sUAS | | Ambient Signal | | | sUAS Signal | | |
|---|---|---|---|---|---|---|---|
| sUAS B | Image | 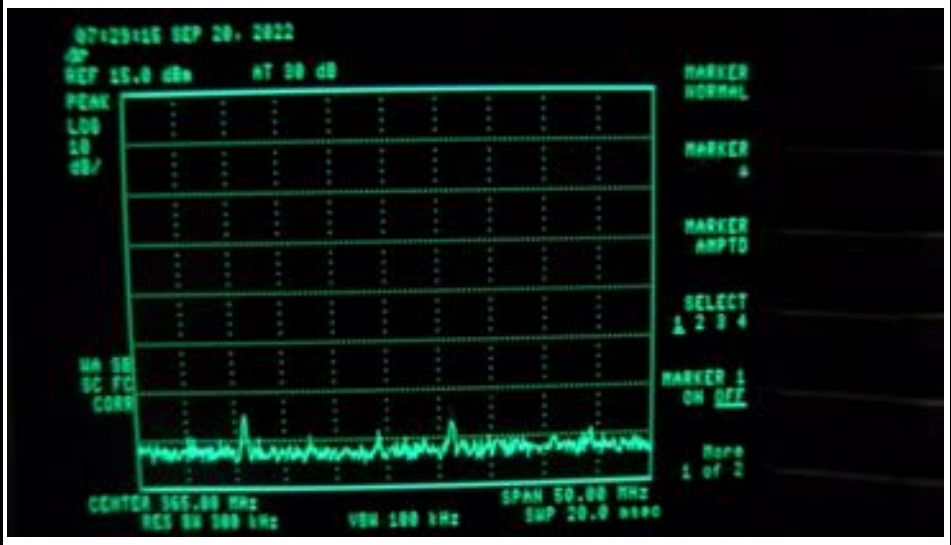 | | | 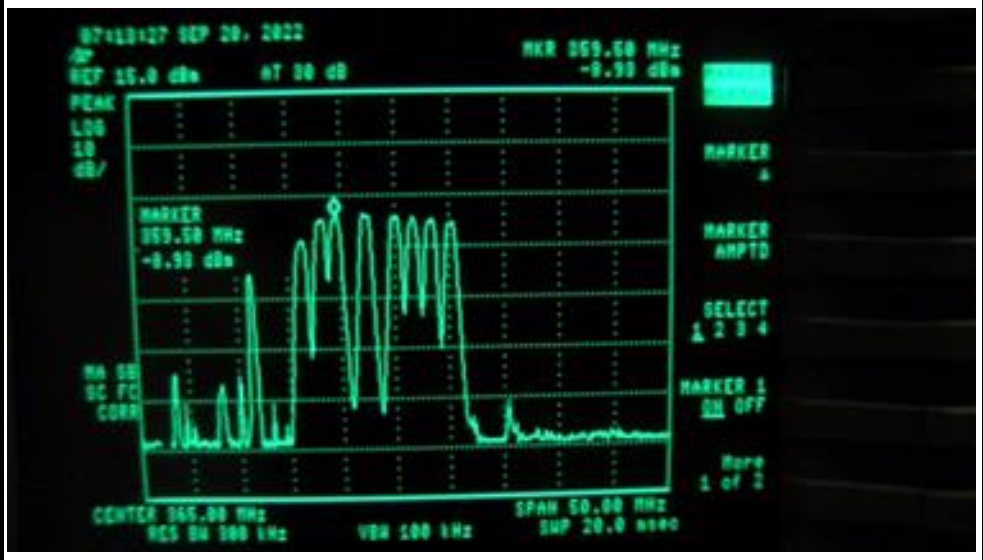 | | |
| | Freq Start (MHz) | 340 | | | 340 | | |
| | Freq End (MHz) | 390 | | | 390 | | |
| | Peaks (MHz) | | | | 355 - 372 | | |
| | | sUAS with Low Power Interference | | | sUAS with High Power Interference | | |
| | Image | 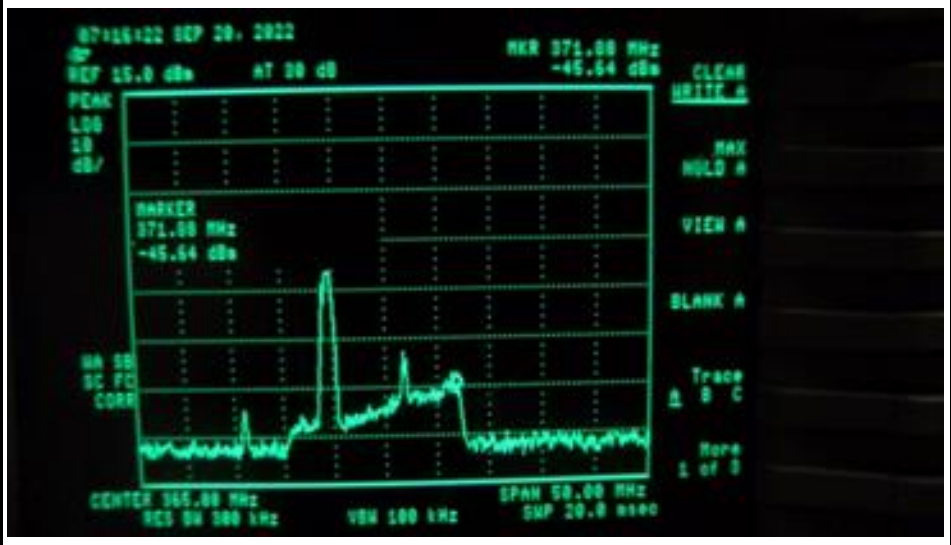 | | | 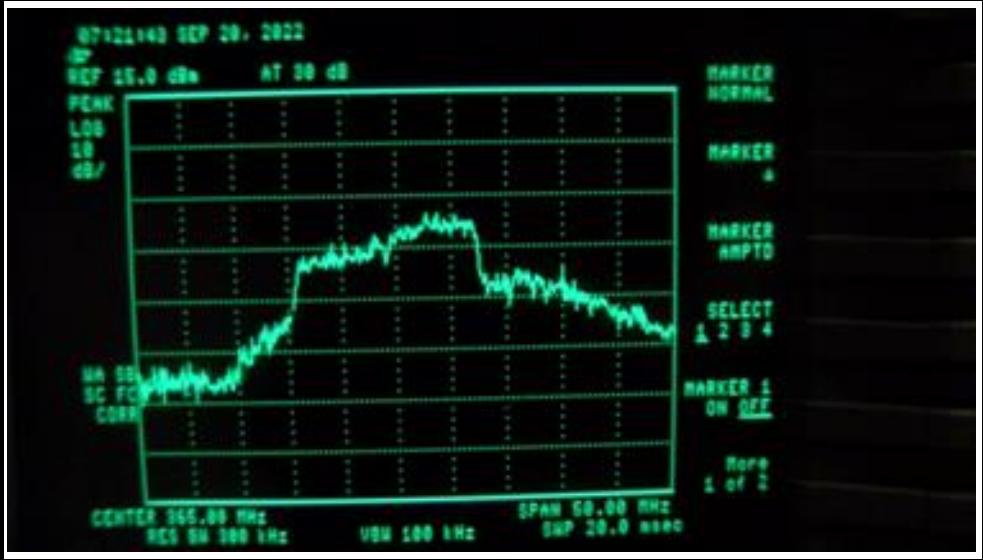 | | |
| | Freq Start (MHz) | 340 | | | 340 | | |
| | Freq End (MHz) | 390 | | | 390 | | |
| | Peaks (MHz) | | | | | | |
| | Metrics | Grounded | Hover | Yaw in Place | Grounded | Hover | Yaw in Place |
| | Takeoff | ✓ | - | - | ✓ | - | - |
| | Video Link | ✓ | ✓ | ✓ | ✓ | X | X |
| | Flight Control | ✓ | ✓ | ✓ | // | X | X |
| | Auto Land | - | X | X | - | ✓ | ✓ |
| | Channel Hop | ✓ | ✓ | ✓ | ✓ | ✓ | ✓ |
| | Input Persist | - | - | N/A | - | - | X |

# Field Readiness

Field readiness comprises several factors for sUAS which will determine whether or not the systems can be effectively utilized for missions in subterranean and other tight, indoor environments. These factors include those related to runtime on a single battery, performance capabilities that can ensure minimal functionality in confined, GPS-denied environments (e.g., ability to perch, land, and remain static for a period of time), the ability to inspect rooms in the environment, and logistics considerations to packing, transporting, and maintaining the systems (e.g., assets that are expendable or non-expendable but easily repairable for predictable appendages). The results of these tests can be used to evaluate the applicability of a sUAS platform to particular mission requirements.

## Runtime Endurance

### Purpose

This test method evaluates sUAS battery life under various operational contexts in order to produce a spread of performance to understand expected mission length.

### Summary of Test Method

The sUAS is continuously maneuvered either for flight throughout an environment or camera movement when stationary to inspect an environment. Three types of Runtime Endurance tests are specified; for all three tests, the operator maneuvers the sUAS as described until either (a) the battery life is exhausted, or (b) the OCU warns the operator of low battery, requiring the sUAS to be flown back to the launch point. All tests can be run in lighted (100 lux or greater) or dark (less than 1 lux) conditions. The three tests are:

Indoor Movement: In an indoor environment, the sUAS navigates within a series of poles and two apertures to form a figure-8 path. The obstacle avoidance settings on the sUAS can be varied (e.g., 1 m clearance to obstacles, no obstacle avoidance, etc.).

Hover and Stare Activities: In either an outdoor or indoor environment, the sUAS launches and hovers at a waypoint, then the operator maneuvers the camera to predefined position and zoom settings, idles for 2 minutes, and then maneuvers to the next camera position, idles again for 2 minutes, and repeats this process. The sUAS can be configured as GPS-enabled or VIO-enabled.

Perch and Stare Activities: In either an outdoor or indoor environment, the sUAS launches, lands on a platform, and the operator maneuvers the camera to predefined position and zoom settings, idles for 2 minutes, and then maneuvers to the next camera position, idles again for 2 minutes, and repeats this process. The sUAS can be configured as GPS-enabled or VIO-enabled.

Metrics from all four tests should be reported in order to demonstrate a spread of expected performance duration, pending the type of mission being performed.

### Apparatus and Artifacts

The apparatuses for each test are specified as follows (see Figure 1):

Indoor Movement: Eight vertical posts at least 2.4 m tall (e.g., wooden 2x4 studs) are positioned 2 m from each other in the configuration shown in Figure 1, in an indoor environment. Two apertures measuring 2 m x 1 m are formed between two pairs of posts, one lower and one higher, with red tape on the side of each aperture facing the other. One figure-8 lap = nominal 13 m distance traveled.

Hover and Stare Activities: One raised platform is positioned on the ground in an outdoor or indoor environment.

Perch and Stare Activities: Two raised platforms are positioned on the ground separated by 4 m center-to-center, in an outdoor or indoor environment.

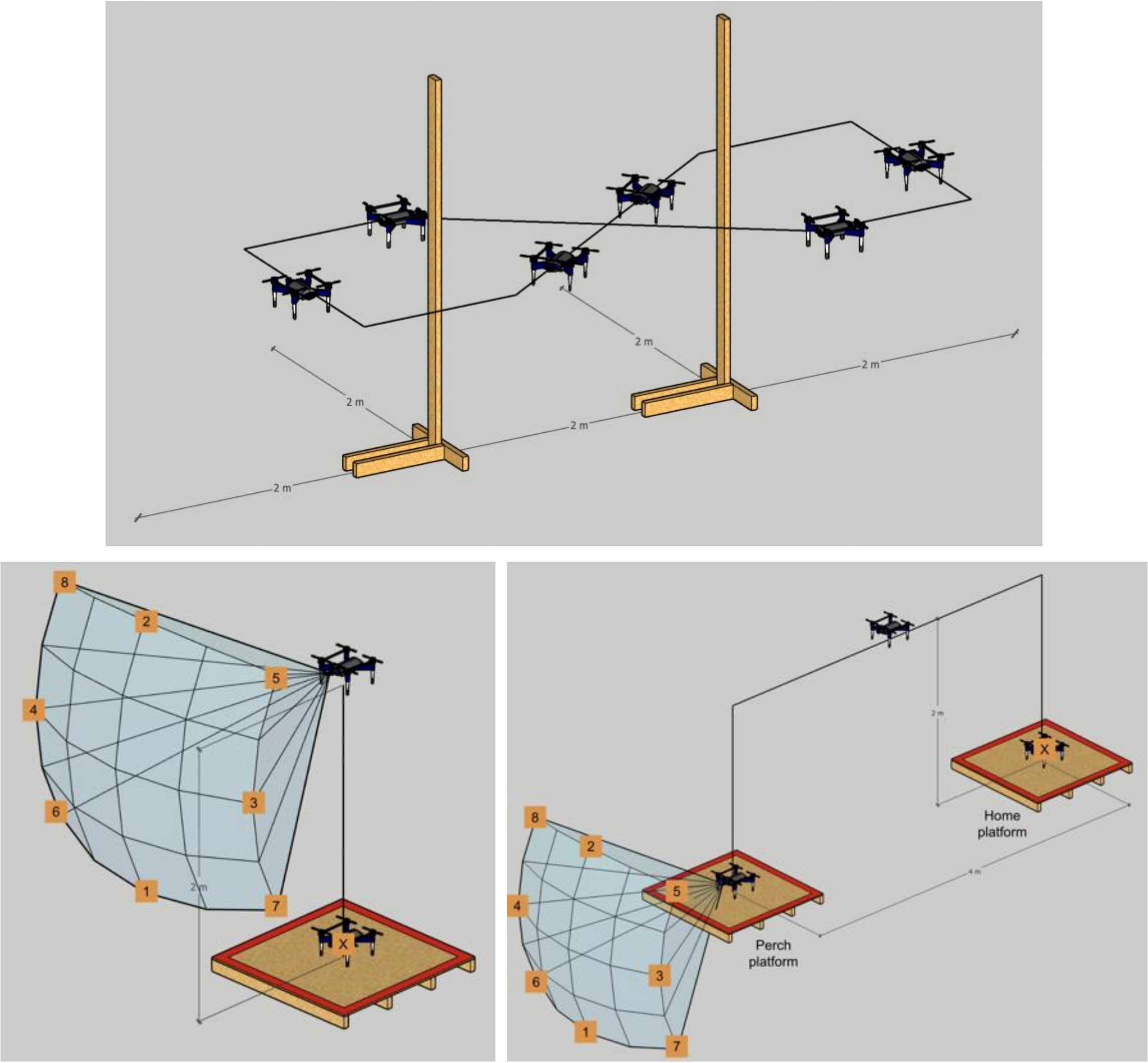

*Figure 1. Apparatuses for the Runtime Endurance test method: indoor movement (top), hover and stare activities (bottom left), perch and stare activities (bottom right).*

## Equipment

A timer is used to record the duration metric.

## Metrics

- <u>Duration</u>: The amount of time to perform the test, starting from when the sUAS launches until either (a) the battery life is exhausted and the sUAS powers down, or (b) the operator returns the sUAS to the launch point due to low battery (i.e., battery life is about to be exhausted). This metric is reported in minutes. Note: most sUAS manufacturers recommend ceasing operation once a certain battery level is reached; the operator should follow these guidelines in order to reach state (b) rather than (a), if possible.
- <u>Distance traveled</u>: The approximate amount of distance traveled during the Indoor Movement tests, measured by counting the number of laps completed and multiplying that by the nominal length of each figure-8 flight path (approximately 13 m). This metric is reported in meters.
- <u>Average traversal speed</u>: This metric is calculated by dividing the distance traveled metric by the duration metric (converted to seconds), reported in meters per second. This metric is applicable to the Indoor Movement test only.

- OCU notification: Whether or not the OCU interface provided a notification of low battery during the test, which could be in the form of a percentage reading, flashing screen warning, audio alert, etc.
- Return to start: Whether or not the sUAS returned to the launch point before the battery expired.
- Collisions: The number of times the sUAS collided with the apparatus (e.g., vertical pole, aperture boundary, raised platform). Note: landing on the platform in the Perch and Stare Activities test does not constitute a collision.

## Procedure

Indoor Movement:

1. Position the posts and apertures 2 m from one another as shown in Figure 1.
2. Launch the sUAS, start the timer, and ascend the sUAS to desired elevation within the 2.4 m limit.
3. Navigate around the vertical poles and through the red border apertures, changing elevation as needed, to form a figure-8 path around the poles, counting the number of figure-8 laps completed.
4. Continue until either (a) the battery expires preventing continued operation flight, or (b) the battery reaches the level wherein the sUAS manufacturer recommends ceasing operations.
5. If state (a) is reached, land the sUAS in place (if possible), stop the timer, and record the duration metric. If state (b) is reached, maneuver the sUAS back to the launch point, land, stop the timer, and record the duration metric.
6. Calculate the metrics.

Hover and Stare Activities:

1. Position the platform in an open area.
2. Launch the sUAS, start the timer, and ascend the sUAS to a nominal 2 m elevation.
3. Maneuver the sUAS camera to maximum position [1] downward, [2] upward, [3] left, [4] right, [5] top-left, [6] bottom-right, [7] bottom-left, or [8] top-right and zoom the camera in to maximum zoom level.
4. Maintain this camera position for 2 minutes.
5. Zoom the camera out to minimum zoom level.
6. Repeat steps 3-5, moving to the next sequential [#] maximum position until all available positions have been exhausted.
7. Repeat steps 3-6, starting at the same first maximum position, until the battery reaches the level wherein the sUAS manufacturer recommends ceasing operations.
8. When this state is reached, land, stop the timer, and record the duration metric.
9. Calculate the metrics.

Perch and Stare Activities:

1. Position the home and perch platforms 4 m center-to-center from one another.
2. Launch the sUAS, start the timer, and ascend the sUAS to a nominal 2 m elevation.
3. Navigate to the perch platform and land on it.
4. Maneuver the sUAS camera to maximum position [1] downward, [2] upward, [3] left, [4] right, [5] top-left, [6] bottom-right, [7] bottom-left, or [8] top-right and zoom the camera in to maximum zoom level.
5. Maintain this camera position for 2 minutes.
6. Zoom the camera out to minimum zoom level.
7. Repeat steps 4-6, moving to the next sequential [#] maximum position until all available positions have been exhausted.
8. Repeat steps 4-7, starting at the same first maximum position, until either (a) the battery expires preventing continued operation flight, or (b) the battery reaches the level wherein the sUAS manufacturer recommends ceasing operations.
9. If state (a) is reached, stop the timer, and record the duration metric. If state (b) is reached, launch the sUAS to a nominal 2 m elevation, navigate back to the home platform, land, stop the timer, and record the duration metric.
10. Calculate the metrics.

## Example Data

- Environment characterization: Lighted
- Performance data:

| sUAS | Test / Metric | Indoor Movement | | | Hover and Stare | Perch and Stare |
|---|---|---|---|---|---|---|
| | | Duration (min) | Distance (m) | Average speed (m/s) | Duration (min) | Duration (min) |
| sUAS A | | 8 | 260 | 0.5 | 9 | 182 |
| sUAS B | | 32 | 299 | 0.2 | 30 | 315 |
| sUAS C | | 31 | 754 | 0.4 | 29 | 222 |

## Takeoff and Land/Perch

### Purpose

This test method is used to determine under what environmental conditions the sUAS is able to takeoff from in order to initiate flight and to land/perch on (e.g., uneven ground, nearby obstructions), which may be inhibited by built-in safety check functionality or stabilization issues.

### Summary of Test Method

A series of conditions are specified that define variations in the ground plane (pitch/roll angle, sensor interference due to material/proximity to external electronics) and nearby obstructions (overhead or lateral obstructions). These variations in the environment can impact a system's ability to takeoff and/or land/perch as it is common for sUAS to have built-in functionality that checks for level ground and/or the presence of obstructions nearby. Such functionality may not allow for the system to launch due to safety concerns (e.g., sUAS behavior may require ascension to a certain height upon takeoff before continuing to operate), or systems without such safety precautions may allow them to attempt takeoff and land/perch regardless of the environment, which may cause collisions or rollovers. Ten conditions are specified in terms of ground plane material, angle, and obstructions:

| Condition | Ground plane material | Ground plane angle | Obstructions |
|---|---|---|---|
| 1 | Wood | 0° | None |
| 2 | Metal with embedded electronics | 0° | None |
| 3 | Wood | 5° roll | None |
| 4 | Wood | 5° pitch | None |
| 5 | Wood | 10° roll | None |
| 6 | Wood | 10° pitch | None |
| 7 | Wood | 0° | 1.2 m (4 ft) overhead |
| 8 | Wood | 0° | 2.4 m (8 ft) overhead |
| 9 | Wood | 0° | 0.6 m (2 ft) lateral |
| 10 | Wood | 0° | 1.2 m (4 ft) lateral |

Two tests are specified:

Takeoff: Starting from a ground position on top of a platform, the sUAS attempts to launch into the air, navigate forward at least 3 m (10 ft).

Land/Perch: Starting hovering in the air at a position 3 m (10 ft) away from the landing platform, the sUAS attempts to navigate toward the platform and land on it.

Both tests can be performed back-to-back so long as takeoff is successful. Each test under each condition is performed multiple times to establish statistical significance and the associated probability of success and confidence levels based on the number of successes and failures (see the metrics section). Tests can be attempted in lighted (100 lux or greater) or dark (less than 1 lux) conditions.

Ideally, the sUAS will not collide with the boundaries of the space (i.e., walls, floor, and ceiling surfaces) during takeoff or landing/perching, but contact is allowed so long as it does not cause the sUAS to crash in a way that requires human intervention for it to resume flight.

## Apparatus and Artifacts

Either a real-world environment or fabricated apparatus can be used for each test condition, so long as the dimensional and positional requirements outlined below are maintained. The descriptions below assume a fabricated apparatus is used. See Figure 1 for a rendering of all test conditions.

For all conditions, a raised platform is used measuring 122 cm (48 in) square by 10 cm (4 in) tall, either made out of wood (conditions 1 and 3-10) or metal. The metal version of the platform also contains embedded electronics that are actively powered in order to exercise potential interference between the sUAS onboard sensors (e.g., magnetometer) and the environment (e.g., hood of an armored car), as this scenario has been shown to cause issues with some sUAS. Specification on the electronics in the metal platform are provided in the equipment section (note: these details have not yet been specified, but will be in a future version of this handbook). A series of wooden posts are used to angle the platform to be 5 or 10 degrees pitch/roll for conditions 3-6.

The overhead obstructions for conditions 7 and 8 using panels (e.g., wood, foam, any material that is a consistent color/surface texture throughout and flat) measuring at least 122 cm (48 in) square or larger, positioned 122 cm (48 in) or 244 cm (96 in) above the platform, hanging level to the platform (i.e., 0 degrees). The overhead obstructions can either hang from the ceiling of the indoor environment being used (e.g., using rope) or can be attached to a nearby wall. However, any nearby walls must not interfere with testing (i.e., potentially cause an issue with taking off or landing/perching in these conditions). If nearby walls are to be used for mounting the overhead obstructions, then condition 10 should first be run to determine if the presence of lateral obstructions 122 cm (48 in) away from the center of the platform does or does not interfere with performance. If not, then the overhead obstructions can be mounted from walls that are 122 cm (48 in) or further away from the center of the platform.

The lateral obstructions for conditions 9 and 10 are freestanding wall panels made of wood or any other material that is a consistent color/surface texture throughout and flat, measuring 122 cm (48 in) wide and at least 244 cm (96 in) tall. Condition 9 uses three wall panels while condition 10 uses six, as shown in Figure 14.

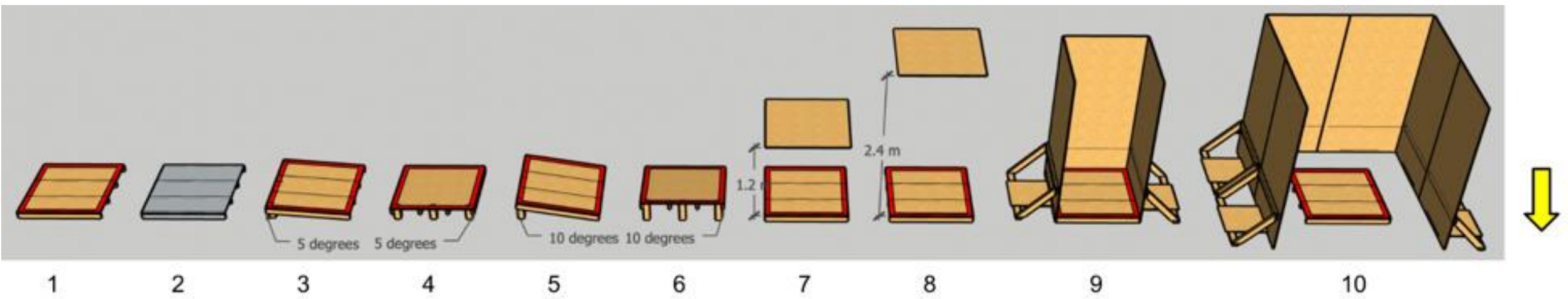

*Figure 1. Test apparatus conditions for the Takeoff and Perch/Land test method. The yellow arrow on the right indicates the required sUAS orientation when grounded at the start of the takeoff test.*

## Equipment

The metal platform used in condition 2 includes embedded electronics that are actively powered. Note: the details for this equipment has not yet been specified, but will be in a future version of this handbook.

## Metrics

- Takeoff efficacy: Whether or not the sUAS is able to takeoff from the platform without failure.
- Land/perch efficacy: Whether or not the sUAS is able to perch/land on the platform without failure.
- Completion: The number of successful trials divided by the total number attempted, reported as a percentage. Each number of successful trials represents an associated probability of success and confidence level, for example:
    - 10 successful trials with no failures for 85% probability of success with 80% confidence
    - 5 successful trials with no failures for 70% probability of success with 80% confidence

- 10 successful trials with 1 failure for 75% probability of success with 85% confidence
- Etc. (see Leber et al. [2019] for more details on the statistics associated with number of successful trials and acceptable failures)

- <u>OCU notification</u>: Whether or not the OCU interface provided a notification indicating issues with non-flat flooring, nearby obstructions, etc., in the event that the sUAS is not able to takeoff from or perch/land on the platform, which could be in the form of a text alert, flashing screen warning, audio alert, etc.
- <u>Collisions</u>: Whether or not the sUAS collided with the apparatus (e.g., overhead or lateral obstructions). Note: landing on the platform does not constitute a collision.
- <u>Rollovers</u>: Whether or not the sUAS rolled over while attempting to takeoff or land/perch on the platform.

## Procedure

1. Set up the apparatus for the desired condition to be tested.

<u>Takeoff test</u>:

2. Position the sUAS in the center of the platform and orient appropriately.
3. Attempt to launch the sUAS and navigate forward at least 3 m (10 ft); see Figure 2.
4. If successful, either land the sUAS or turn the sUAS around while hovering to attempt the perch/land test, starting at step 4 of that procedure (see below).
5. Repeat steps 2-4 until the desired number of successful trials has been achieved.
6. Calculate the metrics.

<u>Land/perch test</u>:

2. Position the sUAS in an open area where it is safe to takeoff that is at least 3 m (10 ft) away from the platform in the opposite direction of the sUAS starting orientation for the takeoff test; see Figure 2.
3. Launch the sUAS.
4. Attempt to navigate toward the platform and land on the platform.
5. If successful, either position the sUAS at the starting position from step 2 or attempt the takeoff test, starting at step 2 of that procedure (see above).
6. Repeat steps 3-5 until the desired number of successful trials has been achieved.
7. Calculate the metrics.

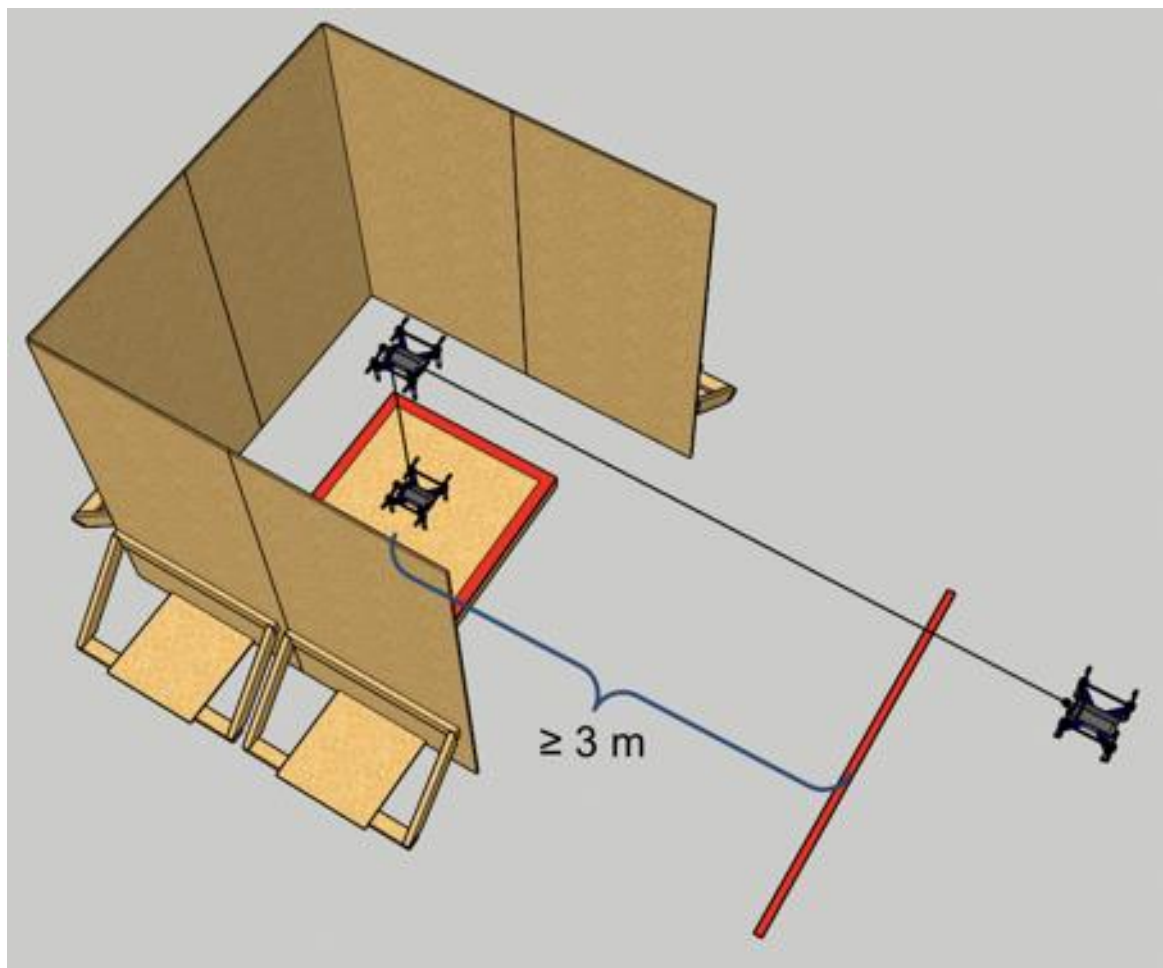

*Figure 2. Rendering condition 8 showing the minimum flight distance to and from the platform.*

## Example Data

- Environment characterization: Lighted
- Performance data:

**Takeoff**

| sUAS | Metrics | Condition | | | | | | | | |
|---|---|---|---|---|---|---|---|---|---|---|
| | | **Flat** | **5° roll** | **5° pitch** | **10° roll** | **10° pitch** | **1.2 m overhead** | **2.4 m overhead** | **1.2 m lateral** | **2.4 m lateral** |
| sUAS A | Completion | 100% | 100% | 80% | 80% | 90% | 0% | 100% | 100% | 100% |
| | Collisions | 0 | 0 | 2 | 2 | 1 | 5 | 0 | 0 | 0 |
| | Rollovers | 0 | 0 | 0 | 1 | 0 | 0 | 0 | 0 | 0 |
| sUAS B | Completion | 100% | 100% | 100% | 100% | 100% | 100% | 100% | 100% | 100% |
| | Collisions | 0 | 0 | 0 | 0 | 0 | 0 | 0 | 0 | 0 |
| | Rollovers | 0 | 0 | 0 | 0 | 0 | 0 | 0 | 0 | 0 |
| sUAS C | Completion | 100% | 100% | 100% | 100% | 100% | 0% | 100% | 0% | 100% |
| | Collisions | 0 | 0 | 0 | 0 | 0 | 5 | 0 | 5 | 0 |
| | Rollovers | 0 | 0 | 0 | 0 | 0 | 0 | 0 | 0 | 0 |

**Land/perch**

| sUAS | Metrics | Condition | | | | | | | | |
|---|---|---|---|---|---|---|---|---|---|---|
| | | **Flat** | **5° roll** | **5° pitch** | **10° roll** | **10° pitch** | **1.2 m overhead** | **2.4 m overhead** | **1.2 m lateral** | **2.4 m lateral** |
| sUAS A | Completion | 90% | 90% | 63% | 88% | 90% | 100% | 100% | 100% | 100% |
| | Collisions | 1 | 1 | 3 | 2 | 1 | 0 | 0 | 0 | 0 |
| | Rollovers | 0 | 0 | 1 | 1 | 0 | 0 | 0 | 0 | 0 |
| sUAS B | Completion | 100% | 100% | 100% | 100% | 100% | 100% | 100% | 100% | 100% |
| | Collisions | 0 | 0 | 0 | 0 | 0 | 0 | 0 | 0 | 0 |
| | Rollovers | 0 | 0 | 0 | 0 | 0 | 0 | 0 | 0 | 0 |
| sUAS C | Completion | 100% | 100% | 100% | 100% | 0% | 100% | 100% | 0% | 100% |
| | Collisions | 0 | 0 | 0 | 0 | 5 | 0 | 0 | 5 | 0 |
| | Rollovers | 0 | 0 | 0 | 0 | 0 | 0 | 0 | 0 | 0 |

# Room Clearing

## Purpose

This test method provides a standard profile for clearing a room, meaning to visually inspect it in order to evaluate the contents of the room for mission planning purposes, such as the presence of people, objects of interests, environment hazards, and ingress/egress points.

## Summary of Test Method

While the environments in a mission-context where room clearing can be performed will vary in terms of room dimensions and types of obstructions in the room, a standard room is specified for this test method to be representative of a room clearing task. The nominally-sized room has a series of visual acuity targets on all surfaces and is without obstructions to provide a clear view of all targets for inspection. A future variant of this test method may be developed that includes one or more sets of standard obstruction layouts. The sUAS can takeoff inside of the room or enter from outside, whichever is preferred; either way, the actual test will not begin until after the sUAS is hovering in the center of the room. From there, the sUAS performs a visual inspection of the room. While not required, it is recommended that the sUAS remain in the center of the room and manipulate its gimbal camera to increase vertical field of view (FOV) for inspecting the floor and ceiling, while yawing in place.

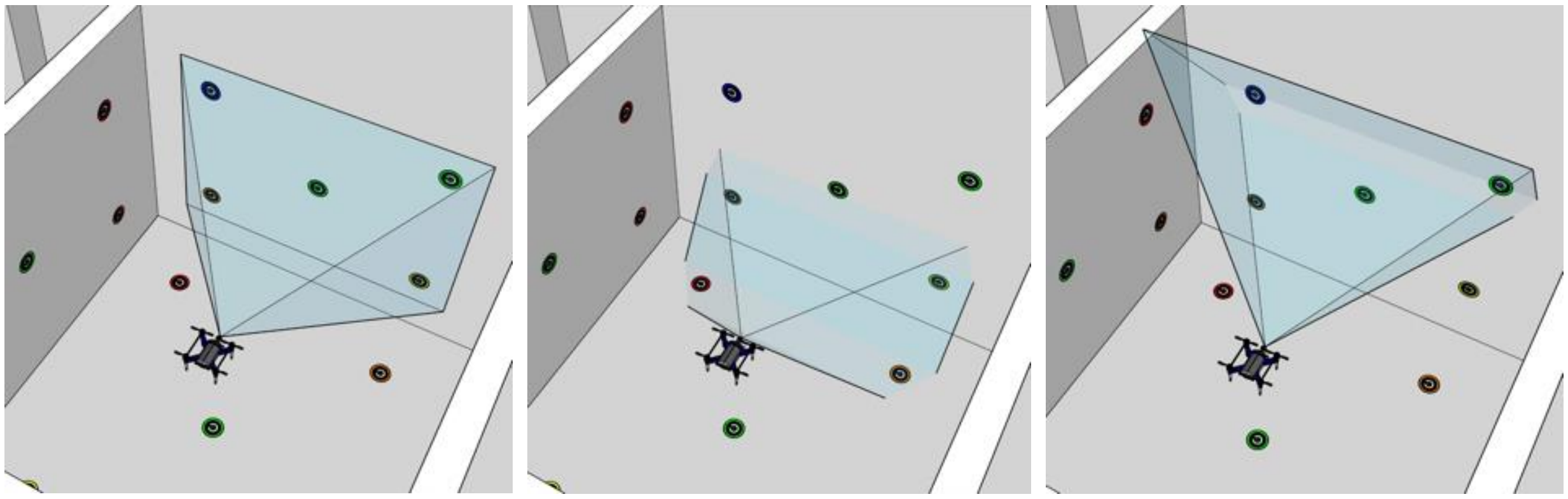

*Figure 1. Rendering of example sUAS field of view when inspecting a wall, floor, and ceiling by manipulating the gimbal camera. Note: ceiling is not shown, but is present during testing.*

The sUAS gimbal movement range and FOV will impact the number of visual acuity targets that can be inspected; e.g., some sUAS will not be able to see the targets on the floor or ceiling due to lack of gimbal capability, while others may be able to see multiple surfaces at once through the use of 360 degree cameras. Additionally, the control and stabilization of the sUAS is exercised by attempting to yaw in place to scan the room. The sUAS is free to move through the room as needed (e.g., navigate forward, back, ascend, descend, etc.), although the room is intentionally narrow to influence a more expedient scanning technique of yawing in place. Two variants of room clearing capability are exercised:

- Static camera: Without the use of camera zoom functionality, likely resulting in faster, coarser room clearing at reduced visual acuity.
- Zoom camera: Allowing for the use of camera zoom functionality if available, likely resulting in slower, finer room clearing at increased visual acuity.

During the test, the operator inspects the visual acuity targets and the test lasts until all visual acuity targets able to be inspected (i.e., those that the sUAS has the capability of inspecting; some targets may not be able to be inspected due to limitations in sUAS gimbal movement), have been successfully inspected.

Video from the sUAS cameras may be higher quality when viewed in post on a high-resolution display monitor rather than through the OCU display screen. To this end, two additional variants of evaluation can be conducted:

- In-situ evaluation: The operator inspects each visual acuity target and calls out the direction of the Landolt Cs while the sUAS is in flight. This is similar to conducting room clearing in real time.

- Post-hoc evaluation: The video of the sUAS cameras is exported and evaluated after the fact, whereby the administrator inspects each visual acuity target and determines the level of acuity able to be achieved in each target. This is similar to conducting room clearing or a mapping exercise after a mission has been conducted.

Lighting in the room is characterized as lighted (100 lux or greater) or dark (less than 1 lux). Ideally, the sUAS will not collide with the boundaries of the room (i.e., walls, floor, and ceiling surfaces), but contact is allowed so long as it does not cause the sUAS to crash in a way that requires human intervention for it to resume flight.

Room clearing be run either as an elemental or operational test:

Elemental Room Clearing: The operator may maintain line-of-sight with the sUAS such as by following the system with the OCU and standing in the doorway to maintain communications link, allowing for room clearing to be evaluated in as close to an ideal setting as possible and reduce potential collisions with the boundaries.

Operational Room Clearing: The operator is positioned away from the room with their back to the doorway, unable to maintain line of sight throughout the test. This is similar to an actual operational mission, including all related situation awareness issues that may arise (e.g., collisions with the boundaries, misunderstanding which wall is being inspected, etc.).

## Apparatus and Artifacts

A room free of lateral obstructions is used with dimensions for width (W), length (L), and height (H) that are each between 2.4 m (8 ft) and 3.7 m (12 ft), such that it can be run in a typical office setting, shipping container, or a fabricated environment for testing purposes. The actual dimensions of the room are reported along with the performance data. Overhead obstructions such as lighting fixtures or HVAC ducts may still be present in a given room; the lowest point of these obstructions should be considered the room's ceiling such that the H dimension spans the floor and this point. One wall will contain an open doorway; any doors or windows on other walls will be closed for the duration of the test.

A set of five visual acuity targets are positioned on all surfaces to form a 5-dice pattern (the wall with the open doorway will only have three visual acuity targets) for a total of 28 targets in the room to be inspected. The visual acuity targets on the ceiling should be mounted such that they are positioned H distance above the floor and just below any overhead obstructions (e.g., lighting fixtures). Each visual acuity target is identified by an alphanumeric code according to the surface it is positioned on (wall front = WF, wall left = WL, wall back = WR, wall right = WR, floor = F, ceiling = C) and a number 1-5 (center = 1, clockwise from top left = 2-5); e.g., WL-2, F-3, C-5, etc. The center of the 5-dice pattern of visual acuity targets (i.e., target 1) is positioned in the center of the surface (e.g., 0.5W by 0.5L on the floor, 0.5W by 0.5H on the wall, etc.) and the corners of the 5-dice pattern are positioned 60 $cm^2$ away from the boundary corners adjacent to other surface boundaries (see Figure 2).

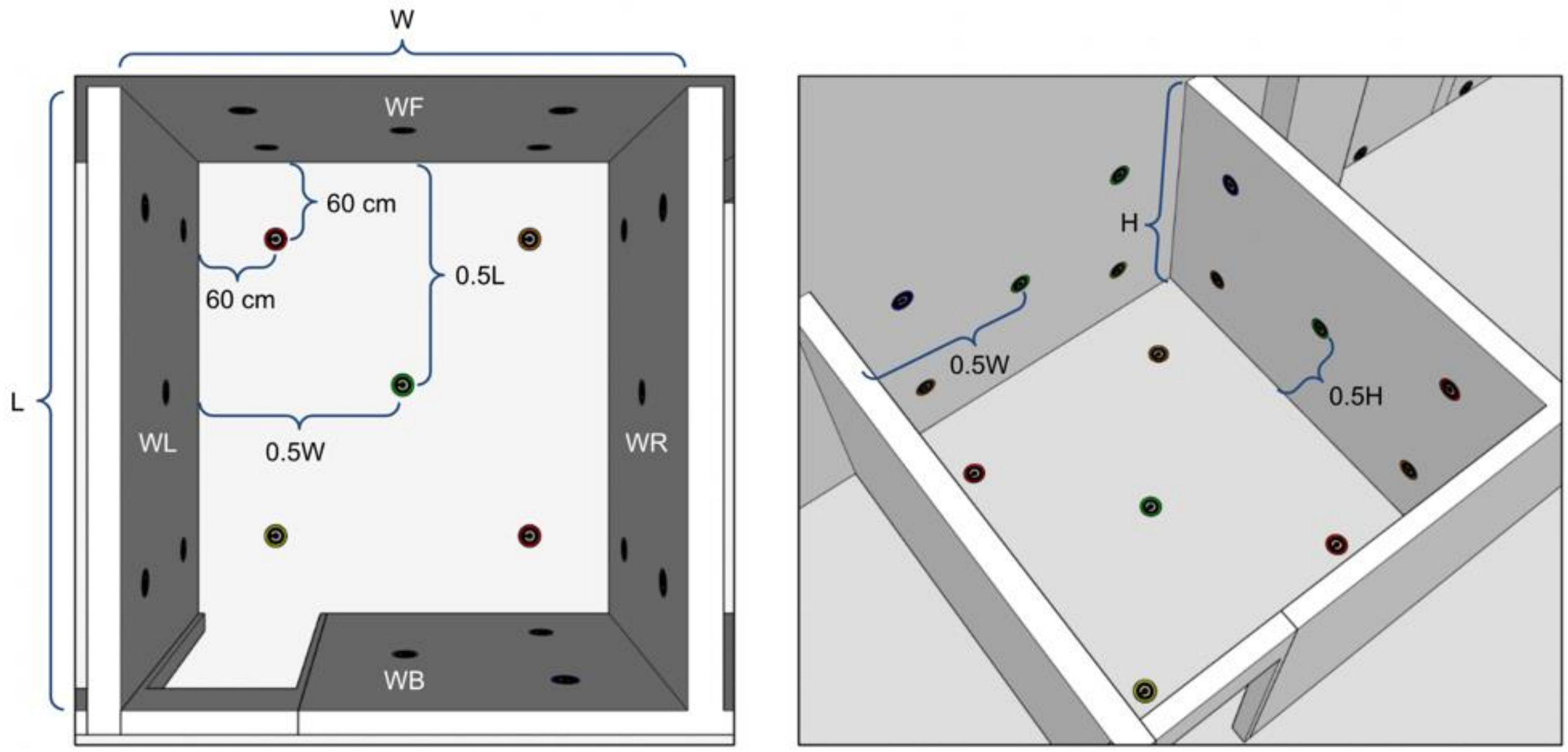

*Figure 2. Placement of visual acuity targets on the walls and floor. While not shown, the ceiling layout matches the floor.*

## Equipment

A timer is used to record the duration metric.

## Metrics

- Duration: The amount of time to perform the test, starting from when the administrator starts the timer until the last remaining visual acuity target that can be inspected by the sUAS has been inspected, reported in minutes.
- Visual acuity: Level of detail that can be resolved in the available Landolt C artifacts during flight, reported per visual acuity target able to be seen. These measures are averaged to calculate an average acuity (with standard deviation) for each surface type (walls, floor, ceiling) and across the entire test.
- Coverage: Number of visual acuity targets able to be seen per surface type (walls, floor, ceiling) and across the entire test, reported as a percentage out of the total number of visual acuity targets available on each surface type (walls = 18, floor = 5, ceiling = 5) and in the entire room (28).
- Collisions: Whether or not the sUAS collided with the apparatus boundary surfaces (walls, floor, or ceiling), reported per surface type and across the entire test.

## Procedure

1. Position the sUAS centered inside of the room. This can be accomplished by either commanding the sUAS to takeoff while inside of the room or doing so outside of the room and navigating in through the doorway.
2. Orient the sUAS such that it is facing either the left wall (for clockwise scanning) or right wall (for counterclockwise scanning) and adjust the elevation of the sUAS to approximately 0.5H such that target WL-1 or WR-1 is level with the sUAS camera pointed forwarded.
3. Starting with the static camera variant (i.e., the operator is not allowed to use the camera zoom functionality of the sUAS, if available), the administrator announces the start of the test to the operator and starts the timer.
4. The operator can choose to follow one of two room clearing profiles:
    a. Recommended room clearing profile: The operator commands the sUAS to yaw in place in order to inspect all visual acuity targets that can be seen on the walls, then gimballing the sUAS camera down and yawing in place to inspect all visual acuity targets that can be seen on the floor, and

then gimballing the sUAS camera up and yawing in place to inspect all visual acuity targets that can be seen on the ceiling.

   b. Custom room clearing profile: The operator commands the sUAS as they deem fit in order to efficiently inspect the visual acuity targets on the walls, floor, and ceiling.

5. If in-situ evaluation is being conducted:
   a. The operator announces the orientations of the Landolt C artifacts down to the smallest size able to be resolved to the administrator for each visual acuity target inspection.
   b. Once all targets able to be inspected have been successfully inspected, the timer is stopped and the test is over.
6. If post-hoc evaluation is being conducted:
   a. Once all walls, floor, and ceiling have been scanned, the timer is stopped, the sUAS lands, and its video is exported.
   b. The administrator watches the recorded video and evaluates each visual acuity target.
7. If the sUAS has zoom camera functionality, then repeat steps 3-6 again, but now the operator is able to zoom the camera during inspection in order to increase visual acuity measures.
8. Calculate the metrics.

## Example Data

- Environment characterization: Lighted, W = 3.2 m, L = 3.5 m, H = 2.4 m
- Performance data: Operational room clearing

| sUAS | Metrics | In-situ Static cam | In-situ Zoom cam | Post-hoc Static cam | Post-hoc Zoom cam |
|---|---|---|---|---|---|
| sUAS A | Duration (min) | 5.2 | 5.2 | 3.1 | 3.1 |
| | Coverage | 83% | 83% | 83% | 83% |
| | Average acuity (mm) | 7.8 | 7.8 | 4.0 | 4.0 |
| sUAS B | Duration (min) | 4.6 | 12.1 | 2.1 | 7.8 |
| | Coverage | 100% | 100% | 100% | 100% |
| | Average acuity (mm) | 3.0 | 1.3 | 2.9 | 1.3 |

# Indoor Noise Level

## Purpose

This test method measures the amount of noise generated by sUAS when operating indoors.

## Summary of Test Method

Evaluating the noise generated by each sUAS is measured simply by using a decibel meter to record the volume in comparison to a baseline (i.e., the environment without the sUAS operating) when operating indoors. The relevant environment characteristics of each environment are recorded (e.g., indoor room dimensions, obstructions between the decibel meter and sUAS) in order to contextualize the results. The sUAS performs basic tasks (takeoff, hovering in place, ascending in place) in each environment while the decibel meter is placed at three distances: 2.5 m (8.2 ft), 5 m (16.4 ft), and 10 m (32.8 ft). Average decibel levels are reported for each condition.

## Apparatus and Artifacts

Real-world indoor environments deemed relevant for the use case can be used. There should be sufficient space that allows for 10 m distance between the sUAS and the decibel meter. It is preferable that the sUAS and decibel meter be placed within line-of-sight when positioned 10 m away. However, if this is not possible and obstructions will exist between, the number and type of obstructions should be recorded (e.g., 1 drywall).

## Equipment

A decibel meter is used to record the noise level. It is recommended it be placed on a tripod during testing.

## Metrics

- Noise level: The average level of noise generated in the environment (ambient noise level) and by the system (sUAS noise level), reported in decibels (dB).

## Procedure

1. Record the ambient noise level of the environment.
2. Position the sUAS and decibel meter at the prescribed distances.
3. Power on the sUAS and OCU and establish connection.
4. Begin continuously recording noise level measurements.
5. Command the sUAS to takeoff, hover in place, and ascend in place.
6. Determine average noise level.
7. Repeat steps 1-6 for each distance.

## Example Data

- Environment characterization:
    - Indoors: 8 x 6 x 5 m room, drywall walls, concrete floor, ambient noise level: 30 dB
- Performance data:

| **sUAS** | **2.5 m Hover** | **2.5 m Ascension** | **5 m Hover** | **5 m Ascension** | **10 m Hover** | **10 m Ascension** |
|---|---|---|---|---|---|---|
| sUAS A | 90 | 92.9 | 88.2 | 91 | 82.2 | 84.6 |
| sUAS B | 54.5 | 55.8 | 53.5 | 54 | 49.5 | 51 |
| sUAS C | 103 | 106.1 | 102 | 105.7 | 93 | 101.9 |

# Logistics Characterization

## Purpose

This practice characterizes the sUAS in terms of its physical properties, operation considerations, and maintenance requirements, allowing for side-by-side comparison of a set of candidate systems.

## Summary of Test Method

A series of characteristics concerning the logistics of operating, maintaining, and collecting data are outlined across seven categories that are to be filled with the relevant information for each sUAS platform being evaluated: physical measurements, power, heat dissipation, safety precautions, body/frame and maintenance, data collection and access, and system survivability. In each category, several fields are posed as prompts/questions for the user to respond to based on empirical evidence and experience of operating and maintaining the system. Some data may be able to be initially derived from vendor-provided specification sheets, but should be verified empirically. All fields are open response, although some may require a specific format of response (e.g., yes/no, dimension units, etc.). The information captured under each field is as follows:

- Physical measurements:
  - Deployed size, Collapsed size, Controller size, Table / Screen size, Dimensions of drone carry case, Weight of drone carry case w/o drone, Dimensions of controller carry case,
  - Weight of frame without battery, Battery weight, Weight of controller, Weight of controller carry case w/o drone, Dimensions of charger carry case, Weight of controller carry case w/o controller, Weight total
- Power:
  - Battery type, Battery charge time, Average flight with full battery, Controller battery type, Controller charge time, Can the power be switched on/off
  - Is battery level displayed to user, Is battery remaining time indicated, Is flight time remaining indicated, Is the user prompted about damaged battery, Are actions prevented at critical battery levels, Are there failsafe lockouts, Can user override failsafe lockouts, What behavior is exhibited at critical power Is battery connection easily accessible
- Heat dissipation and consideration:
  - Operation Temp range, Can the drone idle without overheating, Does it have internal or passive cooling, Is the user prompted on critical heat levels, What happens on overheat, Do batteries need to cool down after use
- Safety precautions:
  - Does operation require hearing protection, Does operation require eye protection, Does operation require head protection, Does operation suggest a respirator
- Frame and maintenance:
  - Are parts serviceable, Are parts reinforced or weatherized, Are parts custom made or off shelf, Is drone 3d printed or mass produced, Are parts interchangeable, Is drone stored fully assembled, Are propellers protected, Can prop guards be attached, Are tools provided with drone
- Data collection and access:
  - Does the drone start recording when armed, Is data stored onboard or in controller, How is data accessed, What format is data stored in, Is software required to process data
- System survivability:
  - Visual detectability, Audible signature, Cybersecurity/encryption
- Capabilities:
  - Does the drone have obstacle avoidance, Does the drone prevent hitting objects, Can obstacle avoidance be disabled, Does the drone have auto takeoff, Does the drone have an auto land, Does the drone have an emergency stop, Is the drone able to carry a payload

Across a set of sUAS platforms, their logistics characteristics can be compared. Additionally, criteria can be set for each characteristic to determine if a system meets the relevant requirements set forth by another entity. For example, soldier feedback provided to the Soldier-Borne Sensor (SBS) program indicated that a minimum of 2 hours of HD video be able to be recorded. This threshold of acceptable performance can be compared to the information provided for candidate sUAS in the *data collection & access* category. For a coarse representation of requirement matching, a percentage can be calculated of the number of fields that match a given set of criteria.

## Apparatus and Artifacts

Deriving responses to some fields based on empirical evidence may require the use of a test method that utilizes its own apparatuses and artifacts, in which case the specifications for that test method should be followed. For example, providing information under the *power* category requires information on the average flight time, which can be derived from the Runtime Endurance test method; that test method requires its own set of apparatuses.

## Equipment

A scale is needed to weigh the sUAS and its components (e.g., batteries, controller, case, etc.) and a tape measure is needed for recording dimensions. A camera can be used to provide supporting photo documentation for certain responses; e.g., responses under the *body/frame and maintenance* category concerning how the sUAS is stored may include a photo of the storage configuration.

## Metrics

- Requirements met: Each field can be evaluated against a provided set of responses that define criteria that the sUAS must meet in order to fulfill a requirement set by an external entity. This metric can be evaluated per field or more coarsely as a percentage by summing the total number of fields that match the provided criteria divided by the total number of fields for which criteria is provided.

## Procedure

1. Review each category and field to be characterized.
2. Respond to each prompt/question for a given sUAS, referring to empirical evidence and/or experience.
3. Compare across several sUAS and/or to a set of criteria to calculate the requirements met metric.

## Example Data

| sUAS | | |
|---|---|---|
| Make | sUAS Make A | sUAS Make B |
| Model | sUAS Model A | sUAS Model B |
| **Physical measurements** | | |
| Deployed size | 15.5 x 15.5 x 14.5 | 20 x 17 x 3.5 |
| Collapsed size | N/A | 8.5 x 3.75 x 3.5 |
| Weight of frame without battery lbs | 2 | 1.5 |
| Battery weight lbs | 1 | 0.5 |
| Weight of controller lbs | 4.5 | 4.5 |
| Controller size | 8.25 x 8.25 x 5 | 10.2 x 5.25 x 2.31 |
| Tablet / Screen size | 9 x 8.5 x 3 | 6 x 3.5 |
| Dimensions of drone carry case | 22.75 x 16.75 x 19 | 18 x 21 x 9 |
| Weight of drone carry case w/o drone lbs | 40 | 10.5 |
| Dimensions of controller carry case | N/A | N/A |
| Weight of controller carry case w/o drone | N/A | N/A |
| Dimensions of charger carry case | N/A | 17 x 13 x 10 |
| Weight of controller carry case w/o controller | N/A | 10.5 |
| Weight total | 47.5 | 27.5 |
| **Power** | | |
| Battery type | Li-po | Li-ion |
| Battery charge time | 1 Hour | 1 hour |
| Average flight with full battery | 9 min | 10 min |
| Controller battery type | Li-ion | Li-ion |
| Controller charge time | 1 Hour | 1 Hour |
| Can the power be switched on/off | No | Yes |
| Are there failsafe lockouts | Yes | Yes |
| Can user override failsafe lockouts | Yes | Yes |
| What behavior is exhibited at critical power | Auto land | fails to maintain altitude |
| Is battery connection easily accessible | No | Yes |
| **Heat dissipation and consideration** | | |
| Operation Temp range | 50 to 86 F | Not Defined |
| Can the drone idle without overheating | No | Yes |
| Does it have internal or passive cooling | Passive | Passive |
| What happens on overheat | Motor and battery damage | Drone disconnects from radio |
| Do batteries need to cool down after use | Yes | Yes |
| **Safety precautions** | | |
| Does the drone have an emergency stop | Yes | Yes |
| Does operation require hearing protection when operator is nearby | Yes | No |
| Does operation require eye protection when operator is nearby | Yes | Yes |
| Does operation require head protection when operator is nearby | Recommended | Recommended |
| **Frame and maintenance** | | |
| Are parts serviceable | Yes | No |
| Are parts reinforced or weatherized | No | No |
| Are parts custom made or off shelf | Custom | Custom |
| Is drone 3D printed or mass produced | 3d printed parts | 3d printed parts |
| Are parts interchangeable | Yes | No |
| Is drone stored fully assembled | Yes | No, props must be removed |
| Are propellers protected | Yes | No |
| Can prop guards be attached | Not needed | No (?) |
| Are tools provided with drone | Yes | Yes |
| Is the drone able to exert force on an object without crashing | Yes | No |
| **Data collection and access** | | |
| Does the drone start recording when armed | Yes | Yes |
| Is data stored onboard or in controller | Both | Onboard |
| How is data accessed | SD card removal or direct connection to drone | SD card removal, tools required |
| What format is data stored in | .mov , .thm, .log | .ts |
| Is software required to process data | Yes | No |
| **System survivability** | | |
| Visual detectability - at what range does the drone become easily visible | The drone is easily detected due to its shape, size, lights, and the noise level of its propellers when in use | The drone would be hard to detect at medium to long range when at high elevation |
| Audible signature- at what range is the drone detectable | Long | Medium |
| Cybersecurity - is the drone encrypted | No | No |
| **Capabilities** | | |
| Does the drone have obstacle avoidance | No | Yes, front and side avoidance |
| Does the drone prevent hitting objects | No | Yes |
| Can obstacle avoidance be disabled | N/A | Yes |
| Does the drone have auto takeoff | | |
| Does the drone have an auto land | Yes | No |
| Does the drone have an emergency stop | Yes | Yes |
| Is the drone able to carry a payload | Yes | Yes |

# Interface

The operator control unit (OCU) for sUAS consists of a series of input modalities (e.g., buttons, touch screen, joysticks) and a display screen for outputting video feeds, thermal camera feeds, and other interface elements (e.g., battery level indicator, communications strength). This category of test methods seeks to characterize the various components of the sUAS interface, including physical characteristics, interface functionality, and nuances in operation that may impact field performance by potentially confusing operators (e.g., changing controller functionality when changing flight modes).

## Operator Control Unit (OCU) Characterization

### Purpose

This practice characterizes the interface shown on the OCU of the sUAS in terms of the properties of input commands by the operator and the modalities used to output information on the display screen, allowing for side-by-side comparison of a set of candidate systems.

### Summary of Test Method

A series of characteristics concerning the OCU, its input functionalities through the controller, and the output provided via display modalities on the interface are outlined across five categories that are to be filled with the relevant information for each sUAS platform being evaluated: controller and UI, power, navigation, camera, and additional functionality and accessories. In each category, several fields are posed as prompts/questions for the user to respond to based on empirical evidence and experience of operating and maintaining the system. Some data may be able to be initially derived from vendor-provided specification sheets, but should be verified empirically. All fields are open response, although some may require a specific format of response (e.g., yes/no, dimension units, etc.). The information captured under each field is as follows:

- Controller and UI:
    - Is the controller labeled, How many non-virtual buttons are there
    - Do flight modes change the configuration of how functionality is mapped to the controller inputs
    - How is the user alerted to critical states, Is flight information fused with nav page, Do settings reset on power cycle
    - Does the drone have obstacle avoidance, Does the drone prevent hitting objects, Can obstacle avoidance be disabled, Are obstacle avoidance notifications shown
    - Does the drone have an auto land, Does the drone have an emergency stop, Are some features disabled in specific modes, Controller display lighting
- Power:
    - Is battery remaining time indicated, Is flight time remaining indicated, Is the user prompted about damaged battery, Are actions prevented at critical battery levels
- Communications link:
    - Does the interface display the current comms link connection level Is the user prompted about reduced comms link Is the user prompted about loss of comms link What happens on comms loss Does the drone alert the user to magnetic interference
- Navigation:
    - What type of navigation system is used, Is the drone GPS capable, Is mapping data displayed during flight, Is the drone GPS Denied compatible, Does its behavior change without GPS
    - Is touch screen used during flight
    - Maximum wind resistance, Do flight modes change wind resistance
    - Do flight modes change obstacle avoidance, Does the drone switch modes automatically, Are critical environmental conditions alerted, Is the drone able to hover in place w/o input
- Camera:
    - Are cameras usable when not armed, Is there a thermal camera
    - Is information fused on nav footage, Is nav cam always visible in menus in flight
    - Is zoom digital or physical

- <u>Additional functionality and accessories</u>:
    - Does the drone have illuminators, Does the drone have IR sensors / emitters, Does the drone have a laser or pointer, Is the drone able to carry a payload

Across a set of sUAS platforms, their OCU characteristics can be compared. Additionally, criteria can be set for each characteristic to determine if a system meets the relevant requirements set forth by another entity. For example, soldier feedback provided to the Soldier-Borne Sensor (SBS) program indicated that, ideally, a system's OCU functionality should not change when operating with GPS or when GPS-denied. This threshold of acceptable performance can be compared to the information provided for candidate sUAS in the *navigation* category. For a coarse representation of requirement matching, a percentage can be calculated of the number of fields that match a given set of criteria.

### Apparatus and Artifacts

Deriving responses to some fields based on empirical evidence may require the use of a test method that utilizes its own apparatuses and artifacts, in which case the specifications for that test method should be followed. For example, the providing information under the *controller and UI* category requires information related to alerting the operator when the sUAS is low on battery, which can be observed during the Runtime Endurance test method once the system's battery is close to expiring; that test method requires its own set of apparatuses.

### Equipment

A camera can be used to provide supporting photo documentation for certain responses; e.g., responses under the *controller and UI* category concerning the various interface modalities and icons available on the interface may include a photo of the OCU that shows the relevant controller or UI feature.

### Metrics

- <u>Requirements met</u>: Each field can be evaluated against a provided set of responses that define criteria that the sUAS must meet in order to fulfill a requirement set by an external entity. This metric can be evaluated per field or more coarsely as a percentage by summing the total number of fields that match the provided criteria divided by the total number of fields for which criteria is provided.

### Procedure

1. Review each category and field to be characterized.
2. Respond to each prompt/question for a given sUAS, referring to empirical evidence and/or experience.
3. Compare across several sUAS and/or to a set of criteria to calculate the requirements met metric.

## Example Data

| sUAS info | | |
|---|---|---|
| Make | sUAS Make A | sUAS Make B |
| Model | sUAS Model A | sUAS Model B |
| **Controller and UI** | | |
| Is the controller labeled | Partially | No |
| Is there an onboard manual / control guide | No | No |
| How many physical buttons are there | 16 | 12 |
| Does the controller use a touch screen? | Yes | Yes |
| Does the controller use a touch screen to start flight | Yes | Yes |
| Is use of the touch screen required during flight | No | No |
| Is the touch screen easily responsive | Yes | No |
| Do flight modes change the configuration of how functionality is mapped to the controller inputs | Yes | No |
| How is the user alerted to critical states | Push prompts on screen and color changes of the screen border | QGC command line, but not visible during flight |
| Do settings reset on power cycle | Yes | Yes |
| Are obstacle avoidance notifications shown | N/A | No |
| If the drone has an auto-land, how/when is it engaged | Yes, but only at critical battery | No, but can detect landing if on ground and throttle is pulled down |
| If the drone has auto takeoff, is it a physical or a virtual button? | N/A | N/A |
| If the drone has an emergency stop that can be initiated by the user, is it a physical or a virtual button? | Physical, pattern on control sticks | Physical, arming button is also used to Estop and disarm |
| Are some features disabled in specific modes | Yes | Yes, OA turns off at higher speeds |
| Controller display lighting | Yes | Yes |
| **Power** | | |
| Is battery remaining time indicated | Yes, % | Yes, % |
| Is flight time remaining indicated | Yes, timer | No |
| Is the user prompted about damaged battery | Yes, batteries track flights, usage, and expiration | No |
| Are actions prevented at critical battery levels | Yes | No |
| **Communications link** | | |
| Does the interface display the current comms link connection level | Yes, bar | No |
| Is the user prompted about reduced comms link | Yes, push notification about strong or weak signal | No |
| Is the user prompted about loss of comms link | Yes, will say communications loss and show UI but blank, loads back to splash page | Yes, after reconnect the controller supplies a popup saying comms lost |
| What happens on comms loss | Auto land on full coms loss, idles props to cool system | Lands and disarms |
| Does the drone alert the user to magnetic interference | No* | No |
| **Navigation** | | |
| Is 3D mapping data displayed during flight | No | No |
| Does its behavior change without GPS | No | Yes, harder to hold position due to drift |
| Is the user prompted about loss of GPS | N/A | Yes, UI GPS indicator on side of screen |
| Is touch screen used during flight | No, after initialization, unless changing settings | No |
| Do flight modes change obstacle avoidance | No | Yes, at different speeds the drone will increase its avoidance distance but turn it off at highest speed |
| Does the drone switch modes automatically | No, user requested | No, user requested |
| Are critical environmental conditions alerted | Yes, popup notification | No |
| **Camera** | | |
| Are cameras usable when not armed | Yes | Yes |
| Do the navigation camera streams also display UI elements (e.g., map, system status) | Yes | No |
| Is nav cam always visible in menus during flight | Yes | No |
| can the camera zoom, is it digital or optical | Digital | Digital |
| Are there multiple cameras used for varied zoom levels | No | No |
| Expected Horizontal Field of View of main camera model Degrees | 114 | Undefined in documentation |
| Expected Horizontal Field of View of thermal camera model Degrees | 56 | Undefined in documentation |
| Horizontal Field of View of main nav camera Degrees | 123 | 102 |
| Vertical Field of View of main nav camera Degrees | 77 | 74 |
| Gimbal Range Up Degrees | 90 | 75 |
| Gimbal Range Down Degrees | 90 | 75 |
| Field of Regard of main nav camera Up Degrees | 128 | 112 |
| Field of Regard of main nav camera Down Degrees | 128 | 112 |
| Thermal Horizontal Field of View Degrees | 60 | 40 |
| Thermal Vertical Field of View Degrees | 43 | 35 |
| Field of Regard of Thermal camera UP Degrees | 122 | 92 |
| Field of Regard of Thermal camera Down Degrees | 122 | 92 |
| **Additional functionality and accessories** | | |
| Does the drone have illuminators | Yes | Yes |
| Does the drone have a laser or pointer | No | Yes |

# Obstacle Avoidance

In subterranean and constrained indoor environments, sUAS face operational risks due to potential collisions with boundaries (i.e., walls, floor, ceiling) in confined space, but also with additional obstacles positioned in the environment that induce even further confinement. This category of obstacle avoidance and collision resilience test methods is intended to evaluate sUAS capability to persist in these environments either through the use of autonomous functionality for obstacle avoidance, or through other means of resilience such as possessing a protective cage or prop guards.

## Obstacle Avoidance and Collision Resilience

### Purpose

The purpose of this test method is to validate the ability of each sUAS to detect, avoid, and/or be resilient to collisions with a predefined set of obstacles common in subterranean or indoor environments.

### Summary of Test Method

The test method consists of flying an sUAS directly towards different types of obstacles and recording the response. The test method is categorized into two classes based on the fundamental capabilities of the sUAS being evaluated. Such systems can usually be classified as having (a) an active collision avoidance system based on a full (or shared) autonomy mode, or (b) a passive collision resilience system (e.g., propeller guards or cage) that limit the impact of collisions with obstacles when they do occur. The different classes of sUAS obstacle capabilities each require their own test methodology. sUAS without any obstacle avoidance or collision resilience capabilities cannot be tested using this method. Metrics such as stopping times and maximum deceleration experienced will be evaluated. The appropriate tests will be performed for three scenarios: (a) head-on-collision course with obstacle (i.e. flight direction is perpendicular to plane of obstacle), (b) collision path that is angled at 45-degrees from the plane of the obstacle, and (c) sideways collision with impact on sUAS starboard or portside. Tests will be performed for the following obstacle elements: walls, chain link fences, mesh materials, and doors.

<u>Obstacle Avoidance</u>: The obstacle avoidance test methodologies pertain to any sUAS systems that possess autonomous obstacle avoidance capabilities, i.e. the sUAS should be able to perceive the presence of an obstacle and take corrective actions to avoid collisions. This test methodology does not cover human-piloted sUAS. The test method evaluates various metrics such as minimum time to collision, minimum distance to collision, and number of collisions. The tests show not only if the system is able to avoid the obstacle but it also assesses sUAS performance for different materials. Some of the materials used in these tests (such as chain link fence and meshes) are significantly more difficult to perceive by sUAS systems as compared to others (such as doors and walls). The tests seek to assess the obstacle avoidance performance in these different scenarios.

<u>Collision Resilience</u>: Collision resilience test methodologies apply to all sUAS, including human-piloted and autonomous systems. These tests evaluate the ability of the sUAS to be resilient to collisions, by analyzing numerical metrics such as maximum deceleration experienced during a collision event, and pre-post collision change in velocity, as well as newly-devised categorical resilience metrics, as discussed in the Metrics section. These test methods are especially useful for analyzing the efficacy of sUAS platforms with additional protection such as propeller guards or cages. The tests with wall, chain link fence, and mesh, have similar methodology which includes flying towards the obstacle in various configurations (forward flight, sideways flight) and at various angles (trajectory is perpendicular to obstacle or at 45 degrees). The collision resilience tests for doors have slightly different setups which include flying towards a door obstacle that is closed, partially open, or open.

### Apparatus and Artifacts

Artifacts that mimic different kinds of obstacles that may be encountered in subterranean or indoor environments are needed. In addition to walls or vertical partitions to act as obstacles, the obstacle avoidance and collision resilience tests will be performed in the presence of plastic mesh, chain link fence, and flat wall obstacles. One 1.5 m x 1.5 m instance of each of these materials will be used. Door-like and wall-like obstacles may be constructed using representative materials such as drywall, wood, plywood etc. The door obstacle tests for avoidance/resilience will focus on closed, partially open (opened at 45-degree angle to adjacent wall partitions), and open doors. Door-like obstacles are designed to have minimum dimensions of 7 ft x 3 ft. Tests for wall-like obstacles can use artifacts with larger dimensions, or use existing, readily-available walls (as discussed below).

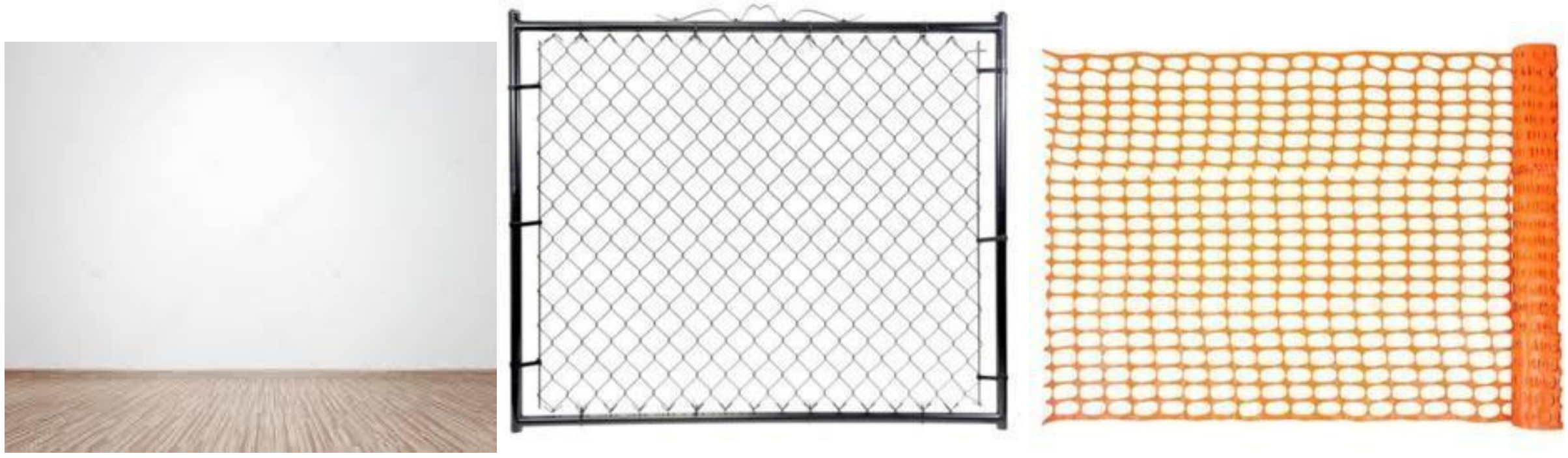

*Figure 1. Materials used for obstacle avoidance/resilience evaluation: flat wall (left), chain link fence (center), and mesh (right).*

In addition, the environment for conducting the evaluation should be bounded with walls to form a free space with dimensions of at least *10d x 15d*, where *d* represents the maximum horizontal dimension of the sUAS being tested (typically prop tip to opposing prop tip). The area should provide enough vertical space to safely takeoff, aviate, and land without fear of colliding with the floor or ceiling. This area will be further divided into a buffer zone (*1d* wide band around the outer edge), flight zone (*8d x 13d*; i.e., the remaining space outside of the buffer zone), and a testing zone within the flight zone (*7.5d x 10d*); see Figure 2. All provided dimensions are minimum measurements. The environment should be free from any obstructions (e.g., structural support columns) to enable safe flight and accurate sensing. Artifacts associated with collision resilience and obstacle avoidance tests should be positioned within the Testing Zone. If required, tests with wall-like obstacles can be conducted outside this zone (e.g. with walls that may bound the Buffer Zone.

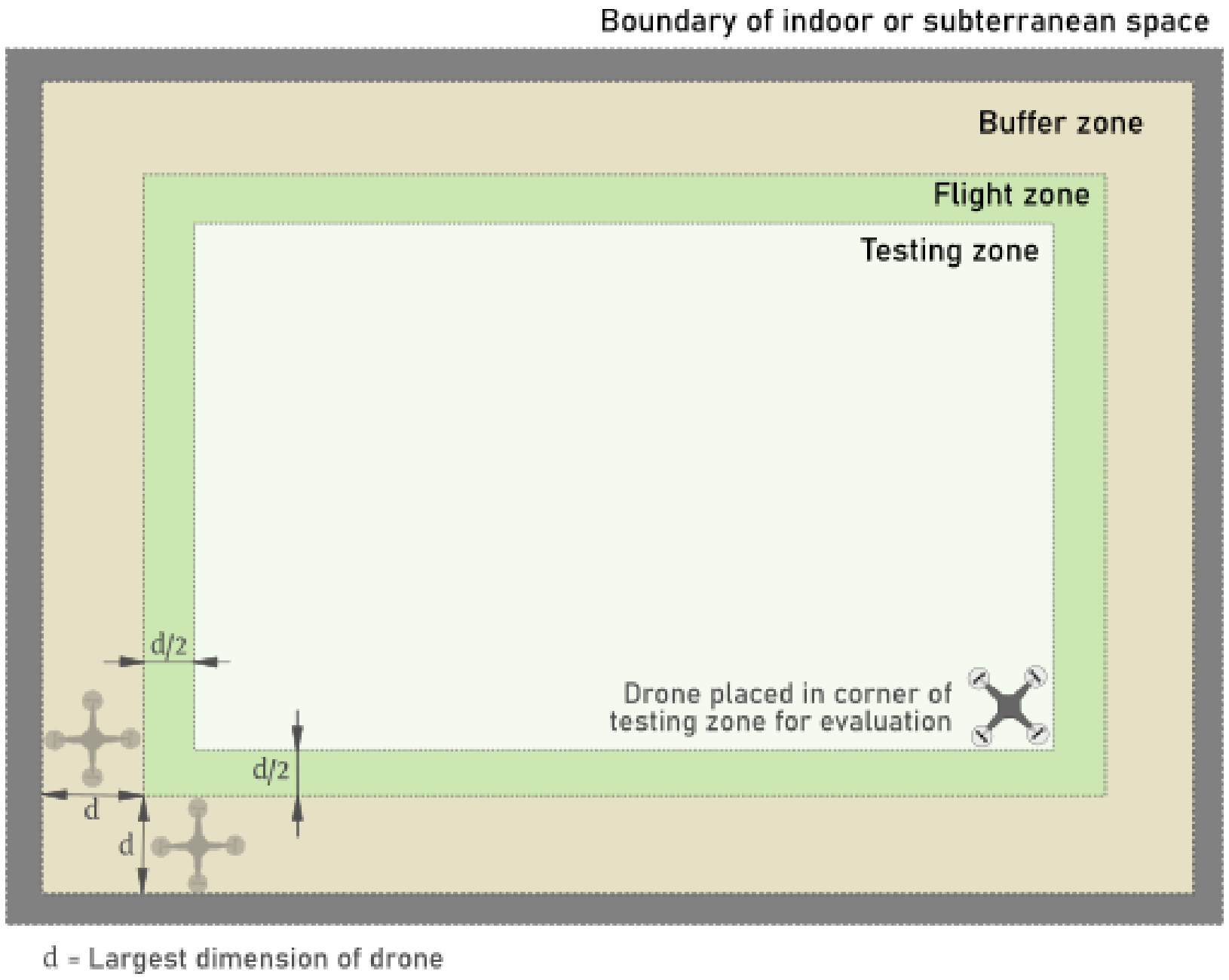

*Figure 2. Environment dimensional requirements.*

## Equipment

Data is collected to measure and evaluate the positional performance of the sUAS in each test. There are two primary options to collect the position information of the sUAS in order to determine its ability to successfully traverse the designated path: (a) internal tracking system existing on-board the sUAS platform, or (b) an external tracking system. External tracking requires hardware and software setup by the evaluators, which should be provided by the vendors of the specific tracking system used by the evaluators. For a given comparison of sUAS platforms using the test method, only one of the data collection methods should be used; i.e., the data collection methods should not be mixed and matched. Additional information about each data collection method is provided below.

Internal tracking system:

- This option requires access to sUAS telemetry, which depends on sUAS vendor assistance or permission. If access to sUAS telemetry is available, then the internal tracking system can be used to obtain sUAS pose and velocity data.
- It is possible that this data has been pre-processed and filtered by on-board systems. If only one sUAS is being evaluated, then such pre-processing does not interfere with the evaluation methods. However, if multiple sUAS are being tested and their performances compared, then it is required that data from the internal tracking systems on-board each sUAS should be available for the evaluation study. If on-board telemetry is not available or inaccessible for even one sUAS platform, then the evaluation and comparison study should either (a) use an external tracking system, or (b) remove the platform from the study. This is necessary in order to avoid inconsistencies in performance comparison between sUAS platforms.

External tracking system:

- In most situations, it may be preferable to deploy an external tracking system to evaluate performance. This approach may enable the evaluator to circumvent issues related to telemetry access and proprietary communication protocols, as well as reduce reliance on pre-processed telemetry streams which may differ from one sUAS platform to another. The use of an external tracking system also enables a more uniform comparison of navigation and traversal capabilities. However, use of an external tracking system may need to be balanced with the additional cost and effort associated with preliminary setup of such a system. Once set up, the tracking system can be used for testing of many sUAS platforms in subsequent evaluations.
- The choice of the external tracking system is left open to the judgment and experience of the evaluation team. Evaluations in outdoor environments may reasonably rely on established localization technologies such as GPS, or cellular network-based localization. However, since these test methods are for indoor or subterranean environments (without potential access to GPS or cellular networks), alternative apparatus is required for external tracking and evaluation. These may include systems based on motion capture, Ultra Wide Band (UWB), or Received Signal Strength Indicator (RSSI) localization technologies, or any other localization system that can be used in indoor or subterranean environments for evaluation studies.
- The authors of the test methods handbook do not recommend one tracking system over another. The design choice is left to the judgment and experience of the evaluation team. The only requirement for the evaluation study is that the same tracking system be used for evaluating the navigation performance of all sUAS that are to be compared.
- Two important notes pertaining to external tracking systems:
    - *Position of localization nodes*: Most external systems require placing a marker or a localization node on the sUAS being tracked. It is important to identify and note the position on the drone that this marker or node is affixed to. Since the tracking system tracks this node and not the sUAS itself, an appropriate rigid-body transformation may be required to obtain the location of the center (of gravity) of the sUAS.
    - *Interference*: Some external tracking systems may rely on the use of radio frequencies or similar communication technologies. These tracking systems are thus vulnerable to interference, or alternatively their usage may interfere with the safe operation of the sUAS itself. The evaluation team should take appropriate measures (such as testing interference and radio frequency bands being used), before conducting the evaluation.

## Metrics

Since the sUAS platforms are categorized into two classes, i.e. those with obstacle avoidance, and those that are collision resilient, the metrics for these will also differ accordingly. If a system has both obstacle avoidance and collision resilient properties, it may be evaluated separately for both criteria. We discuss the numerical and categorical metrics for obstacle avoidance and collision resilience tests below:

**Obstacle Avoidance (OA) metrics:**

The following metrics are used for evaluating obstacle avoidance performance of the sUAS platforms. The first three are numerical metrics, while the last one is a categorical metric.

- Number of collisions: Number of test flights in which the sUAS collided with an obstacle (measured across 5 test flights).
- Minimum distance to obstacle: This metric evaluates the minimum distance between the sUAS and the obstacle during flight, averaged over 5 flights. It determines how close the sUAS came to the obstacle at any point during the test flight. A related work [Tan et al., 2017] studies the impact of separation distance on sUAS operations, but does not evaluate this from the perspective of the minimum distance to obstacle.
- Minimum time to collision (TTC): The minimum Time to Collision (TTC) index is widely used in the study of safety critical operations of vehicles. The TTC index time represents the time between the current time and a potential future collision with the obstacle. Specifically, the TTC index at any time instant *t* is defined as the "time that remains until a collision between two vehicles would have occurred if the collision course and speed difference are maintained" [Minderhoud and Bovy, 2001].

  The TTC for an sUAS with a static obstacle is defined as the ratio of the distance to the obstacle and relative speed of the sUAS and obstacle at time instant *t*. Since most of our tests are with static obstacles, the TTC index has been evaluated as the ratio of the distance to the obstacle and speed of the sUAS at time instant *t*. The TTC index may be higher for an sUAS that is distant from an obstacle but flying at high speed, as compared to an sUAS that is closer to the obstacle, but flying at very low speed. *The minimum TTC index is evaluated as the minimum value of the TTC index at any point during the flight, averaged over 5 flights.*

- Categorical OA metric: This categorical metric is used to determine the various success or failure scenarios of the sUAS operation in indoor or subterranean environments. A lower alphanumeric is better, i.e. OA-A1 indicates better performance than OA-B2. The following table and discussions explain the different types of the Categorical OA metric.

| Categorical OA metric | Description of observed behavior during test |
|---|---|
| OA-A1 | **A: Obstacle avoided:** This represents perfect operation of the obstacle avoidance capabilities of the sUAS during the flight test. In this scenario, the obstacle was successfully detected, and the autonomous avoidance maneuver was successfully completed. |
| OA-B1 | **B: Obstacle was not avoided:** The levels in this category correspond to the scenario where the autonomous sUAS could not avoid a collision with the obstacle.<br>**Category OA-B1** corresponds to the scenario where the sUAS was able to detect the obstacle, but was unable to avoid a collision. Subsequent to the collision with the obstacle, the sUAS survived.<br>An example of this scenario could be that after the obstacle was detected, the sUAS autonomously executed an avoidance maneuver, but this was not successfully completed. However, the sUAS survived the collision, i.e. it remained operational, for example by successfully completing a safe landing. |

| Categorical OA metric | Description of observed behavior during test |
|---|---|
| OA-B2 | Similar to Category OA-B1, **Category OA-B2** corresponds to the scenario where the obstacle was detected, and an avoidance maneuver was unsuccessfully executed leading to a collision, **but** the sUAS did not survive the collision, i.e. it became inoperational.<br>The B1-B2 categorization indicates that, for OA metrics, the ability to detect an obstacle and execute an avoidance maneuver is given more weight than categories B3 and B4 (where an obstacle is not detected). Note that continued operation after a collision (as explained for categories B1 and B3) is weighted more in the collision resilience (CR) metrics. |
| OA-B3 | **Category OA-B3** is similar to OA-B1, except that the obstacle was not detected. In other words, the sUAS did not detect an obstacle (and hence did not execute an avoidance maneuver), leading to a collision, but the sUAS survived the collision. This category represents a failure in the obstacle detection capabilities, even though the sUAS survived the collision, for example, by executing a safe landing. |
| OA-B4 | **Category OA-B4** is similar to OA-B3, in that an obstacle was not detected, so an avoidance maneuver was executed, leading to a collision **but** the sUAS did not survive the collision. |
| OA-C1 | **Category OA-C1** corresponds to a scenario where the evaluation team had to terminate the test to ensure safety and structural integrity of the sUAS platform. In the evaluation team's assessment, the autonomous obstacle avoidance capability of the sUAS appeared to be compromised (either due to algorithmic, sensing, or physical limitations) with the potential of imminent fatal collision. This represents the highest failure modality from the perspective of autonomous obstacle avoidance and continued operation of the sUAS platform. |

**Collision Resilience (CR) metrics:**

Recently, there have been efforts to determine the severity of sUAS collisions, but these have been directed towards impacts with business or commercial jets [Olivares et al., 2017], or with people on the ground [Arterburn et al., 2017]. Consequently, a new set of metrics are required to examine the effects of collisions on the sUAS platform itself. We have developed the following numerical and categorical metrics for measuring the collision resiliency of the tested sUAS platforms in subterranean or indoor environments:

- Modified Acceleration Severity Index (MASI): Traditionally, the Acceleration Severity Index (ASI) has been defined to determine the severity of a vehicle collision with an obstacle or stationary roadside element [Hodgson et al., 1970; Tsoi and Gabler, 2015; Pawlak 2016]. In the performed tests, we use a modified Acceleration Severity Index (MASI) appropriate for use with sUAS platforms. The dimensionless MASI metric is defined as:

$$MASI = \frac{1}{g}\sqrt{a_x^2 + a_y^2 + a_z^2}$$

  where $a_x$ represents the longitudinal acceleration of the sUAS, $a_y$ represents the lateral acceleration of the sUAS, $a_z$ represents the vertical acceleration of the sUAS, and $g$ represents the acceleration due to gravity (9.8 m/s^s^). All quantities are in units of m/s^s^. For the performed tests, sUAS flight only occurred in the horizontal plane, so the we assume that $a_z = 0$ m/s^s^. This is done to avoid including failures (such as sudden drops due to sUAS malfunction) in the analysis. *The MASI metric is calculated as the maximum deceleration of the sUAS recorded during the evaluation test flight, averaged over 5 flights.*

- Maximum Delta-V: The Maximum Delta-V metric calculates the maximum value of the change in velocity before and after collision over a given time window [Shelby, 2011; Wusk and Gabler, 2017]. As stated in 49 CFR Section 563 regulation for event data recorders that measure vehicle collision characteristics, the metric should be calculated over a period of 0.3s, beginning from the time of collision between vehicles [NHTSA, 2006]. For the performed sUAS tests, the Maximum Delta-V is evaluated over a 0.3s window that begins at the time instant of sUAS' collision with the obstacle. A caveat for this test is that the sensing apparatus must record position and velocity information at a frequency of at least 10 Hz to obtain a good estimate of the metric.

- Categorical Collision Resilience (CR) metric: This categorical metric is used to determine the various success or failure scenarios of the sUAS operation in indoor or subterranean environments. Broadly, the three categories represent successful test (A), failed test (B), and test abandonment (C). A lower alphanumeric is better, i.e. CR-A1 indicates better performance than CR-B2. The following table and discussions explain the different types of the Categorical CR metric in more detail.

| Categorical CR metric | Description of observed behavior during test |
|---|---|
| CR-A1 | **A: Resilient:** The category level A represents that the sUAS passed the resiliency test with no or little degradation in operation.<br>Specifically, **Category CR-A1** corresponds to the scenario that<br>that the sUAS platform collided with the obstacle, but did not suffer any failure, and was able to continue operations. This represents perfect collision resilience properties. |
| CR-A2 | **Category CR-A2** is similar to CR-A1 except for the fact that the sUAS temporarily loses operational continuity. For example, after a collision with an obstacle has occurred, the suAS may retreat to a fail-safe mode, such as executing a safe landing. However, this action does not imply lack of resilience, as the sUAS can return to its operational capacity after the fail-safe mode is disabled (such as return to flight after a safe landing). |
| CR-A3 | **Category CR-A3** is similar to CR-A2 except for the fact that in this scenario the sUAS fails to enter a fail-safe mode after the collision event, but is still able to return to operation after the event. Thus, the sUAS suffers only a temporary loss in operational continuity. For example, the sUAS may suffer an uncontrolled descent (i.e. crash) in this scenario as opposed to the scenario in CR-A2 where, for example, the descent was a programmed landing activated by a fail-safe mode. |
| CR-B1 | **B: Lack of resiliency:** The levels in this category correspond to the scenario where the sUAS failed to resolve a collision gracefully.<br>**Category CR-B1** is similar to the sUAS behavior observed in CR-A2 in that the sUAS enters a fail-safe mode after a collision. However, unlike the behavior observed in Category CR-A2, the sUAS is unable to return to operation due to limitations enacted by the underlying control system. Additional human intervention may be required to resume operation. For example, the sUAS may execute a controlled landing in fail-safe mode but reduced sensing (e.g. camera view is restricted due to proximity to the obstacle) may prevent a |

| Categorical CR metric | Description of observed behavior during test |
|---|---|
| | return to operation. All other systems remain operational. |
| CR-B2 | **Category CR-B2** indicates further degradation of behavior as compared to CR-B1. Specifically, in this scenario, the sUAS may execute a successful landing in fail-safe mode, but communication drop-out prevents a return to operation. Without communication, teleoperation of the sUAS cannot be carried out, i.e. sUAS integrity is maintained, but flight capabilities are lost. In this scenario, the sUAS control system may or may not be operational, but since control commands cannot be communicated.<br>This failure mode does not apply to the collision resilience of autonomous sUAS platforms. |
| CR-B3 | **Category CR-B3** is one level of further degradation as compared to CR-B2. In this scenario, the sUAS' attempt to execute a safe landing after the collision is unsuccessful. Specifically, the sUAS may have landed on with a tilt or flipped over, i.e. not in its usual take-off configuration. In this scenario, resume flight operations after a take-off event cannot be guaranteed. In this scenario, the control system and/or the communication channel may or may not be operational. |
| CR-B4 | **Category CR-B4** indicates the final level of performance degradation as it includes loss of structural integrity of the sUAS platform, presenting a *potential permanent loss of operational continuity*. It is possible to provide a minor distinction between actual structural damage and disintegration of some components (such as protective propeller guards), but they both indicate a lack of collision resilience of the sUAS structural frame. |
| CR-C1 | **Category CR-C1** corresponds to a scenario where the evaluation team had to terminate the test to ensure safety and structural integrity of the sUAS platform. |

## Procedure

The same procedure is used for evaluating both Obstacle Avoidance and Collision Resilience tests.

**Initial set-up:**

1. Clear flight test area.
2. If no external tracking system is used, then proceed to step 3. If an external tracking system is used:
    a. Setup the external tracking system as per instructions provided with the tracking system.
    b. Repeat steps c through f before each sUAS evaluation flight
    c. Set one external tracking system in the geometric center of the test space and do not disturb it for the duration of the next step.
    d. Collect position data from this tracker for a duration at least 2 times longer than the duration of the maximum expected flight time in the test. This data will be used later to determine the accuracy and precision of the tracking system, enabling comparison across different tests.
    e. Use the recorded position data and calculate the accuracy (mean value of recorded positions, averaged over the duration of the test) and precision (standard deviation of recorded positions, across the duration of the test) of the external tracking system.
    f. Proceed to the next step to evaluate sUAS obstacle avoidance/collision resilience capabilities

**Evaluation procedure:**

1. Clear the test flight zone.
2. Set up the obstacle (e.g., material partition). In the case of the wall test this might not be necessary if the wall of the room is used.
3. Mark a take-off point at least 1.5m away from the obstacle and place the sUAS at this point.
4. Begin telemetry recording (i.e., internal tracking system), if available.
5. Initiate hover sequence at take-off point.
6. Perform the flight test, once for each of the following flight test paths. (Also see additional notes 1 and 2 regarding speed commands, at the end of Step 6):
   <u>For obstacles such as walls, chain link fences, and plastic meshes, use the following test paths:</u>
    a. *Longitudinal motion and collision*: In this test, the sUAS is given a forward (longitudinal) motion command and the test flight path is perpendicular to the obstacle, resulting in a head-on collision.
    b. *Lateral motion and collision (Starboard side)*: In this test, the sUAS starboard side is positioned closer to the obstacle. The sUAS is commanded to move in a lateral direction perpendicular to the obstacle, with the collision occurring on the starboard side.
    c. *Lateral motion and collision (Port side):* In this test, the sUAS port side is positioned closer to the obstacle. The sUAS is commanded to move in a lateral direction perpendicular to the obstacle, with the collision occurring on the port side.
    *d.* *Longitudinal motion and collision at incidence angle of 45 degrees (Starboard side)*: In this test, the sUAS is commanded to move in a longitudinal direction, approaching the obstacle at an incidence angle of 45 degrees, with the collision occurring on the starboard side.
    *e.* *Longitudinal motion and collision at incidence angle of 45 degrees (Port side)*: In this test, the sUAS is commanded to move in a longitudinal direction, approaching the obstacle at an incidence angle of 45 degrees, with the collision occurring on the port side

   <u>For obstacles such as doors, use the following test paths:</u>
    f. *Longitudinal motion and collision*: In this test, the sUAS is given a forward (longitudinal) motion command and the test flight path is perpendicular to the obstacle, resulting in a head-on collision:
        i. with door fully closed,
        ii. with door partially open at a 45 degree angle, and
        iii. with door fully open

NOTE 1: For an autonomous system, this command may be given by providing a waypoint that coincides with the location of the obstacle.
NOTE 2: For a manually operated sUAS, determine the speed settings of the sUAS. Provide forward flight command to fly the sUAS manually at 0.5 m/s (if speed controls allows this), or whichever speed setting is closest to 0.5 m/s (in case the sUAS has a fixed number of speed settings). Calibration may be required by the flight operator to determine the corresponding control associated with this speed.

7. Terminate flight after collision or when the sUAS reaches a hover state.
8. Stop telemetry recording.
9. Repeat Steps 5-8 for a total of five times for the sUAS platform, and evaluate the metrics. The flight tests may be grouped so that five runs of Step 6a are performed together. A similar grouping approach can be used for Steps 6b-f as well.
10. Repeat Steps 2-9 for various obstacles being tested. The list of obstacles may include (but is not limited to) wall partitions, plastic meshes, and chain link fences. For door obstacles

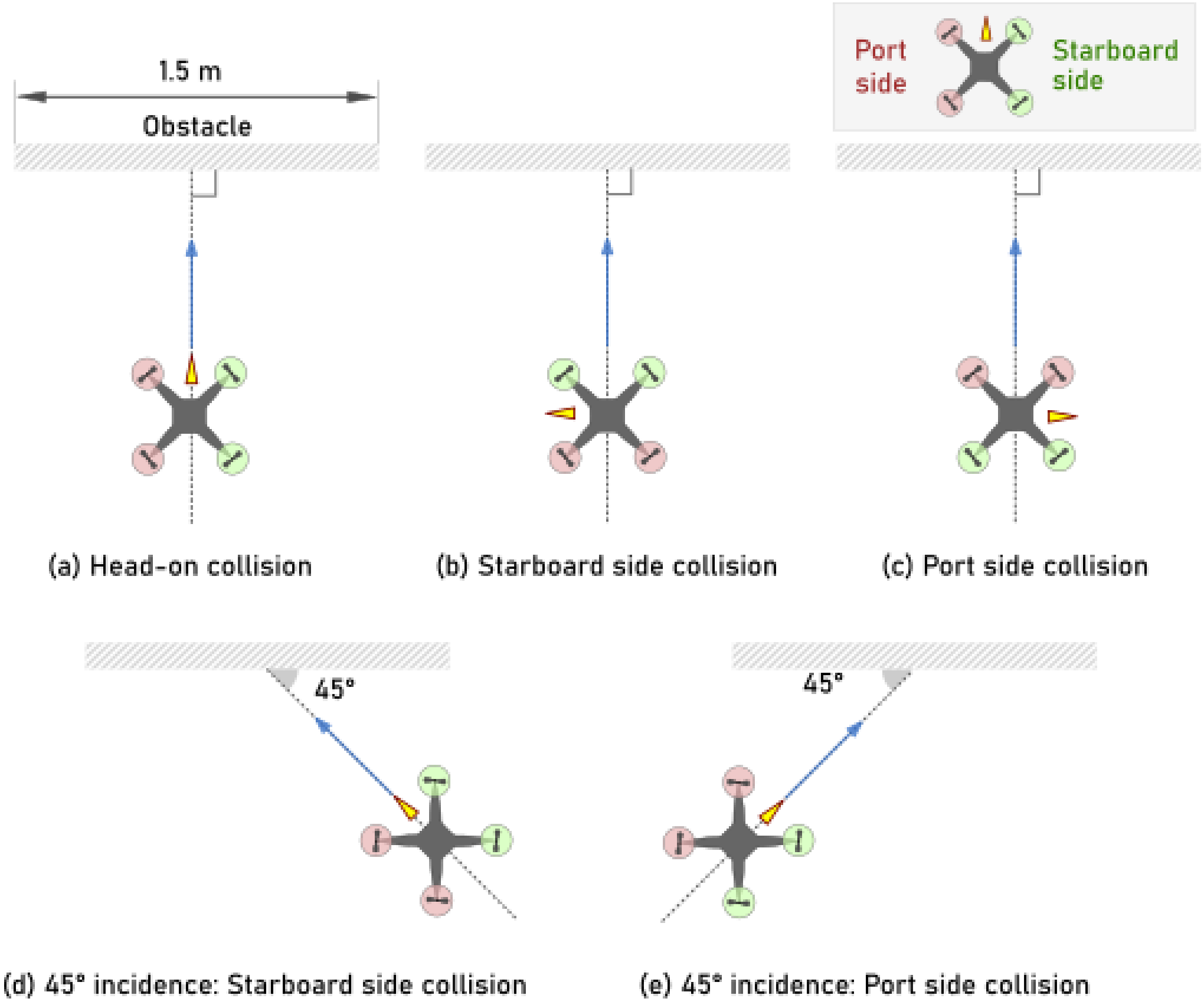

*Figure 3. Obstacle avoidance and collision resilience tests for walls, plastic mesh, and chain link fences*

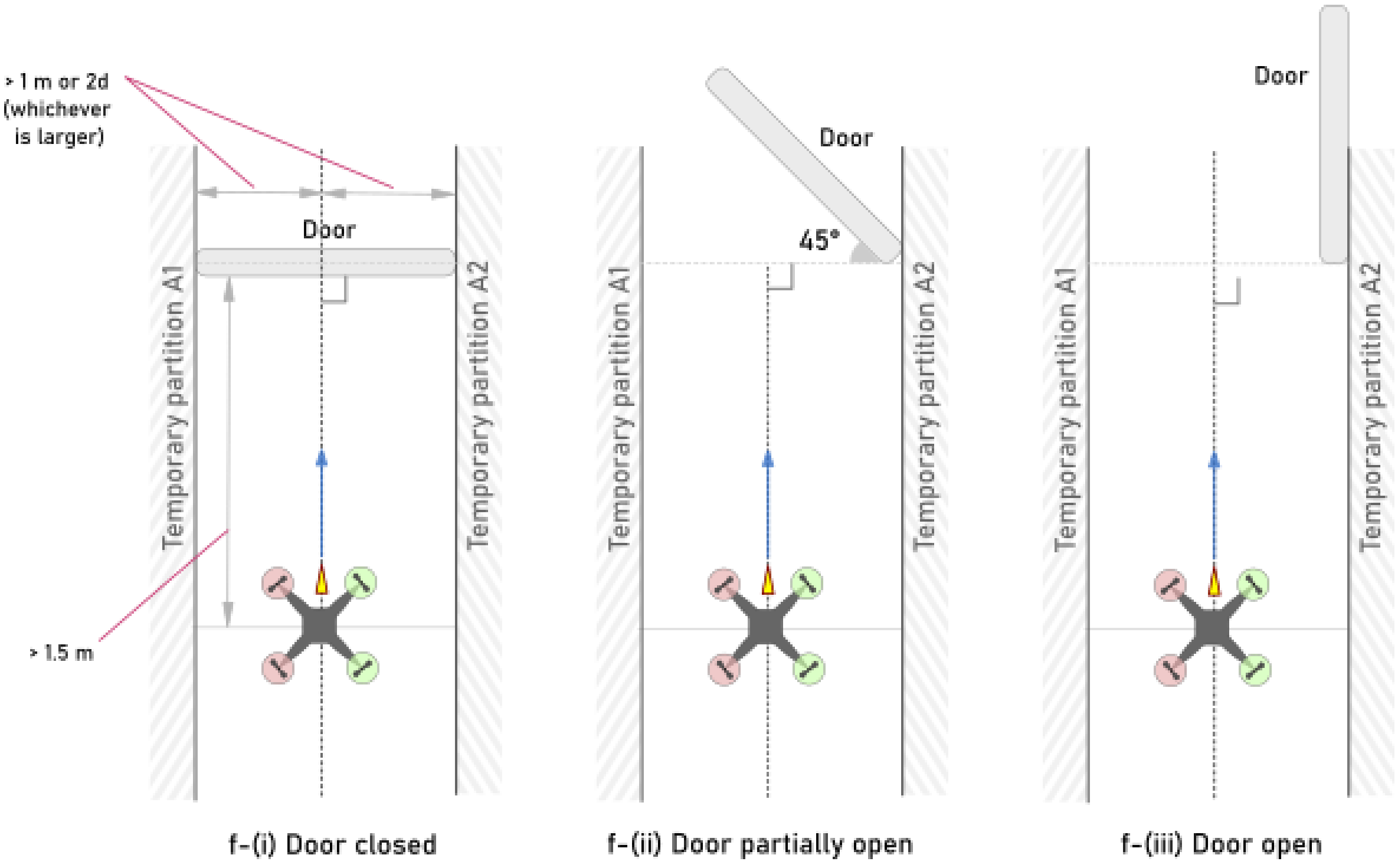

*Figure 4. Obstacle avoidance and collision resilience tests for door obstacles*

## Example Data

Obstacle Avoidance numerical metrics:

| sUAS platform | Number of collisions | Minimum distance to obstacle | Minimum time to collision | Maximum flight deceleration |
|---|---|---|---|---|
| A | x | 0.0 m | 0.0 sec | 1.1 $m/s^2$ |
| | x | 0.0 m | 0.0 sec | 0.6 $m/s^2$ |
| | - | 0.32 m | 0.5 sec | 0.8 $m/s^2$ |
| | - | 0.24 m | 0.2 sec | 0.8 $m/s^2$ |
| | - | 0.26 m | 0.8 sec | 0.9 $m/s^2$ |
| **Count or average** | **2** | **0.164 m** | **0.3 sec** | **0.84 $m/s^2$** |
| B | - | 0.57 m | 0.8 sec | 1.0 $m/s^2$ |
| | x | 0.0 m | 0.0 sec | 1.6 $m/s^2$ |
| | - | 0.8 m | 1.2 sec | 0.9 $m/s^2$ |
| | - | 0.44 m | 0.8 sec | 0.9 $m/s^2$ |
| | - | 0.15 m | 0.2 sec | 0.9 $m/s^2$ |
| **Count or average** | **1** | **0.392 m** | **0.6 sec** | **1.06 $m/s^2$** |

Obstacle Avoidance categorical metric:

| Obstacle type | sUAS Platform A (Distribution over 5 test flights) | | | | | |
|---|---|---|---|---|---|---|
| | Obstacle Avoidance (OA) Category | | | | | |
| | OA-A1 | OA-B1 | OA-B2 | OA-B3 | OA-B4 | OA-C1 |
| **Wall** | 90% | 5% | 0% | 0% | 0% | 5% |
| **Plastic mesh** | 80% | 0% | 10% | 0% | 10% | 0% |
| **Chain link fence** | 40% | 0% | 30% | 0% | 20% | 10% |
| **Door** | 70% | 0% | 0% | 20% | 0% | 10% |

Collision Resilience Severity:

| sUAS platform | Modified Acceleration Severity Index (MASI) | Maximum Delta-V |
|---|---|---|
| A | 0.17 | + 0.7 m/s |
| | 0.2 | + 0.8 m/s |
| | 0.14 | + 1.2 m/s |
| | 0.15 | + 0.8 m/s |
| | 0.16 | + 1.1 m/s |
| **Average** | **0.164** | **+ 0.92 m/s** |
| B | 0.1 | + 0.4 m/s |
| | 0.11 | + 0.4 m/s |
| | 0.05 | + 0.7 m/s |
| | 0.06 | + 0.2 m/s |
| | 0.06 | + 0.1 m/s |
| **Average** | **0.076** | **+ 0.36 m/s** |

Collision Resilience Success table:

| Obstacle type | sUAS Platform A (Distribution over 5 test flights) | | | | | | | |
|---|---|---|---|---|---|---|---|---|
| | Collision Resilience (CR) Category | | | | | | | |
| | CR-A1 | CR-A2 | CR-A3 | CR-B1 | CR-B2 | CR-B3 | CR-B4 | CR-C1 |
| **Wall** | 50% | 10% | 0% | 5% | 20% | 10% | 0% | 5% |
| **Plastic mesh** | 80% | 10% | 0% | 0% | 0% | 0% | 0% | 10% |
| **Chain link fence** | 60% | 10% | 0% | 0% | 30% | 0% | 0% | 0% |
| **Door** | 40% | 10% | 20% | 0% | 0% | 20% | 0% | 5% |

# Navigation

Operations involving sUAS for subterranean and other constrained environments typically include encountering a variety of obstructions to movement in vertical and horizontal space such as doorways, stairwells, shafts, piping, uneven flooring, deadfalls, and other natural and man-made hindrances, which include navigation challenges that are horizontal, vertical, or a combination of the two. The objective of tests in the Navigation category is to determine the ability of the sUAS to physically navigate to desired locations, traverse predefined paths, and navigate through confined spaces and apertures.

## Position and Traversal Accuracy

Affiliated publications: [Meriaux and Jerath, 2022]

### Purpose

The purpose of this test method is to examine the ability of an sUAS to perform indoor traversal along a given path between two waypoints. The test method consists of different flight missions and configurations that evaluate the ability to navigate along combinations of linear paths for horizontal traversal.

### Summary of Test Method

The test method consists of four different tests: (a) wall following, (b) waypoint navigation, (c) straight line path traversals, (d) corner navigation, and (e) aperture navigation. Each test is conducted five times for each sUAS being evaluated and the metrics are aggregated across tests. All tests require the availability of telemetry data either via on-board vendor provided datastreams, or external tracking systems. All tests can be performed with minimal additional apparatus (beyond the tracking system) if appropriate subterranean or indoor environments are available. If existing environments do not meet specifications, they can be constructed using readily available materials. All evaluation flights should be performed using line-of-sight operation as remote FPV operation may confound navigation capabilities of the sUAS.

Wall Following: The wall following test examines the ability of the sUAS to navigate a specific traversal path while operating in the vicinity of a wall at both 1 m (close) and 2 m (far) from the wall. This is a common use case scenario in specific indoor and subterranean operations. This test is performed in two common sUAS orientations for such missions: parallel (i.e., sUAS camera/front is pointed parallel to the wall surface while moving along it, pitching to fly forward) and perpendicular (i.e., sUAS camera/front pointed perpendicular to the wall while moving along it, strafing right or left to fly sideways).

Waypoint Navigation: The waypoint navigation evaluation methodology determines the ability of the sUAS to land at the desired waypoint location.

Linear Path Traversal: The straight line traversal will require the sUAS to fly in a rectangular pattern made of four (4) linear path traversals. Deviations from the rectangular path will be used to evaluate the ability of the sUAS to perform straight line traversals. If a limited flight testing area is available, a single linear path traversal may be used for evaluation (instead of rectangular path).

Hallway Navigation: This test seeks to examine the ability of the sUAS to navigate a confined space with turns (such as a corridor or hallway). To eliminate the confounding factors associated with piloting skills, the test requires a flight pattern such that the corner navigation is performed via an in-place 90-degree turn, rather than smooth turning curves that expert pilots might execute in confined spaces. This test examines the effects of wind eddy currents in cases where there are walls on both sides of the sUAS (such as hallways). Hallway-induced wind eddy currents are expected to generate higher turbulence than the other navigation tests discussed here.

Corner Navigation: This test is similar to the hallway navigation test, but with corner partitions only on one side of the sUAS' flight path

Aperture Navigation: This test evaluates the ability of the sUAS platform to successfully navigate through an aperture. In subterranean environments, sometimes drones need to fly in such cases. They must be able to do so without contact with the surrounding material but if it does it must be able to withstand collision. This is why this test is not numerically evaluated like the other navigation tests but has a tiered result table listed below.

| Result | Condition of test | Explanation of result |
|---|---|---|
| A1 | Pass through no contact | Drone went through aperture did not touch any sides |
| A2 | Pass through contact no rip | Drone went through aperture did touch sides but no tears |
| A3 | Pass through contact ripped | Drone went through aperture did touch sides tear occurred |
| B1 | Failed pass through due to contact | Drone was unable to go through and land properly due to contact with aperture |

## Apparatus and Artifacts

The environment for conducting the evaluation should be bounded with walls to form a free space with dimensions of at least *10d x 15d*, where *d* represents the maximum horizontal dimension of the sUAS being tested (typically prop tip to opposing prop tip). The area should provide enough vertical space to safely takeoff, aviate, and land without fear of colliding with the floor or ceiling.

This area will be further divided into a buffer zone (*1d* wide band around the outer edge), flight zone (*8d x 13d*; i.e., the remaining space outside of the buffer zone), and a testing zone within the flight zone (*7.5d x 10d*); see Figure 1. All provided dimensions are minimum measurements. The environment should be free from any obstructions (e.g., structural support columns) to enable safe flight and accurate sensing.

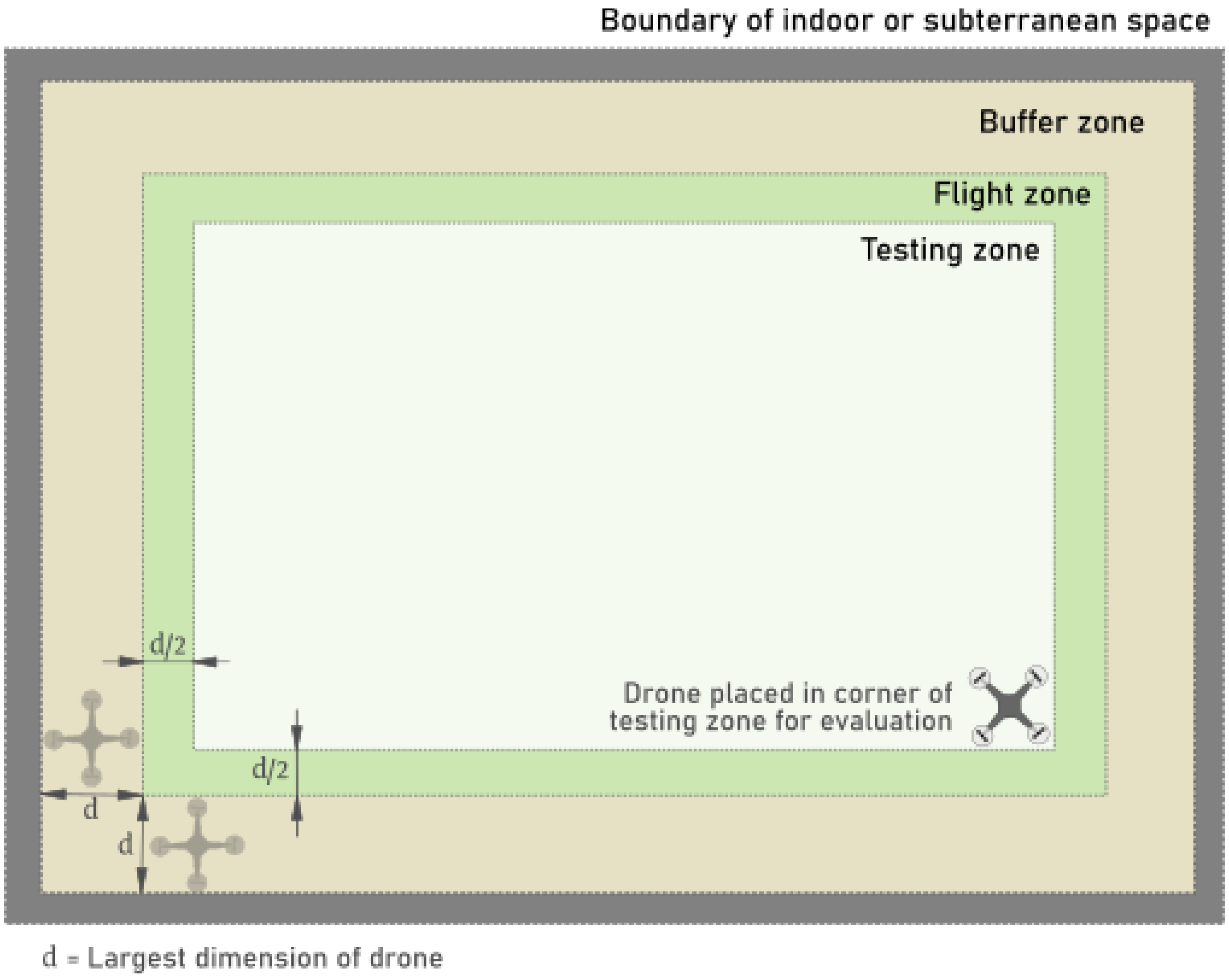

*Figure 1. Environment dimensional requirements.*

The corner navigation test requires additional setup with temporary partitions. Setup the test area with temporary partitions as shown. Ensure that temporary partitions A1 and A2 are parallel to each other and separated from each other by a distance greater than 2 m or *4d* (whichever is larger), where d is the largest dimension of the drone. The temporary partitions should be tall enough such that the sUAS always operates at a height less than that of the partition. They may be constructed with any readily available stiff material (wood, corrugated cardboard, etc.) such that sUAS flight does not displace them. Ensure that temporary partition B1 is parallel to the wall and separated by at least 2 m or *4d* (whichever is larger). The wall may be replaced with another temporary partition B2 if required. An existing corridor or hallway that satisfies these dimensions can also be used.

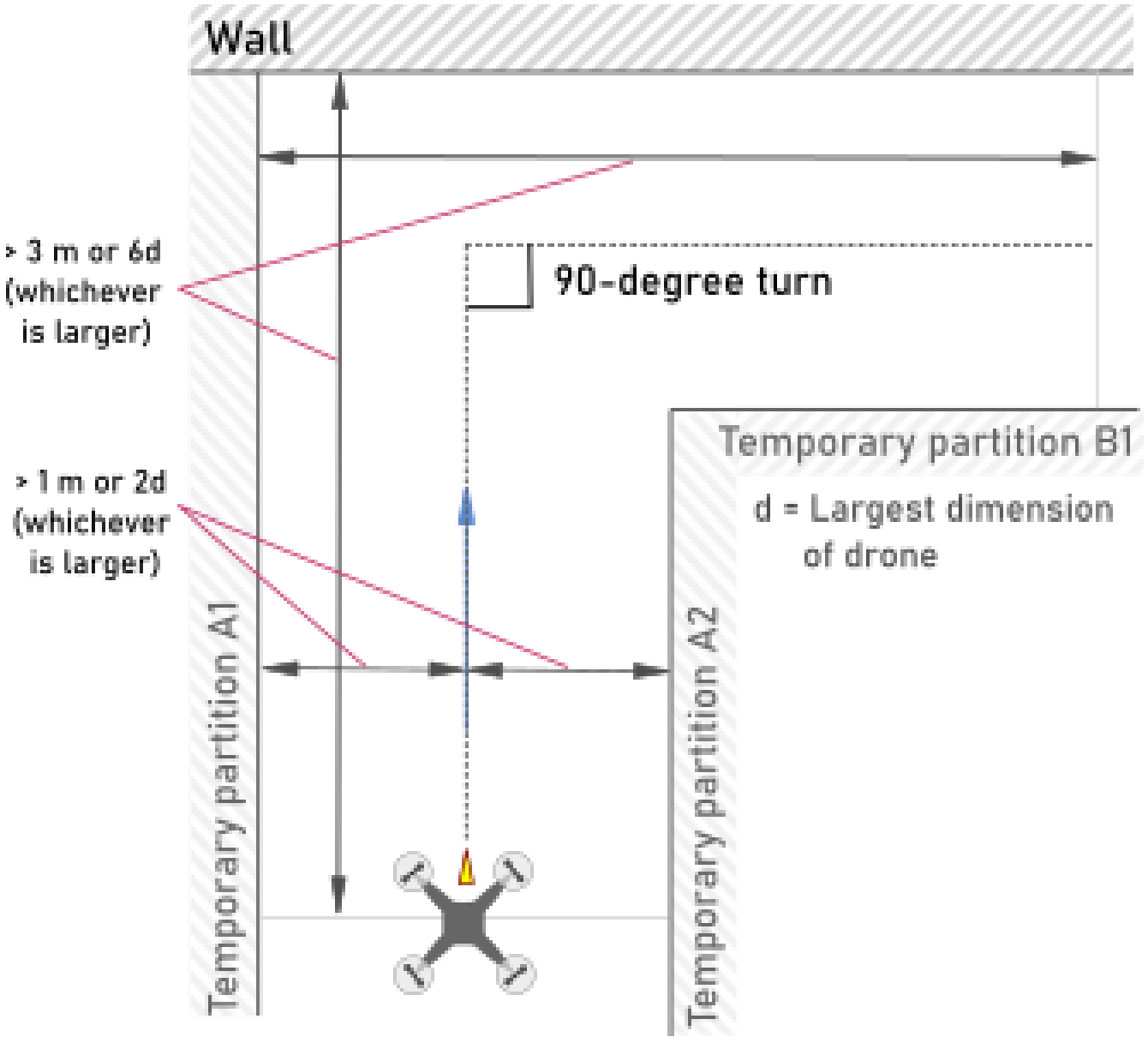

*Figure 2. Test apparatus setup for corner navigation test*

## Equipment

Data is collected to measure and evaluate the positional performance of the sUAS in each test. There are two primary options to collect the position information of the sUAS in order to determine its ability to successfully traverse the designated path: (a) internal tracking system existing on-board the sUAS platform, or (b) an external tracking system. External tracking requires hardware and software setup by the evaluators, which should be provided by the vendors of the specific tracking system used by the evaluators. For a given comparison of sUAS platforms using the test method, only one of the data collection methods should be used; i.e., the data collection methods should not be mixed and matched. Additional information about each data collection method is provided below.

Internal tracking system:

- This option requires access to sUAS telemetry, which depends on sUAS vendor assistance or permission. If access to sUAS telemetry is available, then the internal tracking system can be used to obtain sUAS pose and velocity data.
- It is possible that this data has been pre-processed and filtered by on-board systems. If only one sUAS is being evaluated, then such pre-processing does not interfere with the evaluation methods. However, if multiple sUAS are being tested and their performances compared, then it is required that data from the internal tracking systems on-board each sUAS should be available for the evaluation study. If on-board telemetry is not available or inaccessible for even one sUAS platform, then the evaluation and comparison study should either (a) use an external tracking system, or (b) remove the platform from the study. This is necessary in order to avoid inconsistencies in performance comparison between sUAS platforms.

External tracking system:

- In certain situations, it may be preferable to deploy an external tracking system to evaluate performance. This approach may enable the evaluator to circumvent issues related to telemetry access and proprietary communication protocols, as well as reduce reliance on pre-processed telemetry streams which may differ from one sUAS platform to another. The use of an external tracking system also enables a more uniform comparison of navigation and traversal capabilities. However, use of an external tracking system may need to be balanced with the additional cost and effort associated with preliminary setup of such a system. Once set up, the tracking system can be used for testing of many sUAS platforms in subsequent evaluations.
- The choice of the external tracking system is left open to the judgment and experience of the evaluation team. Evaluations in outdoor environments may reasonably rely on established localization technologies such as GPS, or cellular network-based localization. However, since these test methods are for indoor or subterranean environments (without potential access to GPS or cellular networks), alternative apparatus is required for external tracking and evaluation. These may include systems based on motion capture, Ultra Wide Band (UWB), or Received Signal Strength Indicator (RSSI) localization technologies, or any other localization system that can be used in indoor or subterranean environments for evaluation studies.
- The authors of the test methods handbook do not recommend one tracking system over another. The design choice is left to the judgment and experience of the evaluation team. The only requirement for the evaluation study is that the same tracking system be used for evaluating the navigation performance of all sUAS that are to be compared.
- Two important notes pertaining to external tracking systems:
    - *Position of localization nodes*: Most external systems require placing a marker or a localization node on the sUAS being tracked. It is important to identify and note the position on the drone that this marker or node is affixed to. Since the tracking system tracks this node and not the sUAS itself, an appropriate rigid-body transformation may be required to obtain the location of the center (of gravity) of the sUAS.
    - *Interference*: Some external tracking systems may rely on the use of radio frequencies or similar communication technologies. These tracking systems are thus vulnerable to interference, or alternatively their usage may interfere with the safe operation of the sUAS itself. The evaluation team should take appropriate measures (such as testing interference and radio frequency bands being used), before conducting the evaluation.

A tape measure is used to calculate waypoint accuracy and precision metrics.

## Metrics

For each test performed, the following metrics should be evaluated across five trials:

- <u>Path deviation</u>: Deviation of the sUAS platform's actual trajectory from a defined straight line path. Additional details are provided in the procedure description of the wall following test.
    a. For the Wall Following test, the path is a constant distance away from a wall, parallel to it (i.e., along the edge of the testing zone).
    b. For the Linear Path Traversal test, the path is one or more defined straight line(s) within the testing zone.
    c. For the Corner Navigation test, the path is two straight line(s) that constitute a 90-degree turn trajectory.
- <u>Waypoint accuracy and precision</u>: The accuracy and precision of reaching the desired waypoint, defined using the difference between the desired waypoint location and the final landing position of the sUAS. This metric is used in the Waypoint Navigation test.

## Procedure

**Initial set-up:**

1. Clear flight test area.
2. If an external tracking system is being used:
    a. Setup the external tracking system as per instructions provided with the tracking system.
    b. Repeat steps c through f before each sUAS evaluation flight
    c. Set one external tracking system in the geometric center of the test space and do not disturb it for the duration of the next step.
    d. Collect data from this tracker for a duration at least 2 times longer than the duration of the maximum expected flight time in the test. This data will be used later to determine the accuracy and precision of the tracking system, enabling comparison across different tests.
    e. Use the recorded position data and calculate the accuracy (mean value of recorded positions, averaged over the duration of the test) and precision (standard deviation of recorded positions, across the duration of the test) of the external tracking system.

**Evaluation procedures:**

Wall Following:

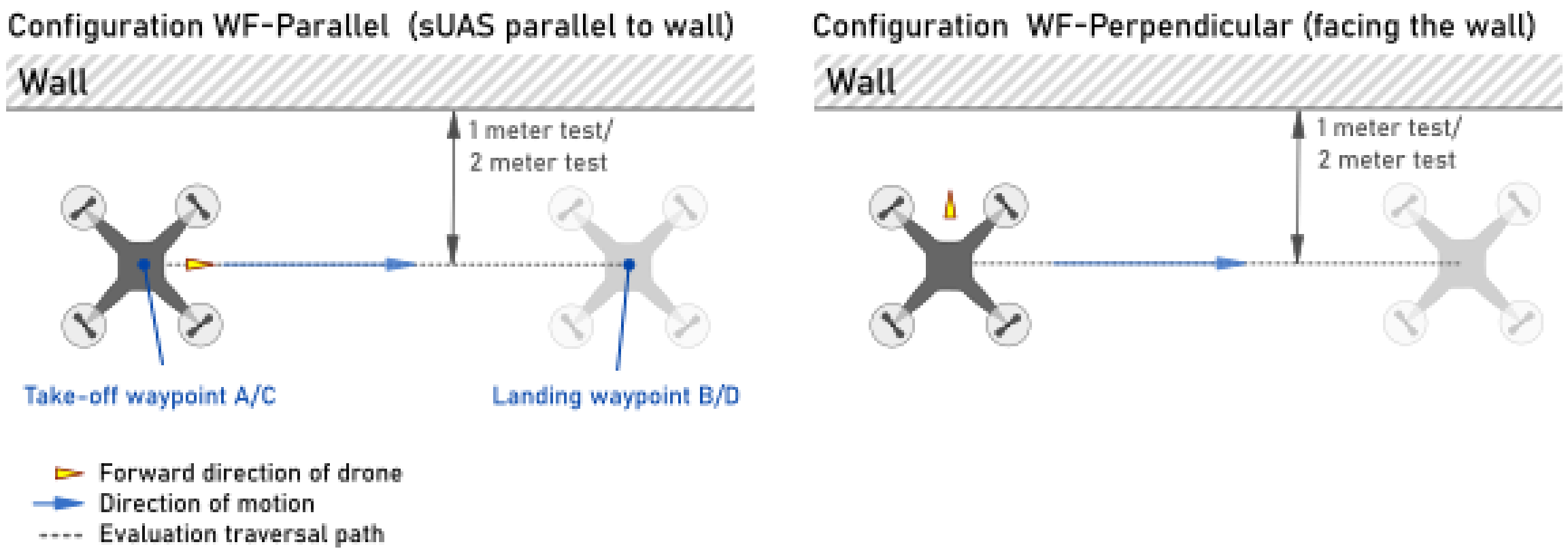

*Figure 3. Two configurations for the Wall Following test: parallel (WF-Parallel, left) and perpendicular (WF-Perpendicular, right).*

1. Mark a take-off waypoint A that is 1 m away from the wall, and a landing waypoint B that is 3 m from the take-off waypoint A and also 1 m away from the wall.
2. Similarly, mark another take-off waypoint C that is 2 m away from the wall, and a landing waypoint D that is 3 m from the take-off waypoint C and also 2 m away from the wall.
3. If available, mount the external tracking marker on the drone and mark its location in relation to the geometric center of the drone. If external tracking is not being used, then proceed to step 4.
    a. Start the pose and velocity recording of the tracking marker using the external tracking system.

For the parallel configuration:

4. Place the sUAS at take-off waypoint A with the front of the drone parallel to the wall (Configuration WF-Parallel).
5. Initiate hover sequence at take-off waypoint A.
6. Manually fly the sUAS forward, while attempting to maintain a constant distance of 1 m from the wall.
7. After 3 m of forward flight, land the sUAS at landing waypoint B.
8. Collect the position data (from internal or external tracking system), and calculate the average deviation (AD) from the straight line traversal path.
9. Repeat steps 4-8 across five flights and calculate average deviation (AD) from straight line traversal path in each flight.
10. Find the mean value and standard deviation of AD. Record these as evaluation metrics for the test.

11. Repeat the test procedure (steps 5-10) with take-off waypoint C (which is 2 m away from the wall) and landing waypoint D, while attempting to maintain a distance of 2 m from the wall.

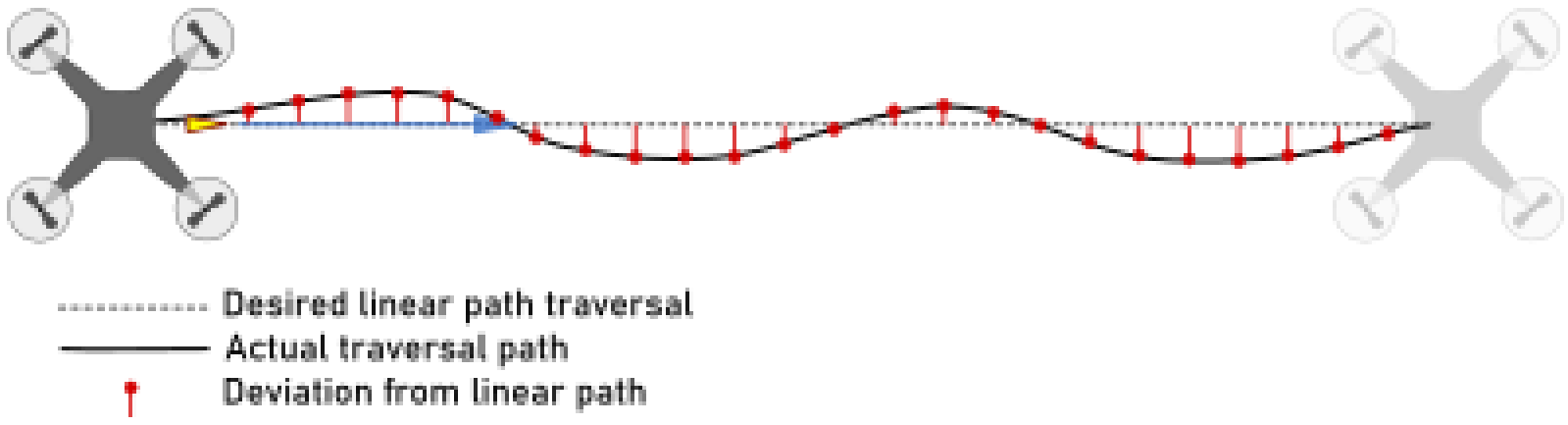

*Figure 4. Calculation of deviation from linear traversal path. The average of all deviations (AD) along the entire traversal path are calculated as a metric for a single trial. The average and standard deviation of AD is evaluated across five trials (or flights).*

For the perpendicular configuration:

4. Place the sUAS 1 m away from a wall with the front of the drone facing the wall.
5. Initiate hover sequence at take-off waypoint A.
6. Manually fly the sUAS sideways, while attempting to maintain a constant distance of 1 m from the wall.
7. After 3 meters of sideways flight, land the sUAS at landing waypoint B.
8. Collect the position data (from internal or external tracking system), and calculate the average deviation (AD) from the straight line traversal path.
9. Repeat Steps 4-8 across five flights and calculate average deviation (AD) from straight line traversal path in each flight.
10. Find the mean value and standard deviation of AD. Record these as evaluation metrics for the test.
11. Place the sUAS 2 meters away from a wall with the front of the drone parallel to the wall (Configuration WF-Perpendicular).
12. Repeat the test procedure (steps 5-10) with take-off waypoint C (which is 2 m away from the wall) and landing waypoint D, while attempting to maintain a distance of 2 m from the wall.

<u>Waypoint Navigation</u>: (note: this test is run concurrent to the Wall Following test)

1. Once the sUAS has landed at waypoint B in the Wall Following test, use a tape measure to determine the distance of the landed sUAS from the destination landing waypoint B.
2. Record the average and standard deviation of the distances obtained in step 1 across five trials.
3. Determine the last location of the sUAS that was obtained using either on-board telemetry or the external tracking system.
4. Record the average and standard deviation of the distances obtained in step 3 across five trials.
5. Repeat steps 1-4 to identify the accuracy and precision of landing at waypoint D.
6. Record the tape measure and telemetry/external tracking as metrics for the test.

<u>Linear Path Traversal</u>:

1. Mark the corners of the rectangular linear flight path, with the shorter and longer edges of the rectangular path no less than *5d* and no less than *7.5d*.
2. Place the sUAS in a corner of the testing zone.
3. Begin logging the internal and/or external tracking data.
4. Initiate hover sequence at the take-off point.
5. Manually fly the sUAS forward towards the next corner of the rectangular linear flight path.
6. Continue to fly in linear paths to the next closest corner of the rectangular flight path while maintaining the same orientation, till the sUAS has reached the original take-off point. As shown in Figure 5, the sUAS will pitch forward, roll right, pitch backward, and roll left to complete the rectangular flight path.
7. Land the sUAS, terminate the flight test and telemetry recording.
8. Use the recorded telemetry or tracking data to evaluate deviations from the rectangular linear flight path. The deviation at each recorded timestamp is calculated as the perpendicular distance from the recorded sUAS location to the desired rectangular linear flight path.

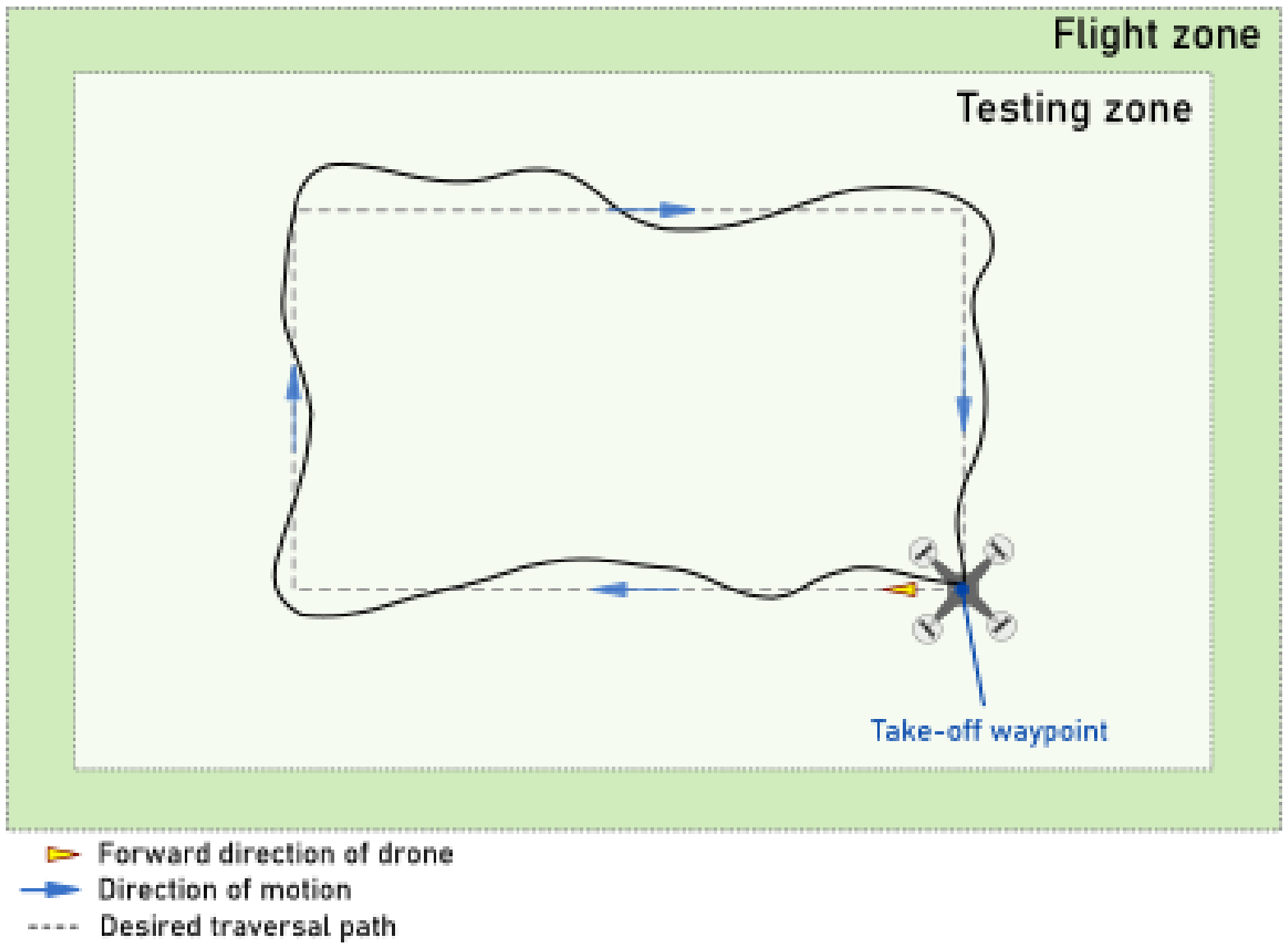

*Figure 5. Flight pattern for linear path traversal test, also indicating potential deviations as the sUAS completes this test.*

Hallway Navigation:

1. Place the sUAS in the middle of partitions A1 and A2, and at a distance of approximately 5d from the wall or partition B1.
2. Begin logging the telemetry or external tracking data.
3. Initiate hover sequence at take-off point.
4. Manually fly the sUAS forward towards the wall or partition B1.
5. At a distance approximately equal to the maximum dimension of the sUAS, execute a 90 degree turn and fly to the end of the trajectory.
6. Land the sUAS, terminate the flight test and telemetry recording.
7. Use the recorded telemetry or tracking data to evaluate deviations from the 90-degree turn flight path. The deviation at each recorded timestamp is calculated as the perpendicular distance from the recorded sUAS location to the desired flight path.

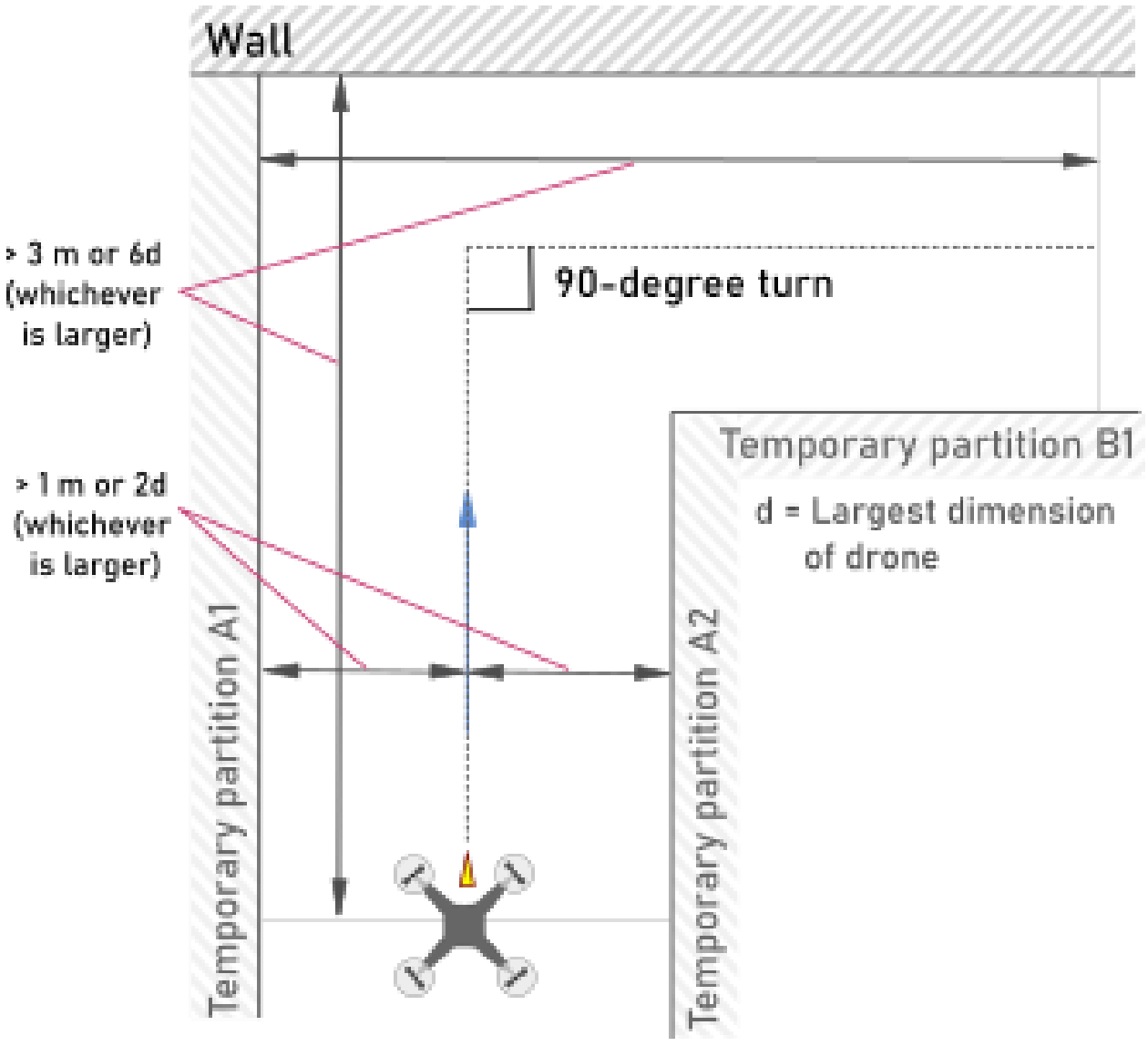

*Figure 6. Executing the Hallway Navigation test.*

Corner Navigation:

1. Place the sUAS near  partition A1, and at a distance of approximately 5d from the wall or partition B1.
2. Begin logging the telemetry or external tracking data.
3. Initiate hover sequence at take-off point.
4. Manually fly the sUAS forward towards the wall. There is no partition but the drone must follow the same path as in the partitioned test.
5. At a distance approximately equal to the maximum dimension of the sUAS, execute a 90 degree turn and fly to the end of the trajectory.
6. Land the sUAS, terminate the flight test and telemetry recording.
7. Use the recorded telemetry or tracking data to evaluate deviations from the 90-degree turn flight path. The deviation at each recorded timestamp is calculated as the perpendicular distance from the recorded sUAS location to the desired flight path.

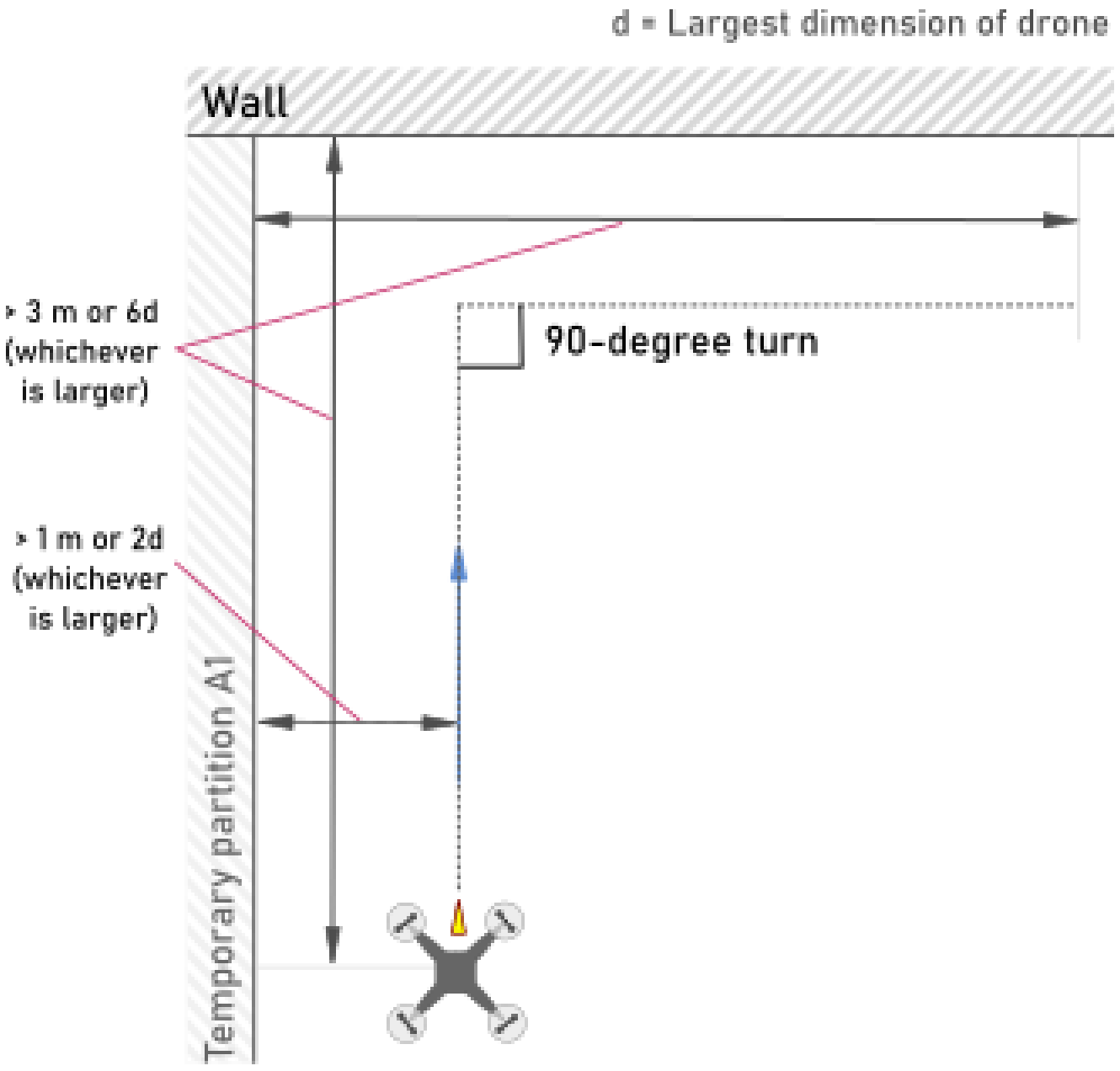

*Figure 7. Executing the Corner Navigation test.*

Aperture Navigation:

1. Place the sUAS on one side of the partition approximately 1 meter from the structure.
2. Begin logging the telemetry or external tracking data.
3. Initiate hover sequence at take-off point.
4. Manually fly the sUAS forward towards the wall or partition.
5. Fly through the aperture’s opening.
6. Land the sUAS 1 meter on the side of the structure that the drone did not takeoff from, terminate the flight test and telemetry recording.
7. Log if the drone collided with the structure and if there is any damage to the structure due to the collision as well as if the drone was able to arrive at the desired landing point.

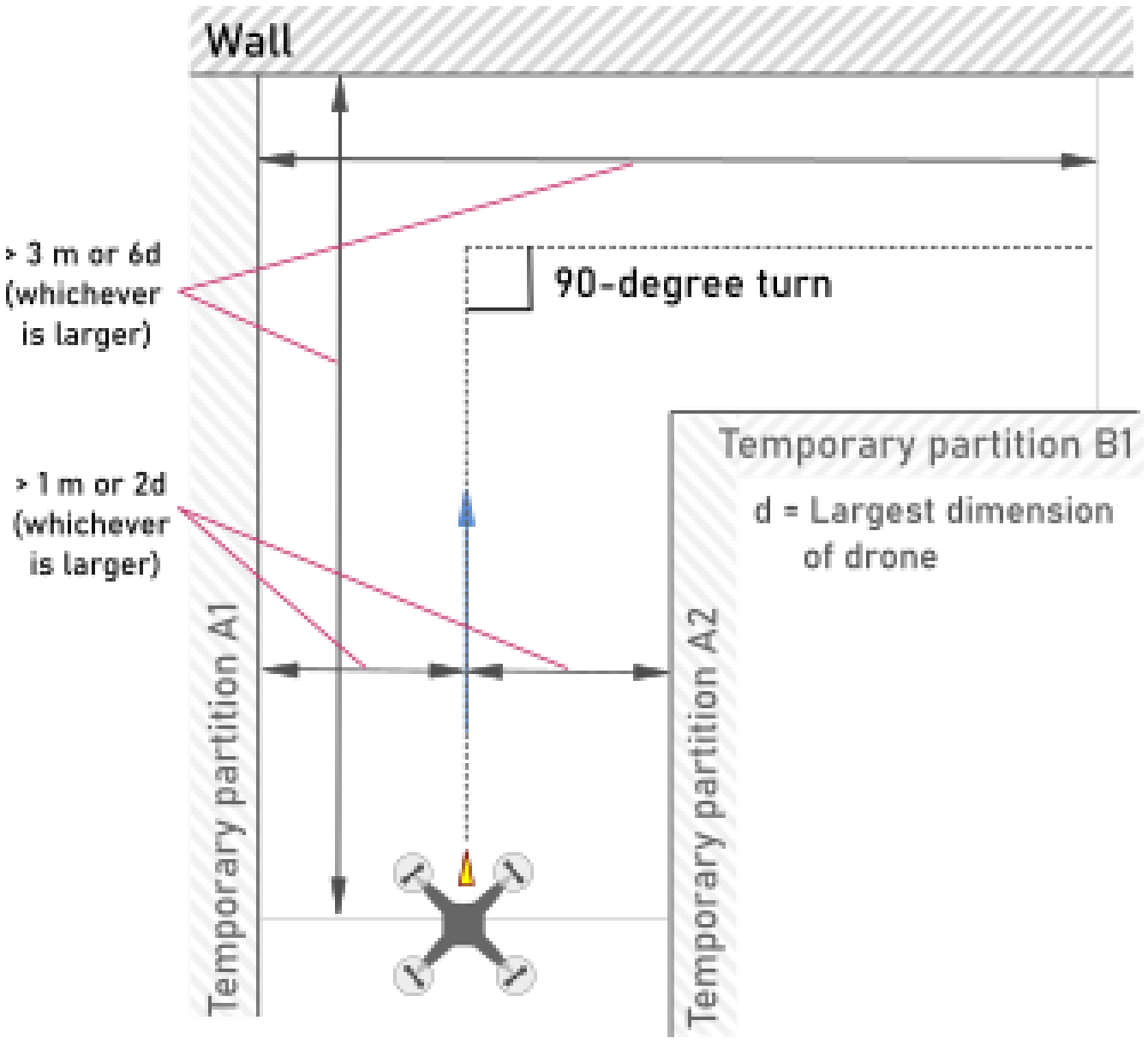

## Example Data

| sUAS platform | Path deviation | | | | | Landing position |
|---|---|---|---|---|---|---|
| | Wall-following (1 m) | Wall-following (2 m) | Linear path traversal | Hallway Navigation | Corner navigation | |
| sUAS Platform A | 0.19 m | 0.07 m | 0.62 m | 0.81 m | 0.67 m | 0.20 m |
| | 0.20 m | 0.07 m | 1.01 m | 1.32 m | 1.02 m | 0.13 m |
| | 0.10 m | 0.13 m | 0.83 m | 1.12 m | 0.99 m | 0.01 m |
| | 0.05 m | 0.05 m | 0.94 m | 1.23 m | 0.66 m | 0.21 m |
| | 0.05 m | 0.06 m | 0.96 m | 0.99 m | 0.95 m | 0.09 m |
| **Average** | **0.12 m** | **0.07 m** | **0.872 m** | **0.858 m** | **1.09 m** | **0.128 m** |
| **Standard deviation** | **0.07 m** | **0.03 m** | **0.16 m** | **0.18 m** | **0.20 m** | **0.08 m** |

Aperture success table:

| Success Condition | sUAS A | sUAS B |
|---|---|---|
| A1 | 80% | 100% |
| A2 | 20% | 0% |
| A3 | 0% | 0% |
| B1 | 0% | 0% |

# Navigation Through Apertures

## Purpose

This test method evaluates sUAS capability to maneuver through apertures wherein the boundaries of the space pose transient constraints on horizontal and vertical movement and hazards for potential collision through a series of narrow opening profiles with variable environmental properties.

## Summary of Test Method

The operator commands the sUAS to navigate through an aperture that either exists already in a real-world environment (e.g., a doorway or window in a building) or a fabricated apparatus that matches the relevant dimensions and shapes for each type of aperture. Three types of apertures are defined for navigation tests, each of which require horizontal or vertical traversal through spaces that are horizontally and/or vertically confined: doorway, window, and manhole. Navigation is performed multiple times to establish statistical significance and the associated probability of success and confidence levels based on the number of successes and failures (see the metrics section). For each trial, the sUAS begins from a starting location that requires it to traverse in a direction not parallel to the navigation route through the aperture, which may also require it to turn. Similarly, the end location for each trial also requires the sUAS to traverse in a direction not parallel to the navigation route. More simply, a single trial constitutes the sUAS traversing from the A side of the apparatus to the B side, navigating through the aperture, then traversing back over to the A side. See Figure 1. Descriptions of each type of aperture are as follows (specifications are provided in the apparatus section):

Doorway Navigation: Horizontal navigation through a horizontally narrow aperture.

Window Navigation: Horizontal navigation through a horizontally and vertically narrow aperture.

Manhole Navigation: Vertical navigation through a horizontally narrow aperture, either via ascension (elevating through the manhole between floors) or descension (lowering through the manhole between floors).

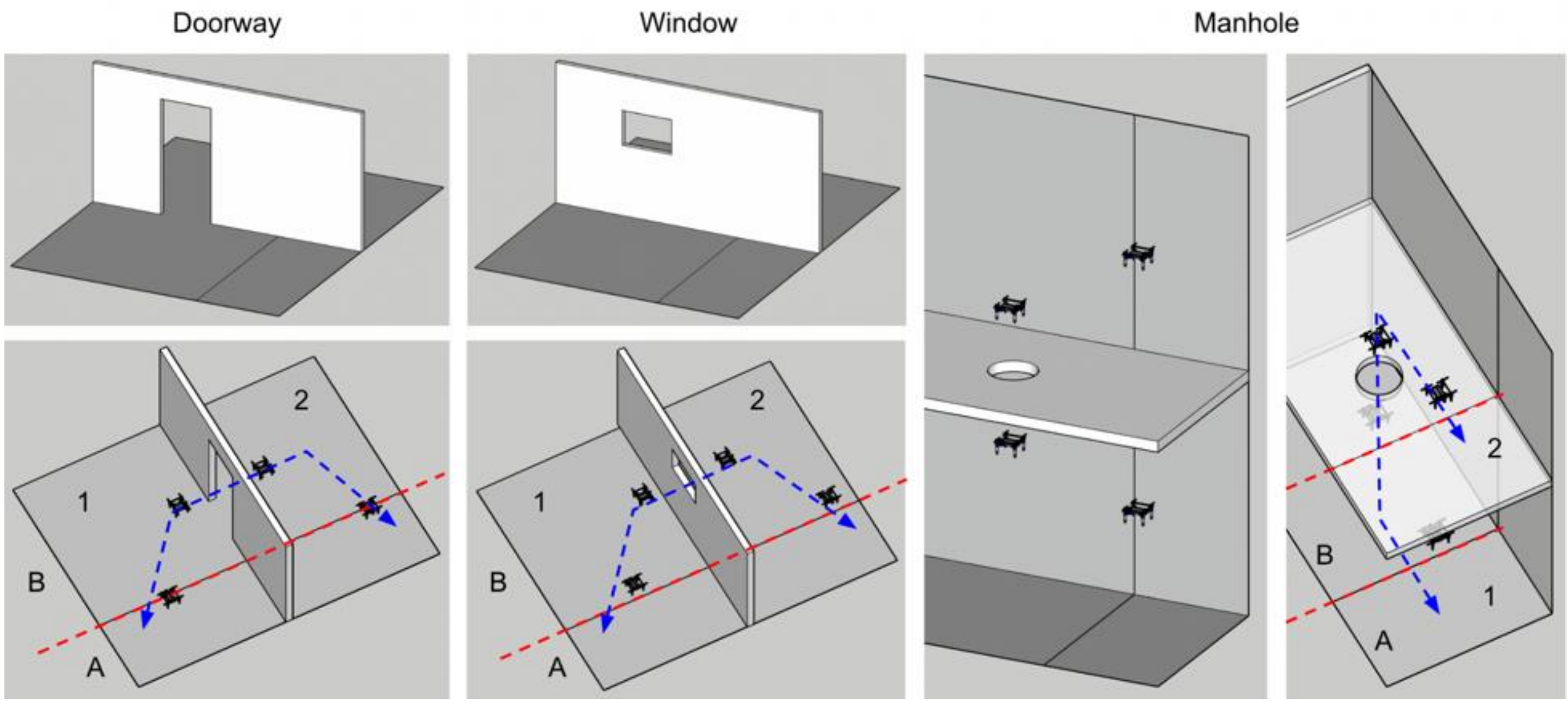

*Figure 1. Each type of aperture navigation test, left to right: doorway, window, and manhole.*

Each navigation test can be run either as elemental or operational navigation:

Elemental Navigation: The operator may maintain line-of-sight with the sUAS such as by following the system with the OCU throughout the environment to maintain communications link, allowing for navigation to be evaluated in as close to an ideal setting as possible.

Operational Navigation: The operator remains at the launch point during execution, unable to maintain line of sight throughout the test. This is similar to an actual operational mission, including all related sUAS communications and operator situation awareness that may arise (e.g., losing comms link at range and/or through obstructions).

Conditions of the environment are also characterized according to lighting and wall and floor surface textures, as these environmental factors are known to impact visual inertial odometry (VIO) capabilities of sUAS and consequently obstacle avoidance, stabilization, etc. The areas outside of the confined space may be individually characterized, identified as 1 and 2 (see Figure 1). Lighting on either side of the aperture is characterized as lighted (100 lux or greater) or dark (less than 1 lux). Several commonly encountered floor and wall surface textures are exemplified as possible textures on either side of the aperture, including concrete, wood, drywall, grass, and cobblestone (see Figure 2). Many apertures in an environment may also exhibit a high-contrast border that may make them easier to distinguish for the VIO of the sUAS. Additionally, each side of the aperture may be located indoors or outdoors, adding further complexity to limit sUAS access to GPS signal or not.

Ideally, the sUAS will not collide with the boundaries of the space (i.e., walls, floor, and ceiling surfaces) while navigating, but contact is allowed so long as it does not cause the sUAS to crash in a way that requires human intervention for it to resume flight.

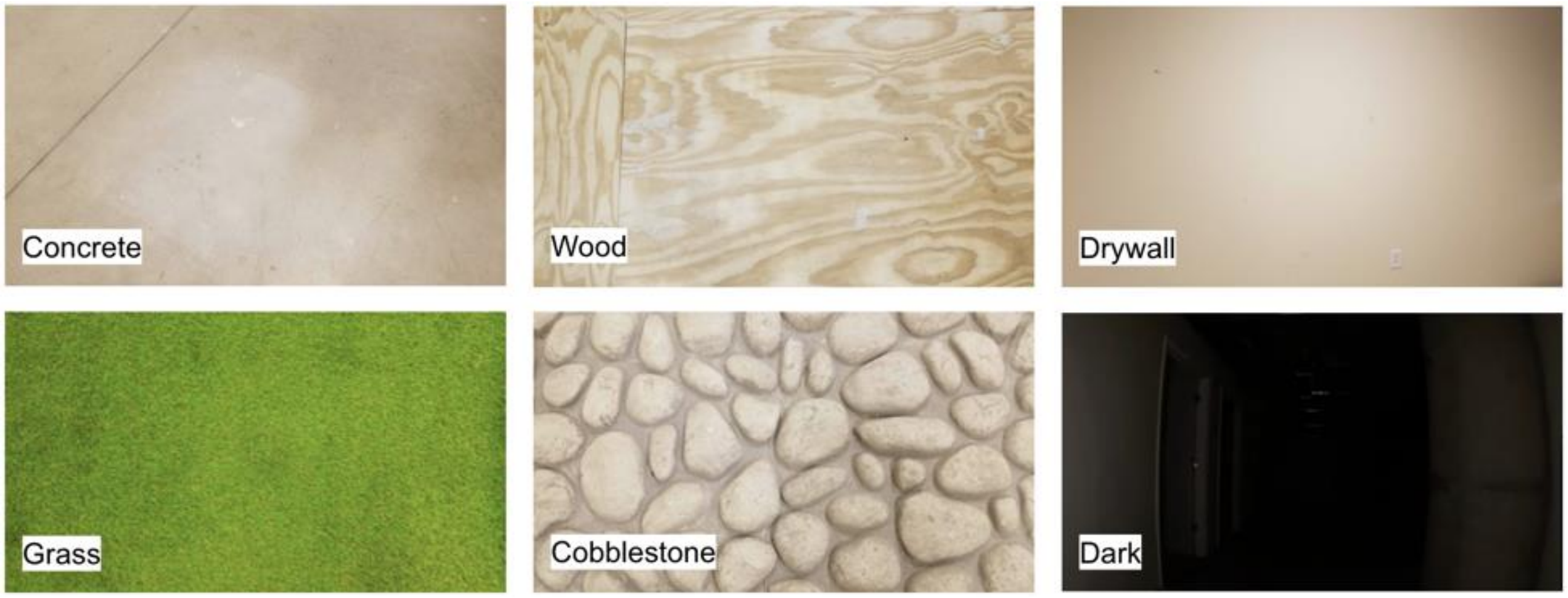

*Figure 2. Examples of surface materials and lighting conditions.*

## Apparatus and Artifacts

Dimensions and specifications of each type of aperture test apparatus are as follows:

Doorway Navigation: A horizontally narrow aperture in a wall measuring between 91 cm (36 in) and 183 cm (72 in) wide by 2 m (80 in) tall (or taller) and 20 cm (8 in) deep (or shallower).

Window Navigation: A horizontally and vertically narrow aperture in a wall measuring between 30 cm (12 in) and 122 cm (48 in) wide by between 30 cm (12 in) and 122 cm (48 in) tall and 20 cm (8 in) deep (or shallower).

Manhole Navigation: A horizontally narrow aperture in a ceiling/floor measuring between 46 cm (18 in) and 122 (48 in) diameter and 20 cm (8 in) deep (or shallower).

The areas on either side of the aperture should measure 3 m (118 in) square or larger to allow for much less obstructed flight than when navigating through the aperture. These areas may contain walls perpendicular to the wall/floor containing the aperture; for example, it is common for doors to be justified to one side of a room. The presence of these obstructions may be problematic for sUAS navigation due to airflow issues when a system flies too close to a wall and/or due to obstacle avoidance functionality (e.g., sUAS may attempt to maintain X distance between it and obstacles for safety, causing it to not be able to navigate through the aperture). If a wall is present on either area outside of the aperture, within 20 cm (8 in) of the edge of the aperture opening, then that area is considered obstructed, producing three possible conditions (see Figure 3):

Unobstructed: No perpendicular walls extend from the wall/floor with aperture located within 1.2 m (48 in) of the edge of the aperture opening.

Partially Obstructed: One perpendicular wall extends from the wall/floor with the aperture in one area outside of the aperture and is located within 20 cm (8 in) of the edge of the aperture opening.

Fully Obstructed: Two perpendicular walls extend from the wall/floor with the aperture in both areas outside of the aperture and are located within 20 cm (8 in) of the edge of the aperture opening.

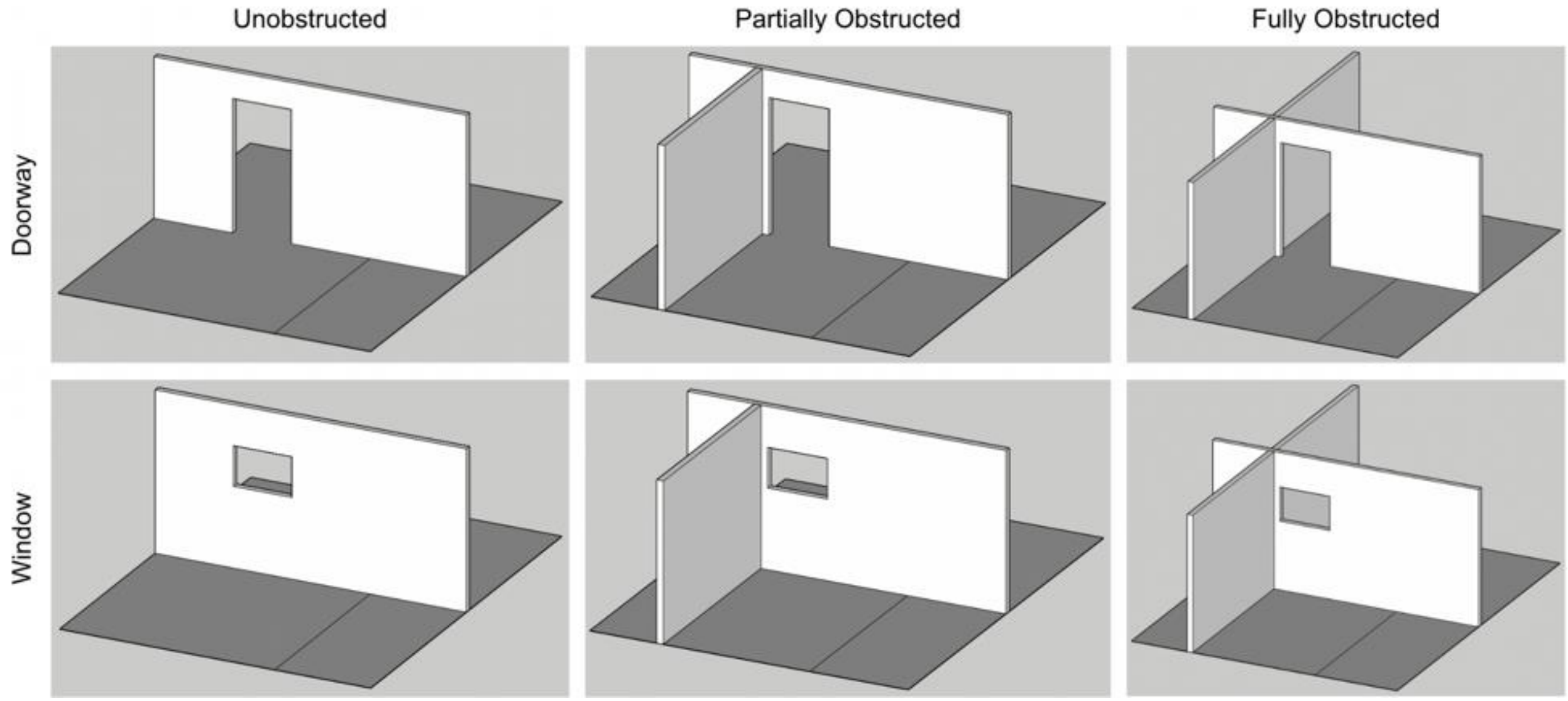

*Figure 3. Variations of the doorway and window aperture navigation tests with and without obstructions.*

Rather than providing exact specifications for each type of surface texture, common examples that are coarsely different from each other in terms of visual color and contrast are provided for subjective classification. Materials used to achieve each surface texture can either be genuine materials (e.g., concrete slab, plywood panel, tile flooring) or a faux material that imitates the same visual qualities (e.g., astroturf for grass, decorative plastic panels that look like cobblestone, printed wallpaper). Any materials used for surface texture should either be heavy enough to not move during sUAS flight or should be held in place using screws, weights, etc. A photo of the aperture should accompany the test data.

## Equipment

A timer is used to record the duration metric.

## Metrics

- Duration: The amount of time to perform the test, starting from when the full body of the sUAS crosses into the B side of the apparatus until it crosses into the A side of the apparatus after completing the number of desired trials.
- Efficacy: Whether or not the sUAS is able to successfully navigate through the confined space, signified by crossing from the A side to the B side, through the confined space, and then back over to the A side on the opposite end of the apparatus.
- Completion: The number of successful trials divided by the total number attempted, reported as a percentage. Each number of successful trials represents an associated probability of success and confidence level, for example:
    - 10 successful trials with no failures for 85% probability of success with 80% confidence
    - 5 successful trials with no failures for 70% probability of success with 80% confidence
    - 10 successful trials with 1 failure for 75% probability of success with 85% confidence
    - Etc. (see Leber et al. [2019] for more details on the statistics associated with number of successful trials and acceptable failures)

- Average navigation speed: The total approximate length traversed across all repetitions divided by the duration of the test, reported in meters per second (m/s).
- Collisions: Whether or not the sUAS collided with the apparatus boundaries (e.g., overhead or lateral obstructions).

## Procedure

1. Select the type of aperture navigation that will be evaluated (doorway, window, manhole), the lighting on either side of the aperture (lighted, dark), the wall and floor surface textures on either side of the aperture (e.g., concrete, wood, drywall, grass, cobblestone), and if each side of the aperture are outdoors or indoors.
    a. If a real-world environment is to be used, record the dimensions, measure the lighting on either side of the aperture, and photograph the wall and floor surface textures. If the opening does not match the dimensional specifications of any of the provided apertures, select one or more that most accurately represent it.
    b. If an apparatus is to be fabricated, build the environment following the dimensional specifications provided in the apparatus section, measure the lighting either side of the aperture, and photograph the wall and floor surface textures.
2. Launch sUAS on the A side of the apparatus.
3. Command the sUAS to navigate to the B side of the apparatus. Once the full body of the sUAS crosses into the B side of the apparatus, start the timer.
4. Command the sUAS to navigate through the aperture to the opposite end of the apparatus. If the sUAS crashes in a way that requires human intervention for it to resume flight, then that trial is considered a failure.
5. Command the sUAS navigate to side A of the apparatus, signifying the successful completion of a trial.
6. Repeat steps 3-5 until the desired number of successful trials has been achieved. Once the full body of the sUAS crosses into the A side of the apparatus on the final successful trial, stop the timer.
7. Calculate the metrics.

## Example Data

- Environment characterization: 91 x 201 x 20 cm (36 x 80 x 8 in)

| Aperture type | Outside aperture, area 1 | | | | | Outside aperture, area 2 | | | | |
|---|---|---|---|---|---|---|---|---|---|---|
| | **Lighting** | **Walls** | **Floor** | **Indoor/ outdoor** | **Obstructed** | **Lighting** | **Walls** | **Floor** | **Indoor/ outdoor** | **Obstructed** |
| Doorway, fully obstructed | Lighted | Drywall | Concrete | Indoor | Yes | Lighted | Drywall | Concrete | Indoor | Yes |
| | 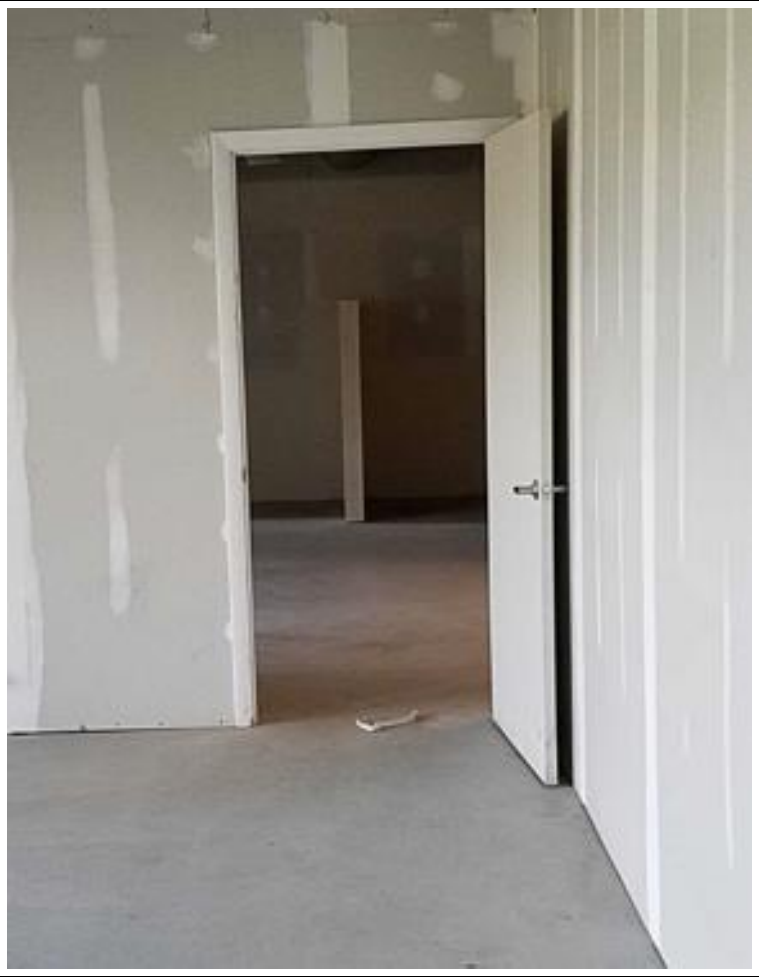 | | | | | 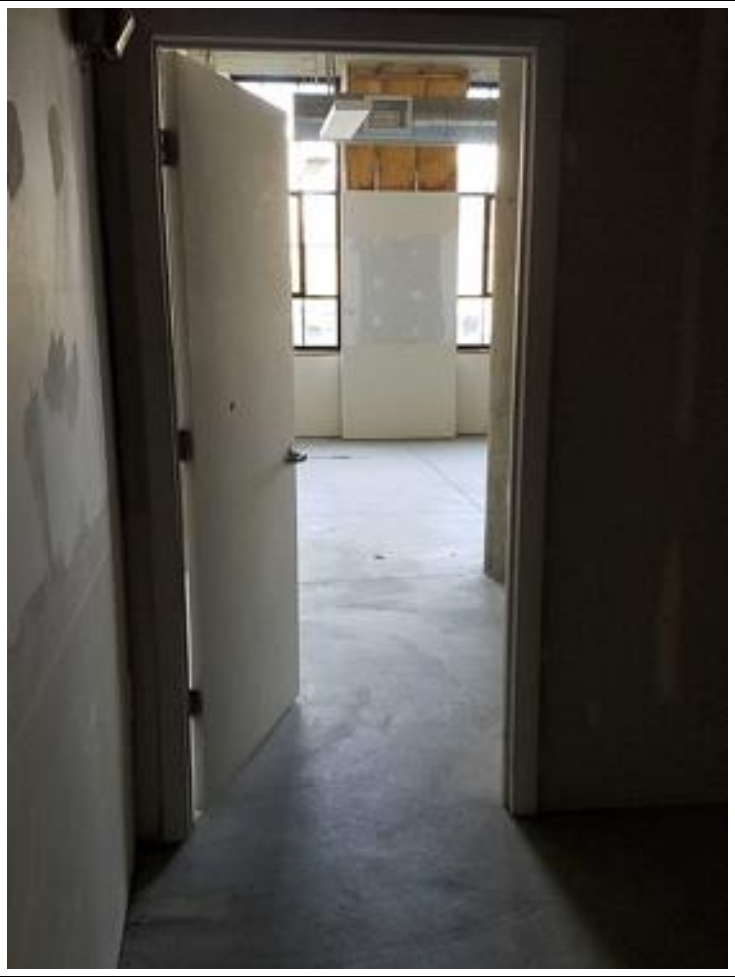 | | | | |

- Performance data: Operational navigation

| sUAS | Metrics | Trials, area 1 -> area 2 | | | | | Trials, area 2 -> area 1 | | | | |
|---|---|---|---|---|---|---|---|---|---|---|---|
| | | **1** | **2** | **3** | **4** | **5** | **1** | **2** | **3** | **4** | **5** |
| A | Efficacy | ✓ | ✓ | ✓ | ✓ | ✓ | ✓ | ✓ | ✓ | ✓ | ✓ |
| | Collisions | | 1 | | 2 | | | 2 | 3 | | 1 |
| | Completion | 100% | | | | | 100% | | | | |
| | Duration | 5 min | | | | | 5 min | | | | |
| | Average navigation speed | 0.13 m/s | | | | | 0.13 m/s | | | | |
| B | Efficacy | ✓ | ✓ | ✓ | X | X | ✓ | ✓ | ✓ | ✓ | X |
| | Collisions | | | | 1 | 1 | | | | | 1 |
| | Completion | 60% | | | | | 80% | | | | |
| | Duration | 8 min | | | | | 7 min | | | | |
| | Average navigation speed | 0.83 m/s | | | | | 0.10 m/s | | | | |

# Navigation Through Confined Spaces

## Purpose

This test method evaluates sUAS capability to maneuver through environments wherein the boundaries of the space pose continuous constraints on horizontal and/or vertical movement and hazards for potential collision through a series of confined space profiles with variable environmental properties.

## Summary of Test Method

The operator commands the sUAS to navigate through a confined space that either exists already in a real-world environment (e.g., a hallway or stairwell in a building) or a fabricated apparatus that matches the relevant dimensions and shapes for each type of confined space. Four types of confined spaces are defined for navigation tests, each of which require horizontal and/or vertical traversal through spaces that are horizontally and/or vertically confined: hallway, tunnel, stairwell/incline, and shaft. Navigation is performed multiple times to establish statistical significance and the associated probability of success and confidence levels based on the number of successes and failures (see the metrics section). For each trial, the sUAS begins from a starting location that requires it to traverse in a direction not parallel to the navigation route through the confined space, which may also require it to turn. Similarly, the end location for each trial also requires the sUAS to traverse in a direction not parallel to the navigation route. More simply, a single trial constitutes the sUAS traversing from the A side of the apparatus to the B side, navigating through the confined space, then traversing back over to the A side. See Figure 1. Descriptions of each type of confined space are as follows (specifications are provided in the apparatus section):

Hallway Navigation: Horizontal navigation through an enclosed horizontally narrow space.

Tunnel Navigation: Horizontal navigation through an enclosed horizontally and vertically narrow space.

Stairwell/Incline Navigation: Horizontal and vertical navigation through an enclosed diagonally narrow space (i.e., horizontally and vertically narrow), either via ascension (elevating throughout the stairwell between the 1st and 2nd floor) or descension (lowering throughout the stairwell between the 2nd and 1st floor).

Shaft Navigation: Vertical navigation through an enclosed horizontally narrow space, either via ascension (elevating throughout the shaft between the 1st and 2nd floor) or descension (lowering throughout the shaft between the 2nd and 1st floor).

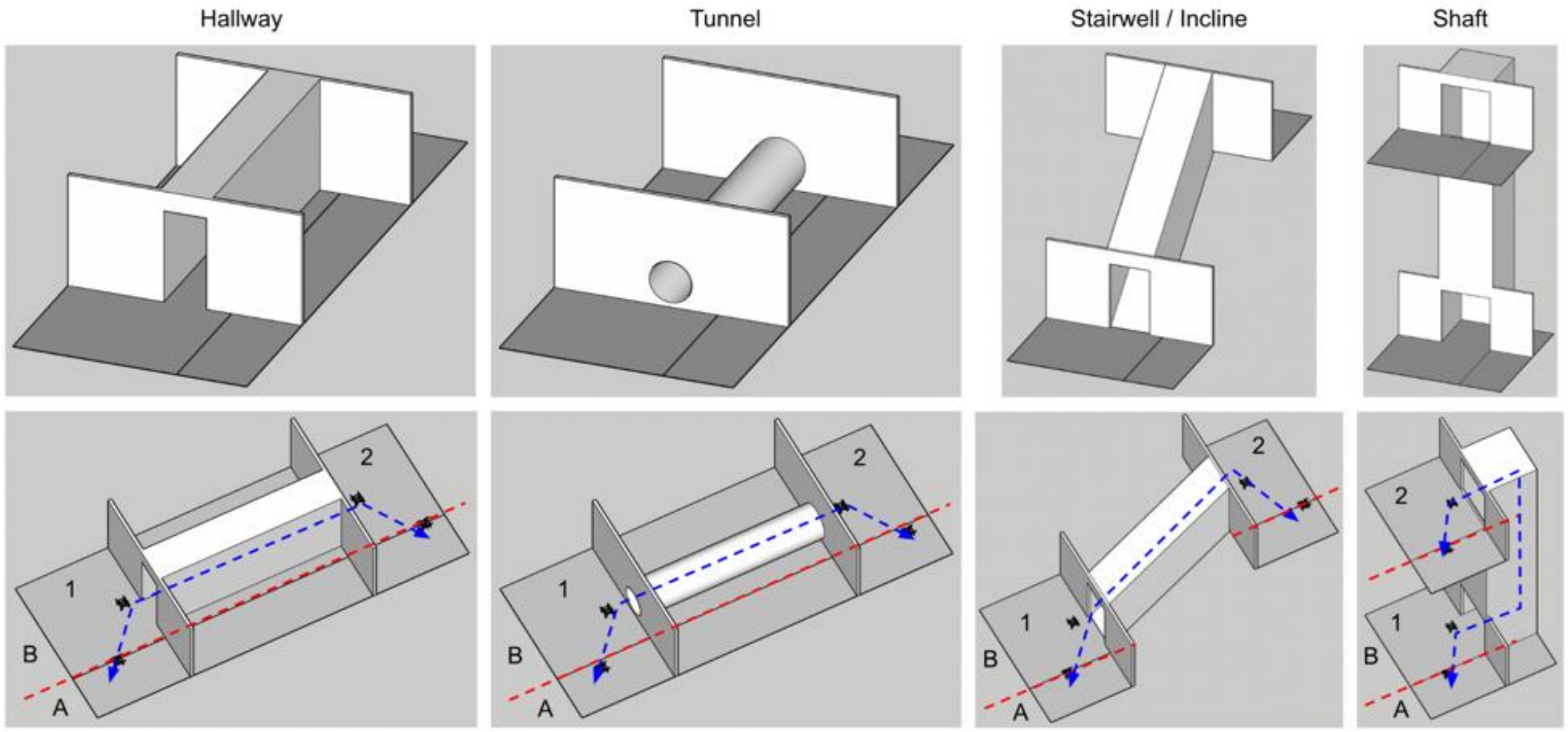

*Figure 1. Each type of confined space navigation test, left to right: hallway, tunnel, stairwell/incline, and shaft.*

Each navigation test can be run either as elemental or operational navigation:

Elemental Navigation: The operator may maintain line-of-sight with the sUAS such as by following the system with the OCU throughout the environment to maintain communications link, allowing for navigation to be evaluated in as close to an ideal setting as possible. Note: this may not be possible for the entirety of shaft navigation.

Operational Navigation: The operator remains at the launch point during execution, unable to maintain line of sight throughout the test. This is similar to an actual operational mission, including all related sUAS communications and operator situation awareness that may arise (e.g., losing comms link at range and/or through obstructions).

Conditions of the environment are also characterized according to lighting and wall and floor surface textures, as these environmental factors are known to impact visual inertial odometry (VIO) capabilities of sUAS and consequently obstacle avoidance, stabilization, etc. The areas outside of the confined space may be individually characterized, identified as 1 and 2 (see Figure 1). Lighting inside and outside of the confined space are characterized as lighted (100 lux or greater) or dark (less than 1 lux). Several commonly encountered floor and wall surface textures are exemplified as possible textures inside and outside of the confined space, including concrete, wood, drywall, grass, and cobblestone (see Figure 2). Additionally, outside of each confined space may be located indoors or outdoors on either end, adding further complexity to limit sUAS access to GPS signal or not.

Ideally, the sUAS will not collide with the boundaries of the space (i.e., walls, floor, and ceiling surfaces) while navigating, but contact is allowed so long as it does not cause the sUAS to crash in a way that requires human intervention for it to resume flight.

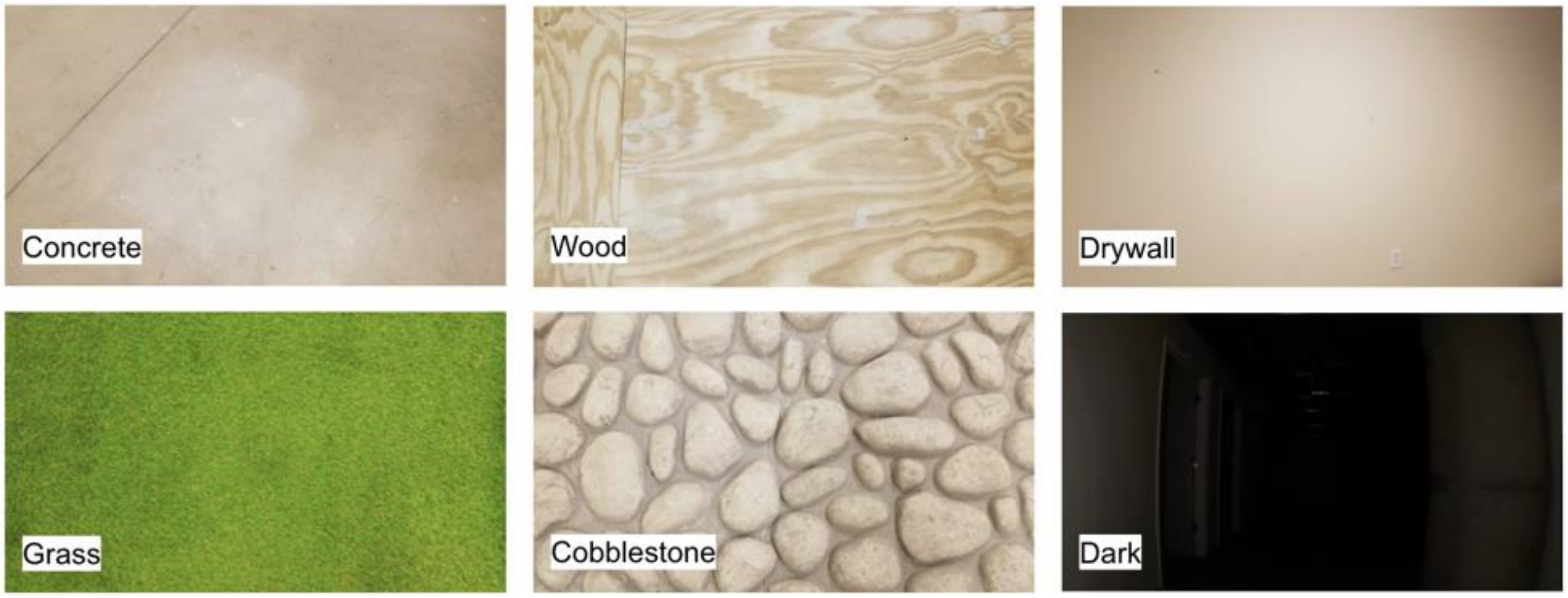

*Figure 2. Examples of surface materials and lighting conditions.*

## Apparatus and Artifacts

Dimensions and specifications of each type of confined space test apparatus are as follows:

Hallway Navigation: A horizontally narrow space measuring between 91 cm (36 in) and 183 cm (72 in) wide by 2 m (80 in) tall (or taller) and 6 m (20 ft) long (or longer).

Tunnel Navigation: A horizontally and vertically narrow space (cylindrical or rectangular) measuring between 91 cm (36 in) and 183 cm (72 in) wide (diameter or square) and 6 m (20 ft) long (or longer). The elevation of the bottom boundary of the tunnel can vary (e.g., coincident with the floor plane or elevated above; see Figure 3.

Stairwell/Incline Navigation: A diagonally narrow space (i.e., horizontally and vertically narrow) measuring between 91 cm (36 in) and 183 cm (72 in) wide by 2.4 m (96 in) tall (or taller) and 6 m (20 ft) long (or longer) on an incline between 30-45 degrees. Two variations of the stairwell are specified (see Figure 3):

- Obstructed Stairwell/Incline: The ceiling plane is similarly inclined to the floor (either a flat surface or inversely stepped like the stairs) resulting in continuous vertically confined space as the sUAS navigates through the stairwell measuring 2.4 m (96 in) tall (or taller) consistently throughout.
- Unobstructed Stairwell/Incline: The ceiling plane is parallel to the floor (i.e., not inclined like the stairs) resulting in a vertically confined space that narrows as the sUAS ascends the stairwell, measuring 2.4 m (96 in) tall (or taller) at the narrowest point.

Shaft Navigation: A horizontally narrow space measuring between 91 cm (36 in) and 183 cm (72 in) square and 6 m (20 ft) tall (or taller).

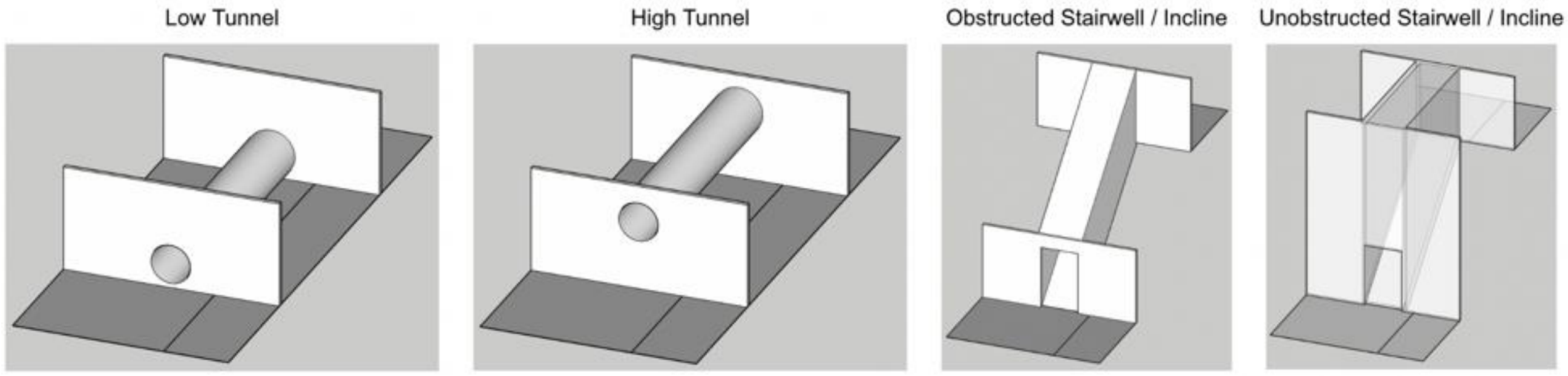

*Figure 3. Variations of the tunnel and stairwell/incline confined space navigation tests.*

The areas outside of the confined space should measure 3 m (118 in) square or larger to allow for much less obstructed flight than when navigating through the confined space. These areas may contain walls perpendicular to the wall/floor outside of the confined space; for example, it is common for stairwells to be justified to one side of an area within a building. The presence of these obstructions may be problematic for sUAS navigation due to airflow issues when a system flies too close to a wall and/or due to obstacle avoidance functionality (e.g., sUAS may attempt to maintain X distance between it and obstacles for safety, causing it to not be able to navigate through the confined space). If a wall is present on either area outside of the confined space, within 20 cm (8 in) of the edge of the opening to the confined space, then that area is considered obstructed, producing three possible conditions (see Figure 3):

Unobstructed: No perpendicular walls extend from the wall/floor outside of the confined space located within 1.2 m (48 in) of the edge of the confined space opening.

Partially Obstructed: One perpendicular wall extends from the wall/floor in one area outside of the confined space and is located within 20 cm (8 in) of the edge of the confined space opening.

Fully Obstructed: Two perpendicular walls extend from the walls/floors in both areas outside of the confined space and are located within 20 cm (8 in) of the edge of the confined space opening.

Rather than providing exact specifications for each type of surface texture, common examples that are coarsely different from each other in terms of visual color and contrast are provided for subjective classification. Materials used to achieve each surface texture can either be genuine materials (e.g., concrete slab, plywood panel, tile flooring) or a faux material that imitates the same visual qualities (e.g., astroturf for grass, decorative plastic panels that look like cobblestone, printed wallpaper). Any materials used for surface texture should either be heavy enough to not move during sUAS flight or should be held in place using screws, weights, etc.

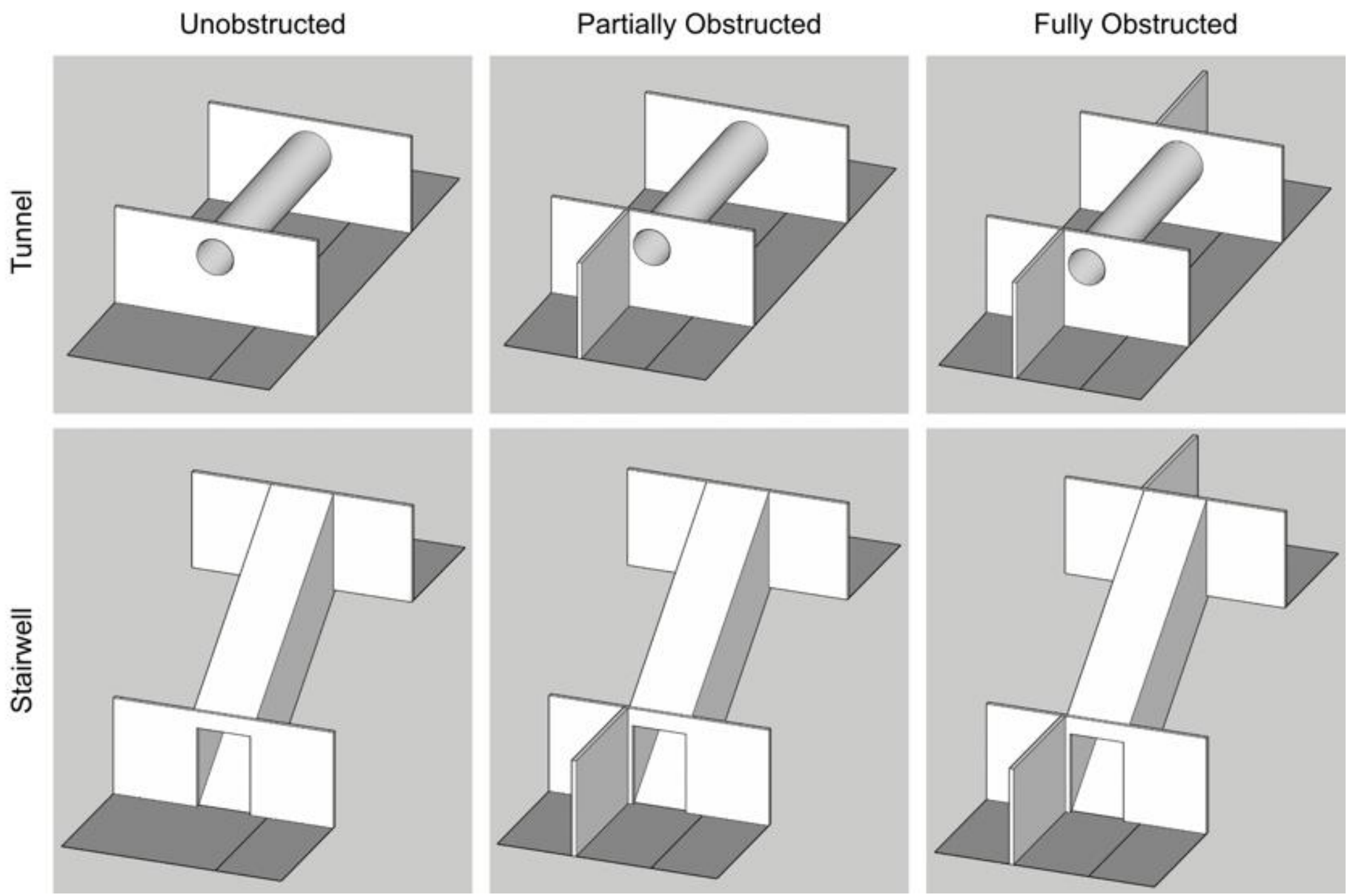

*Figure 4. Variations of the tunnel and stairwell navigation tests with and without obstructions.*

## Equipment

A timer is used to record the duration metric.

## Metrics

- Duration: The amount of time to perform the test, starting from when the full body of the sUAS crosses into the B side of the apparatus until it crosses into the A side of the apparatus after completing the number of desired trials, reported in minutes.
- Efficacy: Whether or not the sUAS is able to successfully navigate through the confined space, signified by crossing from the A side to the B side, through the confined space, and then back over to the A side on the opposite end of the apparatus.
- Completion: The number of successful trials divided by the total number attempted, reported as a percentage. Each number of successful trials represents an associated probability of success and confidence level, for example:
  - 10 successful trials with no failures for 85% probability of success with 80% confidence
  - 5 successful trials with no failures for 70% probability of success with 80% confidence
  - 10 successful trials with 1 failure for 75% probability of success with 85% confidence
  - Etc. (see Leber et al. [2019] for more details on the statistics associated with number of successful trials and acceptable failures)
- Average navigation speed: The total approximate length traversed across all repetitions divided by the duration of the test, reported in meters per second (m/s).
- Collisions: Whether or not the sUAS collided with the apparatus boundaries (e.g., overhead or lateral obstructions).

## Procedure

1. Select the type of confined space navigation that will be evaluated (hallway, tunnel, stairwell/incline, shaft), the lighting inside and outside of the confined space (lighted, dark), the wall and floor surface textures inside and outside of the confined space (e.g., concrete, wood, drywall, grass, cobblestone), and if the areas outside of the confined space are outdoors or indoors.
    a. If a real-world environment is to be used, record the dimensions, measure the lighting on the inside and outside of the confined space, and photograph the wall and floor surface textures. If the space does not match the dimensional specifications of any of the provided confined spaces, select one or more that most accurately represent it.
    b. If an apparatus is to be fabricated, build the environment following the dimensional specifications provided in the apparatus section, measure the lighting on the inside and outside of the confined space, and photograph the wall and floor surface textures.
2. Launch sUAS on the A side of the apparatus.
3. Command the sUAS to navigate to the B side of the apparatus. Once the full body of the sUAS crosses into the B side of the apparatus, start the timer.
4. Command the sUAS to navigate through the confined space to the opposite end of the apparatus. If the sUAS crashes in a way that requires human intervention for it to resume flight, then that trial is considered a failure.
5. Command the sUAS navigate to side A of the apparatus, signifying the successful completion of a trial.
6. Repeat steps 3-5 until the desired number of successful trials has been achieved. Once the full body of the sUAS crosses into the A side of the apparatus on the final successful trial, stop the timer.
7. Calculate the metrics.

## Example Data

- Environment characterization:

| Confined space type, dimensions | Outside confined space, area 1 | | | | | Inside confined space | | | | Outside confined space, area 2 | | | | |
|---|---|---|---|---|---|---|---|---|---|---|---|---|---|---|
| | **Lighting** | **Walls** | **Floor** | **Obstructed** | **Indoor/ outdoor** | **Lighting** | **Walls** | **Floor** | **Indoor/ outdoor** | **Lighting** | **Walls** | **Floor** | **Obstructed** | **Indoor/ outdoor** |
| Tunnel<br><br>104 cm diameter, 4 cm elevated | Lighted | n/a | Grass | Yes | Outdoor | Dark | Concrete | Concrete | Indoor | Dark | Concrete | Gravel | No | Indoor |
| | 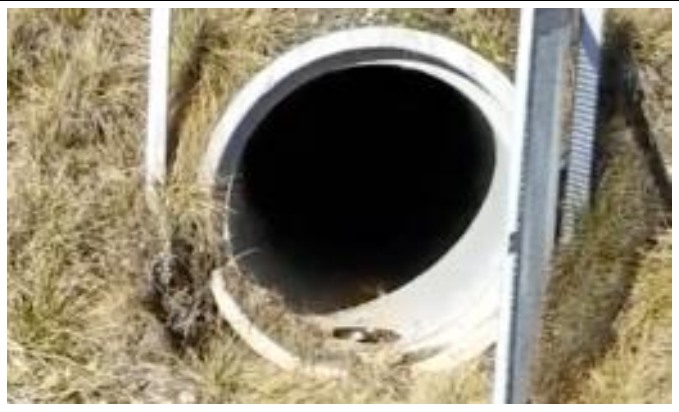 | | | | | 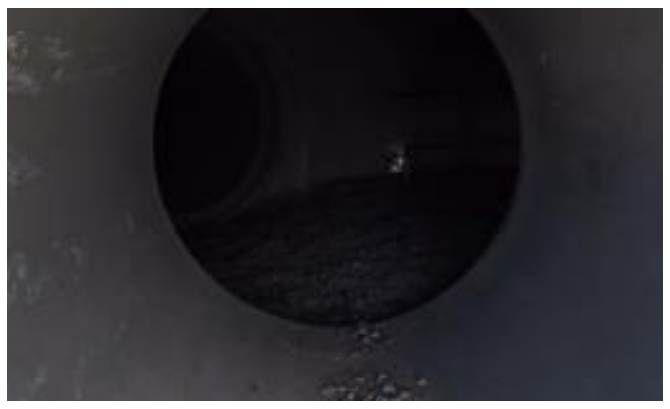 | | | | 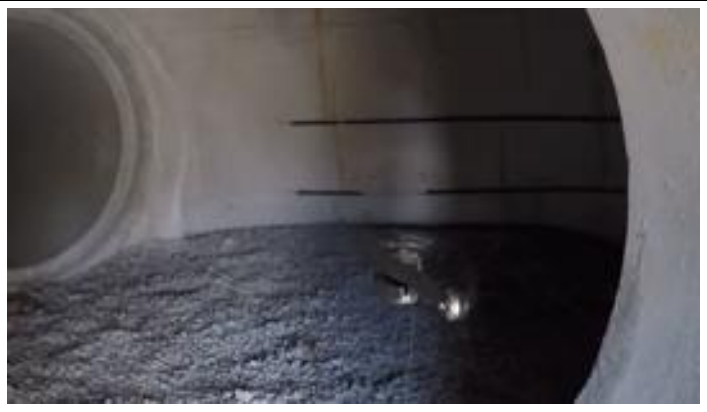 | | | | |

- Performance data: Elemental navigation

| sUAS | Metrics | Trials, area 1 -> area 2 | | | | | Trials, area 2 -> area 1 | | | | |
|---|---|---|---|---|---|---|---|---|---|---|---|
| | | **1** | **2** | **3** | **4** | **5** | **1** | **2** | **3** | **4** | **5** |
| A | Efficacy | ✓ | ✓ | ✓ | ✓ | ✓ | ✓ | ✓ | ✓ | ✓ | ✓ |
| | Collisions | | 1 | | 2 | | | 2 | 3 | | 1 |
| | Completion | 100% | | | | | 100% | | | | |
| | Duration | 7 min | | | | | 8 min | | | | |
| | Average navigation speed | 0.15 m/s | | | | | 0.14 m/s | | | | |
| B | Efficacy | ✓ | ✓ | ✓ | X | X | ✓ | ✓ | ✓ | ✓ | X |
| | Collisions | | | | 1 | 1 | | | | | 1 |
| | Completion | 60% | | | | | 80% | | | | |
| | Duration | 15 min | | | | | 13 min | | | | |
| | Average navigation speed | 0.07 m/s | | | | | 0.08 m/s | | | | |

# Mapping

These test methods are used to measure the elemental 2D topology, 3D shapes, and/or photographic features of the environment that can be detected and mapped by the sUAS. Competencies evaluated include those related to distance sensing, dead reckoning/odometry, mapping resolution, map generation (onboard or offboard), camera resolution, and image stitching. The results of these tests can be used to evaluate the ability of sUAS to map environments of certain dimensions, features, and complexity.

## Indoor Mapping Resolution

Affiliated publications: [Norton et al., 2021]

### Purpose

This test method is used to evaluate (1) the standoff distances required to map interior spaces and at what level of accuracy and visual acuity can be expected at each distance, and (2) the ability to accurately map features in the environment.

### Summary of Test Method

This test method is comprised of two separate tests for evaluating indoor mapping resolution:

Interior Boundaries: The operator maneuvers the sUAS along a specified approximate trajectory through a standard set of clearances which define a consistent standoff distance between the centroid of the sUAS and the walls and floor. While moving along the trajectory, the sUAS maps the environment while tilting and panning its mapping camera(s) and sensors as needed at each defined point, while maintaining a forward orientation (i.e., not yawing in place). Each point is navigated to sequentially and then back out in reserve order. If the sUAS is not able to reach a point due to confined space restrictions, it may reverse at any point in the sequence. The test can be run in lighted (100 lux or greater) or dark (less than 1 lux) conditions.

Shape Accuracy: The operator maneuvers the sUAS around a single split-cylinder fiducial mounted on two sides of a wall in order to map it, following manufacturer recommendations on effective flight and camera maneuvering techniques for mapping. There are no restrictions on the flight path taken. The collected data is downloaded to generate a map. The test can be run in lighted (100 lux or greater) or dark (less than 1 lux) conditions.

These tests should be run as a prerequisite to the Indoor Map Accuracy test.

### Apparatus and Artifacts

Interior Boundaries: A series of walls, elevated platforms, and ceiling panels are positioned in the space as shown in Figure 1. Colored tape is laid out to mark the approximate trajectory of the sUAS and along all edges where two intersecting planes meet (e.g., floor and wall, wall and ceiling). Visual acuity targets are mounted on the walls, floors, and ceilings. The trajectories taken by the sUAS are between two points in space, labeled 0 through 4. Each point-to-point trajectory corresponds to a consistent standoff distance from all surfaces, as follows:

- 0-1: 180 cm (72 in) on left, right, front, and below
- 1-2: 120 cm (48 in) on left, right, front, below, and above
- 2-3: 60 cm (24 in) on left, right, front, below, and above
- 3-4: 30 cm (12 in) on left, right, front, below, and above

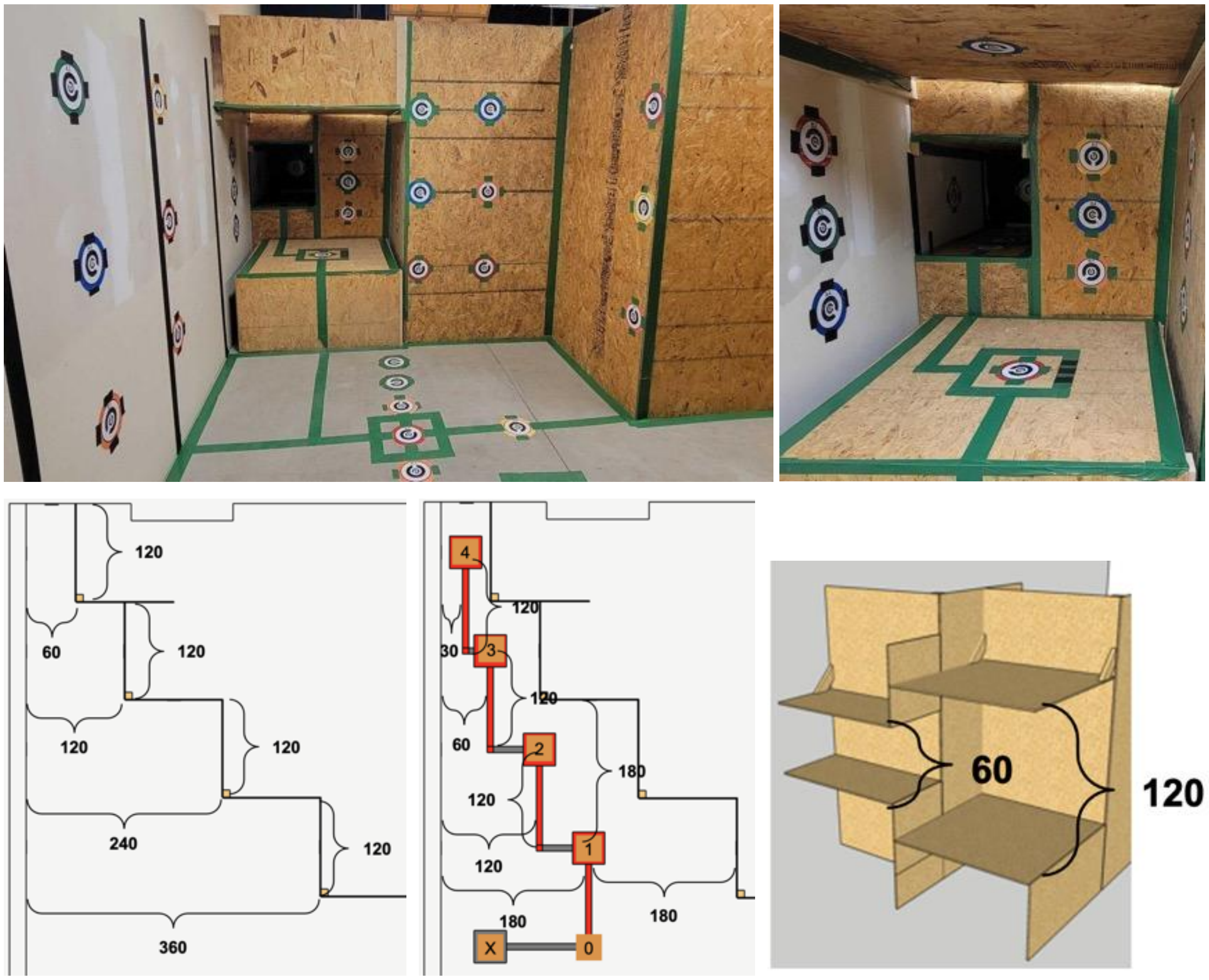

*Figure 1. Indoor Mapping Resolution: Interior Boundaries apparatus. All measurements are shown in centimeters.*

<u>Shape Accuracy</u>: A single wall with a split-cylinder fiducial mounted on either side of it with sufficient unobstructed space on all sides for the sUAS to comfortably maneuver around it (see Figure 2). The difficulty metrics for the fiducial in this test method are 5 m (16 ft) for minimum traversal distance and 2 turns for minimum orientation changes.

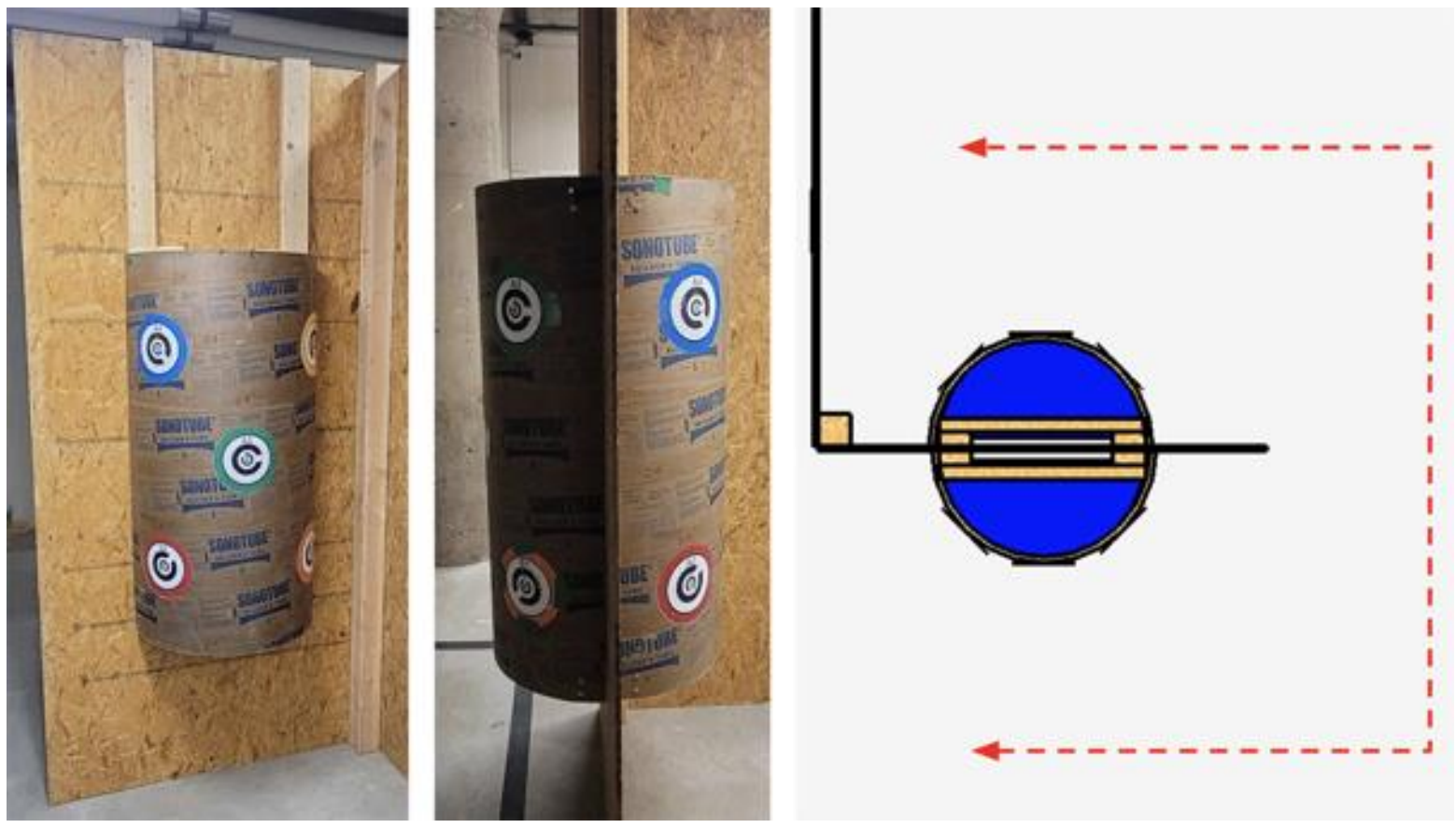

*Figure 2. Apparatus and approximate sUAS trajectory (seen from overhead) for the Indoor Mapping Resolution: Shape Features test method.*

## Equipment

A timer is used to record mapping time and processing time metrics.

## Metrics

- Dimensional accuracy: Dimensions and angles of wall, floor, and ceiling boundaries compared to ground truth. This metric is calculated by dividing the sum of all reported dimensions by the sum of all corresponding ground truth dimensions per standoff distance, reported as a percentage, for average dimensional accuracy.
- Shape accuracy: Alignment and orientation of each half of a split-cylinder fiducial to the other corresponding half, evaluated from distance data generated either through point cloud or photogrammetry data when viewed from overhead. Each pair of halves is evaluated as "complete" (if the two halves form a complete circle), "incomplete" (if a significant portion of each fiducial half is missing), or "shifted" (if both halves are mapped, but are significantly apart from one another).
- Field of view (FOV): Amount of boundaries for the associated wall, floor, or ceiling surface able to be mapped. The boundaries may be measured from point cloud data or photogrammetry data by measuring the visible colored tape lines. This metric is calculated by dividing the number of boundaries that are 50% or more visible by the total number possible per standoff distance, reported as a percentage.
- Visual acuity: Level of detail that can be resolved in the available Landolt C artifacts, evaluated from photogrammetry data, reported in millimeters. The highest level of acuity achieved on a single target is reported per the surface where it is mounted (interior boundaries test) or per each half of the split-cylinder fiducial (shape accuracy test) is reported. These measures are averaged to calculate an average acuity (with standard deviation) per standoff distance.
- Mapping time: Amount of time required to conduct the flight for mapping the apparatus, timed from takeoff to landing, reported in minutes.
- Processing time: Amount of time required to download the mapping data, load it into the mapping software, and produce the map, reported in minutes. This metric is evaluated separately for point cloud and photogrammetry data. The time it takes to perform any manual adjustments made to the map (e.g., scaling) are included in this metric.

## Procedure

Interior Boundaries:

1. Generate ground truth dimensions of the apparatus. Given that the apparatus is built to known dimensions, additional measurement should not be necessary.
2. Launch sUAS to 180 cm elevation, start the timer, and maneuver to point 0.
3. Initiate mapping functionality and follow flight path.
4. After point 1 is reached, descend to approximately 120 cm.
5. At each point, hover sUAS in place, tilting and panning camera(s) as needed while maintaining forward orientation.
6. Navigate to each point sequentially and then back out in reverse order and orientation.
7. Once complete, maneuver the sUAS back to the launch point, land, and stop the timer. Record the mapping time metric.
8. Start the timer again and then download, process, and generate the map.
9. Once the map is generated and ready for evaluation, stop the timer. Record the processing time metric.
10. Calculate the metrics in software (e.g., CloudCompare for point clouds, Pix4Dmapper for photogrammetry).

Shape Accuracy:

1. Generate ground truth dimensions of the apparatus. Given that the apparatus is built to known dimensions, additional measurement should not be necessary.
2. Launch sUAS to desired elevation and start the timer.
3. Initiate mapping functionality and maneuver around both sides of the fiducial to map it.
4. Once complete, land the sUAS and stop the timer. Record the mapping time metric.
5. Start the timer again and then download, process, and generate the map.
6. Once the map is generated and ready for evaluation, stop the timer. Record the processing time metric.

7. Calculate the metrics in software (e.g., CloudCompare for point clouds, Pix4Dmapper for photogrammetry).

## Example Data

<u>Interior Boundaries</u>:

- Environment characterization: Dark
- Performance data:

| | | Right | | | Left | | | Front | | | Below | | | Above | | | Average | | |
|---|---|---|---|---|---|---|---|---|---|---|---|---|---|---|---|---|---|---|---|
| **sUAS** | **Stand off (cm)** | **Dim Acc (%)** | **FOV (%)** | **Acuity (mm)** | **Dim Acc (%)** | **FOV (%)** | **Acuity (mm)** | **Dim Acc (%)** | **FOV (%)** | **Acuity (mm)** | **Dim Acc (%)** | **FOV (%)** | **Acuity (mm)** | **Dim Acc (%)** | **FOV (%)** | **Acuity (mm)** | **Dim Acc (%)** | **FOV (%)** | **Acuity (mm)** |
| A | 180 | 100 | 50 | 20 | 100 | 50 | 8 | 100 | 100 | 3 | 100 | 75 | 20 | n/a | n/a | n/a | 100 | 69 (+/- 24) | 13 (+/- 9) |
| | 120 | 100 | 100 | 3 | 100 | 100 | 20 | 100 | 100 | 3 | 100 | 100 | 20 | n/a | n/a | n/a | 100 | 100 | 12 (+/- 10) |
| | 60 | 100 | 100 | 3 | 100 | 100 | 8 | 100 | 100 | 3 | 100 | 100 | 8 | 100 | 100 | 8 | 100 | 100 | 6 (+/- 3) |
| | 30 | 100 | 100 | 8 | 100 | 100 | 3 | 100 | 100 | 8 | 100 | 100 | 3 | 100 | 100 | 8 | 100 | 100 | 6 (+/- 3) |

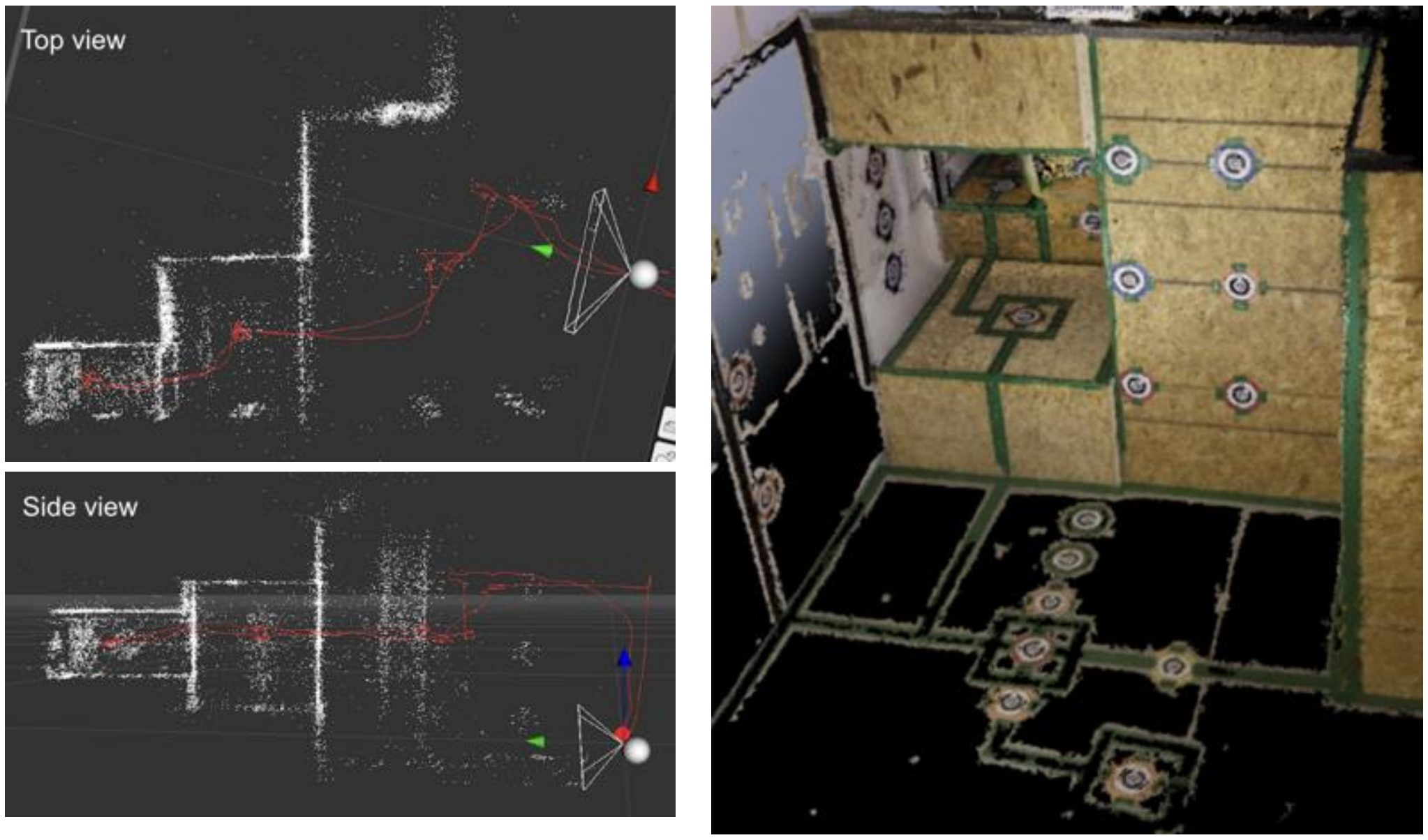

*Figure 3. Example point cloud and photogrammetric maps generated from the Interior Boundaries test.*

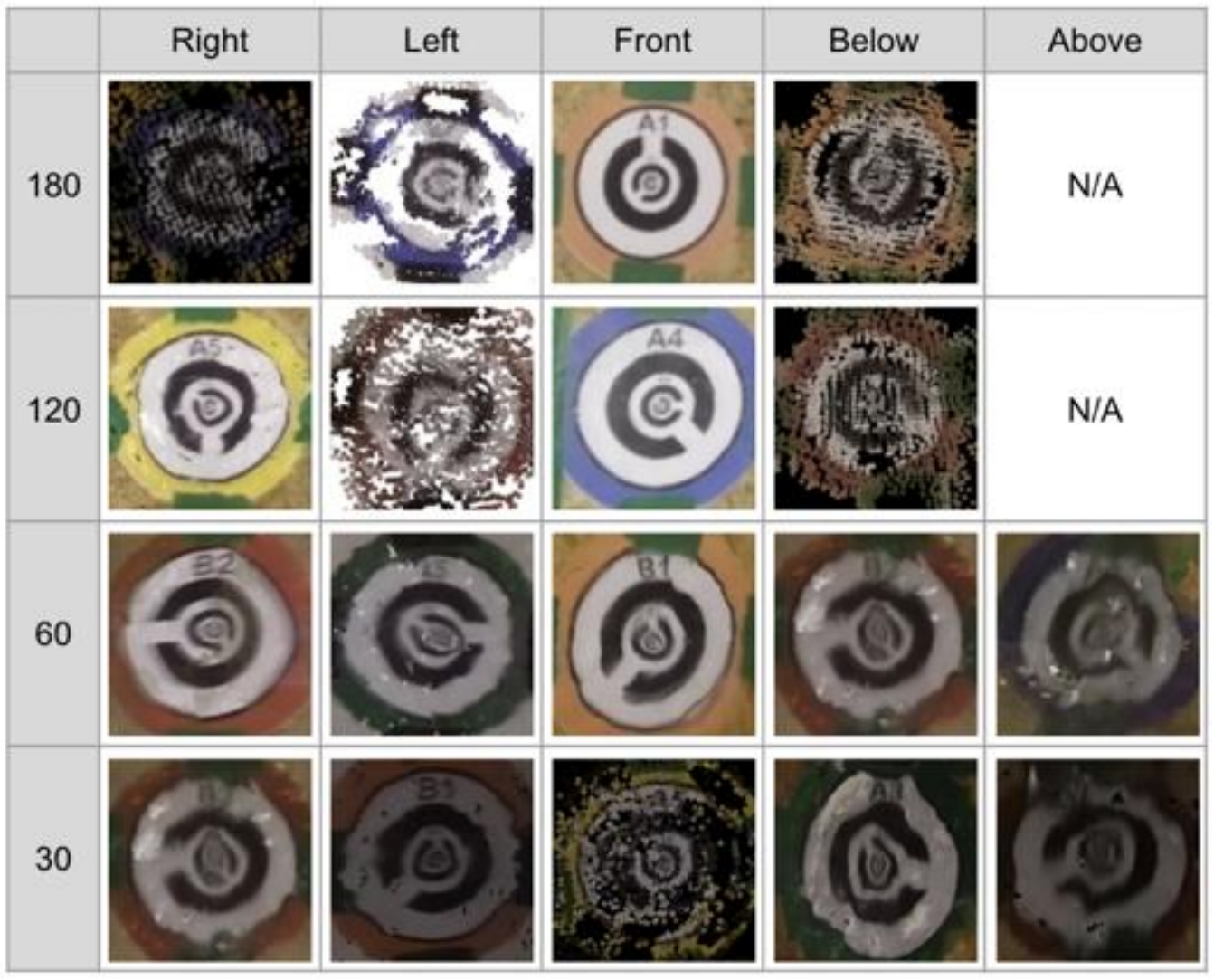

*Figure 4. Example singulated visual acuity targets from photogrammetric maps at each standoff and surface generated from the Interior Boundaries test.*

<u>Shape Accuracy</u>:

- Environment characterization: Dark
- Performance data:

| sUAS | Shape accuracy | Acuity (mm) | Mapping time (min) | Processing time (min) |
|---|---|---|---|---|
| A | Complete | 3 | 7 | 7 (point cloud)<br>120 (photogrammetry) |
| B | Complete | 8 | 5 | 15 (photogrammetry) |

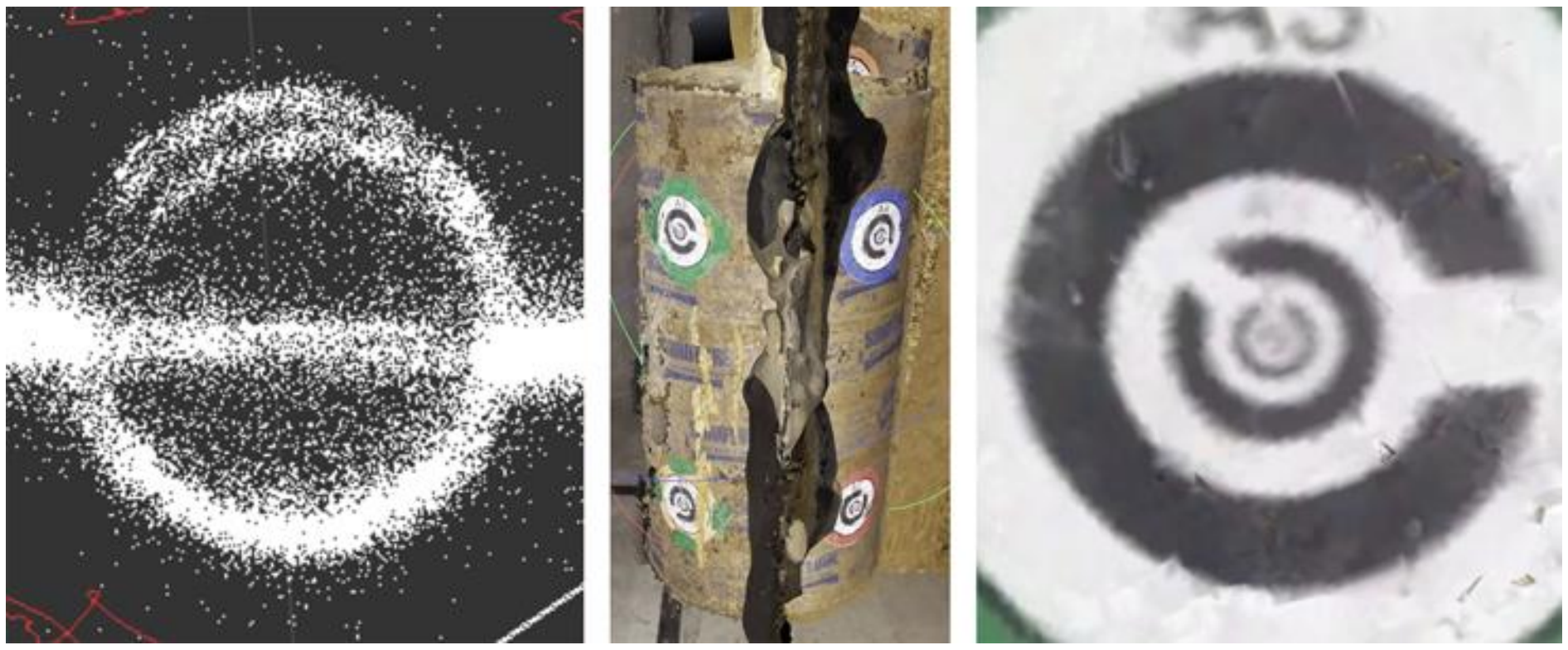

*Figure 5. Example data from the Shape Accuracy test. Left to right: top view of point cloud; side view of photogrammetric map; example visual acuity target in photogrammetric map.*

## Indoor Mapping Accuracy

Affiliated publications: [Norton et al., 2021]

### Purpose

This test method evaluates generated maps of real-world spaces under either ideal conditions (elemental mapping) or in the context of a mission (operational mapping) in a larger environment where drift issues may be exacerbated.

### Summary of Test Method

The operator flies the sUAS through an environment to collect mapping data. A series of split-cylinder fiducials are positioned throughout the environment. An accurate ground truth of the environment must be available in order to compare the sUAS map. The ground truth for comparison is a map of the environment that is generated using a more precise method with high confidence of accuracy (e.g., industrial ground robot, handheld lidar system). An accurate 3D ground truth may be very expensive and/or difficult to generate, whereas a 2D map can be more easily gathered (e.g., architectural layout, dimensional measurement). Multiple flights may be conducted in order to change batteries. The collected data is downloaded to generate a map; if multiple flights were conducted, multiple maps will be generated for each incremental flight (e.g., map of flight 1, map or flights 1+2, etc.) and evaluated separately. For maps generated from multiple flights, evaluations should differentiate between automatic alignment of the individual maps and manual alignment performed by an operator. The test can be run in lighted (100 lux or greater) or dark (less than 1 lux) conditions, as either an elemental or operational test:

Elemental Mapping: The operator may maintain line-of-sight with the sUAS such as by following the system with the OCU throughout the environment to maintain communications link. This allows the system's map generation capability to be evaluated in as close to an ideal setting as possible.

Operational Mapping: The operator remains at the launch point during execution, unable to maintain line of sight throughout the test, without prior knowledge of the layout of the space. This is similar to an actual operational mission, including all related sUAS communications and operator situation awareness that may arise (e.g., losing comms link at range and/or through obstructions, monitoring battery life such that the sUAS can be flown back before it dies). If multiple flights are conducted, the sUAS must be flown back to the launch point where the operator is stationed in order to change batteries.

### Apparatus and Artifacts

A real-world indoor environment is outfitted with split-cylinder fiducials horizontally separated on either side of a wall and/or vertically at different elevations. The environment must be a closed loop (i.e., conducting a wall follow would result in returning back to the start). At least one half of a fiducial should be present in each room and/or floor/landing; this ensures that calculation of the coverage metric more closely represents coverage of each room and/or floor/landing in the environment. Environments can be classified as those that can be largely navigated horizontally (e.g., warehouse floor), vertically (e.g., stairwells), or a combination of both (e.g., multi-level facility). See Figure 1 for examples.

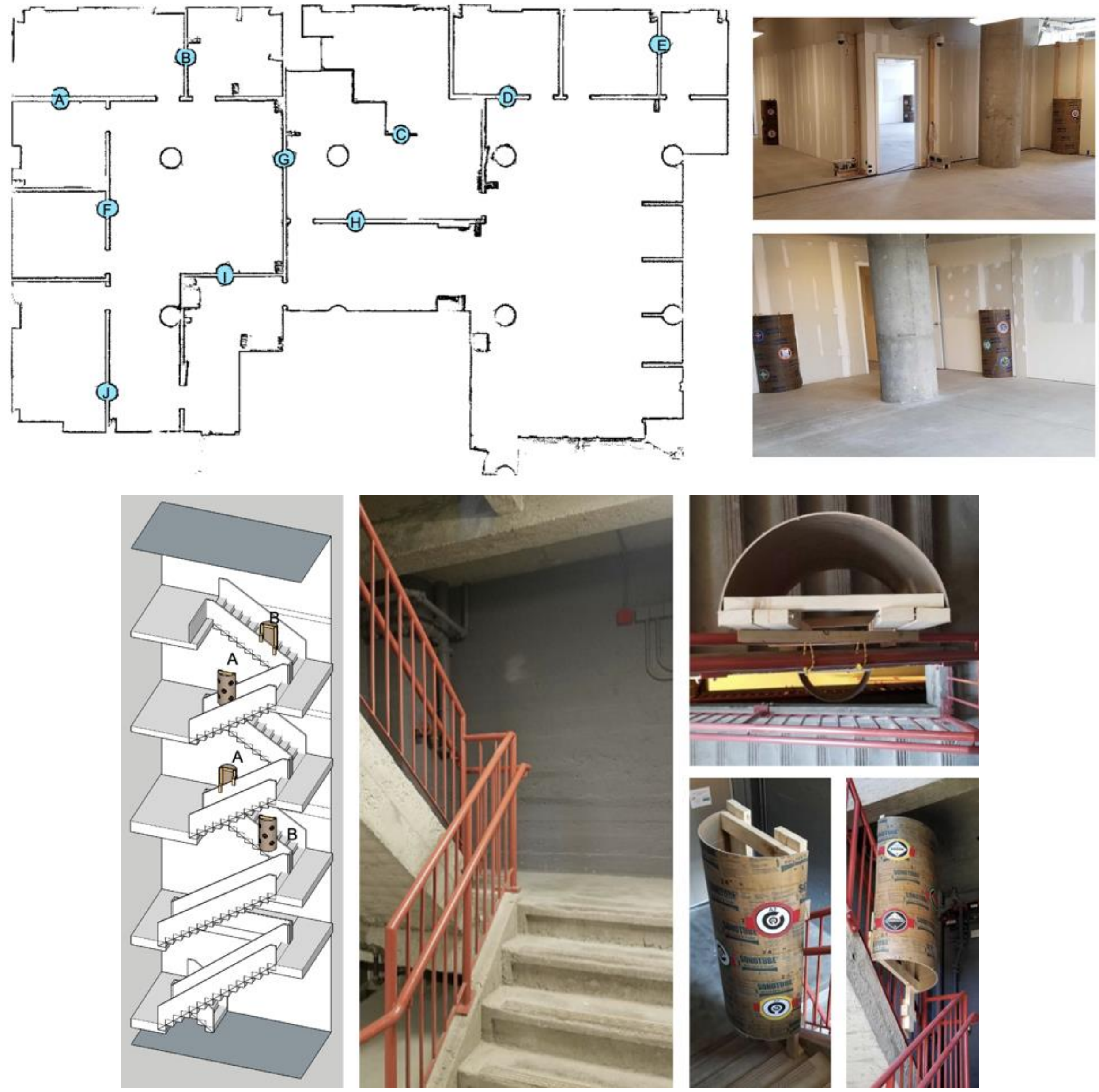

*Figure 1. Example apparatuses for the Indoor Map Accuracy test method. Top: example horizontal navigation environment with 10 fiducials labeled A through J. Bottom: example vertical navigation environment with 2 fiducials labeled A and B.*

## Equipment

A timer is used to record mapping time and processing time metrics.

## Metrics

- Shape accuracy: Alignment and orientation of each half of a split-cylinder fiducial to the other corresponding half, evaluated from distance data generated either through point cloud or photogrammetry data when viewed from overhead. Each pair of halves is evaluated as “complete” (if the two halves form a complete circle), “incomplete” (if a significant portion of each fiducial half is missing), or “shifted” (if both halves are mapped, but are significantly apart from one another). This metric is reported as a percentage by dividing the number of complete fiducials by the total number of fiducials in the environment.

- Visual acuity: Level of detail that can be resolved in the available Landolt C artifacts, evaluated from photogrammetry data, reported in millimeters. The highest level of acuity achieved on a single target across all available per each split-cylinder fiducial is reported. These measures are averaged to calculate an average acuity (with standard deviation) for the entire test.
- Global error: Relative measurements of fiducial positions to one another across the entire map compared to ground truth, averaged across all fiducials, and reported as an average amount of error in centimeters (scaled from pixels). A custom software tool is available that evaluates this metric after manual input of fiducial locations in the ground truth and evaluation map.
- Coverage: Number of fiducial halves mapped (partially or completely) compared to the amount available, reported as a percentage. A custom software tool is available that evaluates this metric after manual input of fiducial locations in the ground truth and evaluation map. A second coverage metric can be calculated using square footage for the area that has been mapped, whereby the square footage of a room is counted if at least a portion of all of its wall boundaries has been mapped.
- Mapping time: Amount of time required to conduct the flight for mapping, timed from takeoff to landing, reported in minutes. If multiple flights are conducted, this metric is calculated as a sum of the individual flights plus the time taken to change batteries in between flights.
- Processing time: Amount of time required to download the mapping data, load it into the mapping software, and produce the map, reported in minutes. This metric is evaluated separately for point cloud and photogrammetry data. The time it takes to perform any manual adjustments made to the map (e.g., scaling, position and orientation adjustments to manually align multiple maps) are included in this metric.
- Number of flights: The number of flights performed to generate the map. If multiple flights are conducted, the method used to merge the maps from each fight into a single map is characterized as either “automatic” (i.e., the map merging software used scales and aligns them automatically) or “manual” (i.e., the operator must scale and align each map to the other manually).

## Procedure

1. Generate ground truth dimensions of the environment and layout of fiducials for comparison.
2. Launch sUAS to desired elevation and start the timer.
3. Initiate mapping functionality and maneuver throughout the space in order to map it.
    a. Elemental mapping test: The operator can follow the sUAS through the space during flight to maintain optimal communications link.
    b. Operational mapping test: The operator must remain at the launch point during execution, unable to maintain line of sight throughout the test.
4. Conduct multiple flights if needed based on battery life. The location and timing of any battery changes should be recorded.
    a. Elemental mapping test: The sUAS can land anywhere in the environment to have its battery changed.
    b. Operational mapping test: The sUAS must be flown back to the launch point where the operator is stationed in order to have its battery changed.
5. Once complete, land the sUAS and stop the timer. Recording the mapping time metric.
6. Start the timer again and then download, process, and generate the map.
7. Once the map is generated and ready for evaluation, stop the timer. Record the processing time metric.
8. Calculate the metrics by observing and measuring the map in software (e.g., CloudCompare for point clouds, Pix4Dmapper for photogrammetry). When evaluating 3D maps, 2D versions of those maps (either single layers or multiple layers compressed together) can be generated and evaluated against 2D ground truths.

## Example Data

- Environment characterization: Using the example horizontal navigation environment in Figure 1, Dark

| Metrics | Fiducials | | | | | | | | | |
|---|---|---|---|---|---|---|---|---|---|---|
| | **A** | **B** | **C** | **D** | **E** | **F** | **G** | **H** | **I** | **J** |
| Minimum traversal distance (m) | 11 | 8 | 35 | 5 | 12 | 7 | 27 | 7 | 16 | 10 |
| Minimum orientation changes (turns) | 2 | 2 | 7 | 2 | 3 | 2 | 5 | 2 | 3 | 2 |
| Difficulty rating | M | L | H | L | M | L | H | L | M | L |

- Performance data: Elemental mapping
  (note: sUAS A does not have photogrammetric mapping capabilities, while sUAS B does)

| sUAS | Flight, Merge type | Metrics | Fiducials | | | | | | | | | | Map |
|---|---|---|---|---|---|---|---|---|---|---|---|---|---|
| | | | **A** | **B** | **C** | **D** | **E** | **F** | **G** | **H** | **I** | **J** | |
| A | 1 | Shape accuracy | C | C | | | | C | I | | S | | 60% |
| | | Visual acuity (mm) | n/a | n/a | n/a | n/a | n/a | n/a | n/a | n/a | n/a | n/a | n/a |
| | | Coverage (%) | ✓ | ✓ | X | X | X | ✓ | // | X | ✓ | ✓ | 55 |
| | | Mapping time (min) | | | | | | | | | | | 9 |
| | | Processing time (min) | | | | | | | | | | | 9 |
| | | Global error (cm) | | | | | | | | | | | 8 |
| | 1+2 Automatic | Shape accuracy | C | C | C | S | C | C | I | C | S | C | 70% |
| | | Visual acuity (mm) | n/a | n/a | n/a | n/a | n/a | n/a | n/a | n/a | n/a | n/a | n/a |
| | | Coverage (%) | ✓ | ✓ | ✓ | ✓ | ✓ | ✓ | ✓ | ✓ | ✓ | ✓ | 100 |
| | | Mapping time (min) | | | | | | | | | | | 18 |
| | | Processing time (min) | | | | | | | | | | | 18 |
| | | Global error (cm) | | | | | | | | | | | 64 |
| B | 1+2+3+4 Manual | Shape accuracy | C | C | C | C | I | I | C | C | | I | 67% |
| | | Visual acuity (mm) | 8 | 8 | 8 | 20 | 8 | 8 | 8 | 8 | | 3 | 8.8 (+/- 4.5) |
| | | Coverage (%) | ✓ | ✓ | ✓ | ✓ | // | // | ✓ | ✓ | X | // | 75 |
| | | Mapping time (min) | | | | | | | | | | | 32 |
| | | Processing time (min) | | | | | | | | | | | 720 |
| | | Global error (cm) | | | | | | | | | | | 5 |

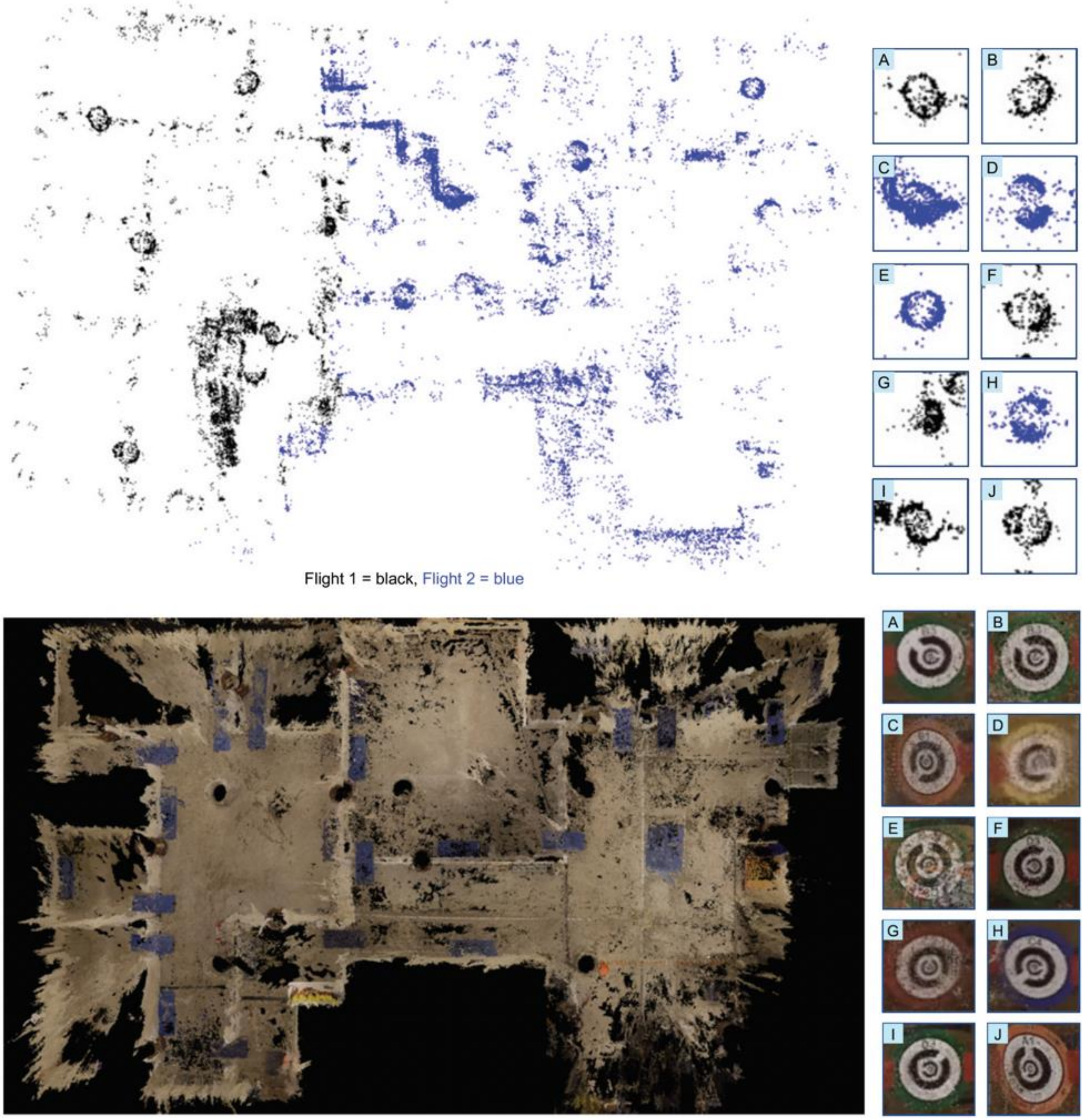

*Figure 2. Example point cloud map with singulated fiducials (top) and photogrammetric map with singulated fiducials (bottom) generated from the Indoor Map Accuracy test.*

# Autonomy

These tests aim to calculate, evaluate, and compare the autonomy score of sUAS in both a general mission-independent and a more specific mission-dependent context. These two test methods are referred to as the Non-Contextual Autonomy Ranking and the Contextual Autonomy Ranking. Through these evaluations, the Autonomy of sUAS will be ranked; the difference between rankings will also be calculated. The results of these tests can be used for selecting a system by considering its autonomy score compared to other available systems.

## Non-Contextual Autonomy Ranking

Affiliated publications: [Hertel et al., 2022]

### Purpose

This test method is used to calculate a potential autonomy score without considering any mission-specific features. It allows the user to rank multiple sUAS according to their overall autonomy score which comprises several features including its operating time, camera feed characteristics, and number of sensors.

### Summary of Test Method

In this test, the user selects a set of mission-independent features (e.g., maximum range, number of smart behaviors) and measures those features for a set of sUAS in order to generate a series of non-contextual autonomy scores for comparison across sUAS. These data points can be gathered either by referencing vendor-provided specification sheets for each sUAS or measured and verified through experimentation using the test methods specified in this document (which is preferred for an ideal evaluation). The method used to derive the measure for each feature should be reported to delineate between assumed and demonstrated measures. The collected data is used to calculate several metrics including sUAS Autonomy Level, sUAS Component Potential, and ultimately the sUAS Combined NCAP Score.

The user can add or remove the mission-independent features they deem relevant. These are hardware-based features that their value can change from system to system, but does not change for a system when in different situations and missions. The following features are relevant examples for subterranean and constrained indoor operations, and can typically be derived from vendor-provided specification sheets and/or using the test methods specified in this document:

- Flight time: Time for a sUAS to deplete a fully charged battery. This is the limit for how long each sUAS is able to fly without landing and replacing the battery, measured in minutes.
- Charge time: Time required to fully charge a completely discharged battery. This should be a running average for each time the battery is completely discharged to completely charged, measured in minutes.
- Stream resolution: The resolution of the video feed that is streamed from the sUAS to the OCU, measured in pixels.
- Field of view (FOV): Angle away from the center of the camera on the sUAS which is visible in the images captured by the camera (i.e., the area of the inspection or observation captured on the sensor and measured in degrees along x and y axes), measured in degrees.
- Maximum range: The maximum distance at which the sUAS is able to maintain communications and operate, measured in meters. It may be difficult to evaluate this feature through experimentation due to limited access to kilometer distance ranges, so the vendor-provided specification is expected to be used.
- Thermal camera resolution: The resolution of the thermal camera onboard the sUAS, measured in pixels.
- Weight: The weight of the sUAS, measured in grams.
- Maximum flight speed: The maximum speed attainable by the sUAS in flight, measured in meters per second. It may be difficult to evaluate this feature through experimentation due to safety concerns with flying sUAS above 20 m/s, so the vendor-provided specification is expected to be used.
- Number of sensors: The number of different sensors onboard each sUAS, including video cameras, thermal cameras, inertial measurement units (IMUs), lidars, etc. Both proprioceptive and exteroceptive sensors are enumerated.

- Number of smart behaviors: The number of onboard preprogrammed autonomous behaviors of the sUAS, such as auto-launching, auto-landing, obstacle avoidance, waypoint navigation, etc.
- Signal-to-noise ratio (SNR): A measure of the signal-to-noise ratio present in each camera of the sUAS.

## Apparatus and Artifacts

If the data is being gathered from vendor-provided specification sheets, then no apparatus or artifacts are needed. However, if some or all of the features are to be measured and verified through experimentation, then proper artifacts according to the procedure for measuring each feature are required. For example, if signal-to-noise ratio (SNR) is to be included as a feature for the set of sUAS, then an enclosed room with visual acuity targets on the walls of the room is necessary, as well as a thermal acuity target. This room needs enough space to be able to operate the sUAS from an 8 meter distance to a wall.

## Equipment

If the data is being gathered from vendor-provided specification sheets, then no additional equipment is needed. However, if some or all of the features are to be measured and verified through experimentation, corresponding test equipment is required such as a timer to measure battery charge time and a scale to weigh the sUAS.

## Metrics

- sUAS Autonomy Level ($N_{AL}$): This level represents the non-contextual potential of sUAS capabilities in four areas (defined below). The total score increases by 1 if the system possesses abilities in each of the four areas. Although Perception is evaluated, a system that can only sense the world (e.g., possession of a working camera), but not perceive, model, plan, or execute autonomously (e.g., identifying the presence of a person or car in the camera image), will still have a $N_{AL}$ score of 0. The ability for a system to exhibit any of these capabilities shows an increased level of autonomy, and is used as one of the measurements of the NCAP score.
    - Perception: The ability for the sUAS to perceive its surroundings (e.g., automatic identification of a person in a camera image).
    - Modeling: The ability for the sUAS to model or map its surroundings (e.g., generate a point cloud representation of a doorway).
    - Planning: The ability for the sUAS to develop path planning in a map or other representation of the environment to be executed by either the system or the user.
    - Execution: The ability of the sUAS to execute a plan autonomously, without input from the user to perform the required actions (e.g., navigating to a waypoint).
- sUAS Component Potential ($N_{CP}$): This is calculated based upon the performance of the components in each sUAS, evaluating each of the selected mission-independent features (e.g., flight time, charge time, stream resolution, FOV, etc.). Methods to evaluate each feature through experimentation are referenced or specified in the procedure.
- sUAS Combined NCAP Score: This metric combines the values of $N_{AL}$ and $N_{CP}$, generating NCAP Component Scores, NCAP Coordinates, and the Potential Autonomy Distance to allow for comparison of sUAS. First create a table with rows for each sUAS being evaluated and columns for each feature being evaluated.
    - NCAP Component Score: The values of $N_{CP}$ for each feature being evaluated must be weighted. This could be either an even weight distribution or another system such as degree of autonomy, where each feature is assigned a degree of separation from pure autonomy, which will then be used to calculate a weight. This will be $2^{-n}$, where n is the degree of autonomy. The weightings that are chosen, however, must then be normalized so all the weights add up to a value of 1.
    - Using these weights, a weighted product *P* for each sUAS across each feature is calculated by using the equation in Equation 1, where the set of features is defined as $\phi_i$ for $i \in [0, N]$ where *N* is the number of features. $w_i$ is defined as the weight assigned to each feature. This equation states that the value of *P* for each sUAS will be the product of the value of each feature after being raised by the weight associated with that feature, for each sUAS. In this, a condition in

which a smaller score for a feature is ideal, when the score for that feature is raised by the weight, the weight will be negative.

- NCAP Coordinate: Each sUAS gets plotted on a chart, with the x-axis pertaining to the sUAS $N_{AL}$ value, and the y-axis pertaining to the $N_{CP}$ value. In both cases, higher is better.
- Potential Autonomy Distance: This is a measure of the distance from the origin to the NCAP Coordinate for each sUAS. A larger distance represents a higher level of autonomy, therefore ranking the sUAS. It is possible to compare each sUAS in a specific $N_{AL}$ level to each other, as well as comparing systems between $N_{AL}$ levels. A relative Autonomy Distance Difference is also calculated, which compares the distance between the NCAP Coordinates for each sUAS. This allows for the comparison of systems in a batch to determine not only which sUAS will perform the best, but the rankings of the next best systems as well.

$$P = \prod_{i}^{N} \phi_{i}^{w_i}$$

*Equation 1. Weighted multiple equation utilized to combine values for an overall score.*

## Procedure

1. Identify mission-independent features of sUAS that are to be evaluated.
2. Collect relevant sUAS feature measures, derived either through vendor-provided specification sheets or through experimentation. If deriving the features through experimentation, refer to the following test methods or follow procedures specified per feature:
    a. Flight time: Derive this measure from the Field Readiness: Runtime Endurance test method.
    b. Charge time: Follow this procedure:
        i. Acquire fully depleted battery from the sUAS.
        ii. Plug the battery into the charger and begin the timer.
        iii. When the battery is fully charged, stop the timer.
        iv. Repeat at least 10 times.
    c. Stream resolution: Derive this measure by using ASTM E2566-17a Standard Test Method for Evaluating Response Robot Sensing: Visual Acuity, or by following this procedure:
        i. Place the sUAS on the ground in a safe environment to launch.
        ii. Position a visual acuity target 2 meters away on a wall at a 1.2 meter height.
        iii. Mark positions 2, 4, 6, and 8 meters away from the wall.
        iv. If possible, take photos and a short video with the main camera while the sUAS is stationary, centered on the visual acuity target, while at each position. If not, launch and hover the sUAS such that the visual acuity target is in the center of the camera feed and take photos and a short video at each position.
        v. Land the sUAS and extract the photos and videos.
        vi. Analyze the photos and verify the resolution of the images and video.
    d. Field of view (FOV): Derive this measure by using ASTM E2566-17a Standard Test Method for Evaluating Response Robot Sensing: Visual Acuity, or by following this procedure:
        i. Place the sUAS on the ground in a safe launch environment to launch
        ii. Set up a visual acuity target 2 meters away on a wall at a 1.2 meter height.
        iii. Mark positions every 0.25 meters away from the visual acuity target, up to 2 meters on each side of the target on the wall, up to 1 meter above and below the target on the wall, and up to 1 meter away from the wall.
        iv. If possible, take photos and a short video with the main camera while the sUAS is stationary, centered on the visual acuity target, while at each position. If not, launch and hover the sUAS such that the visual acuity target is in the center of the camera feed and take photos and a short video at each position.
        v. Land the sUAS and extract the photos and videos.

vi. Analyze the photos and verify the FOV of the images and video. The verification can be done by following FOV calculation methods using existing tools [example].

e. Maximum range: Derive this measure by using ASTM E2855-12 Standard Test Method for Evaluating Emergency Response Robot Capabilities: Radio Communication: Non-Line-of-Sight Range.

f. Thermal camera resolution: Derive this measure by using ASTM E2566-17a Standard Test Method for Evaluating Response Robot Sensing: Visual Acuity, or by following this procedure:

   i. Place the sUAS on the ground in a safe environment to launch.
   ii. Position a thermal visual acuity target 2 meters away on a wall at a 1.2 meter height.
   iii. Mark positions 2, 4, 6, and 8 meters away from the wall.
   iv. If possible, take photos and a short video with the main camera while the sUAS is stationary, centered on the thermal visual acuity target, while at each position. If not, launch and hover the sUAS such that the thermal visual acuity target is in the center of the camera feed and take photos and a short video at each position.
   v. Land the sUAS and extract the photos and videos.
   vi. Analyze the photos and verify the resolution of the images and video.

g. Weight: Using a scale, weigh the sUAS.

h. Maximum flight speed: Maneuver the sUAS to reach its maximum flight speed and measure the speed in meters per second. This can be done by placing markers in the test environment and recording video of the sUAS flying near/above the markers at full speed. The speed can then be calculated by reviewing the footage and dividing the displacement by time spent to reach each marker.

i. Number of sensors: Count and list the number of sensors listed in the vendor-provided specification sheet, then follow this procedure:

   i. Place the sUAS in a safe launch environment, and begin the startup procedure.
   ii. Perform a search of the sUAS physical form for the expected sensors, and verify their existence.
   iii. Explore the OCU, searching for any indications of sensors through the interface (e.g., icon for GPS signal strength).
   iv. If necessary, verify the functionality of each sensor through flight.
   v. Take note of the actual number of sensors present in the system.

j. Number of smart behaviors: Count and list the number of smart behaviors listed in the vendor-provided specification sheet and operation manuals, then follow this procedure:

   i. Place the sUAS in a safe launch environment, and begin the startup procedure.
   ii. Explore the OCU, searching for any indications of smart behaviors onboard through the interface (e.g., menu options to initiate waypoint following), and take note of any unexpected smart behaviors found.
   iii. If necessary, verify the functionality of each smart behavior through flight.
   iv. Take note of the actual number of smart Behaviors present in the system.

k. Signal-to-Noise Ratio (SNR): Follow this procedure:

   i. Place the sUAS in a safe launch environment that includes at least one wall or surface that has some visual texture (i.e., not a single flat color), and begin the startup procedure.
   ii. Set up a visual acuity target 2 meters away on the wall at a 1.2 meter height.
   iii. If possible, take 11 photos with the main camera while the sUAS is stationary, centered on the visual acuity target. If not, launch and hover the sUAS such that the visual acuity target is in the center of the camera feed and take 11 photos.
   iv. Land the sUAS and extract the photos.
   v. Utilize the provided SNR calculation software [link] on a folder of the extracted photos which will calculate the SNR value.

3. Using these measures, calculate the sUAS Autonomy Level ($N_{AL}$), sUAS Component Potential ($N_{CP}$), and the sUAS Combined NCAP Score metrics.

## Example Data

**Data acquisition**:

| sUAS | Flight Time (min) | Charge Time (min) | Stream Res. | FOV | Max. Range (m) | Thermal Camera Res. | Weight (g) | Max. Flight Speed (m/s) | # of Sensors | # of Smart Behaviors |
|---|---|---|---|---|---|---|---|---|---|---|
| A | 15 | 50 | FHD | 100° | 2000 | N/A | 370 | 3 | 3 | 2 |
| B | 10 | 90 | FHD30p | 114° | 500 | 160x120 | 1450 | 6.5 | 10 | 7 |

For calculating the component potential ($N_{CP}$) with the weighted product formula (see Equation 1), the weights can be uniform (e.g., all 0.1) or user defined (e.g., 0.07, 0.03, 0.1, 0.1, 0.05, 0.1, 0.05, 0.05, 0.15, 0.30).

If the value for a feature is N/A, it will be replaced with the minimum possible value for that feature. For the camera resolutions, the lowest value in our data set is given a value of "1", and the next lowest value is given a "2", and if there are more resolutions, their value for the purpose of the weighted product follows this same path. For our set, FHD is given a value of "3", FHD30p is given a value of "2", and the thermal camera resolution 160x120 is given a "1".

The Autonomy Distance can measured as an absolute measure, which measures the distance from the origin (0,0) to the Autonomy Coordinate ($N_{AL}$, $N_{CP}$), or as a relative measure, which can be used to measure the distance of each system relative to the best performing system, defined by the system with the largest absolute Autonomy Distance.

**Uniform weights**:

| sUAS | Component Potential ($N_{CP}$) | Autonomy Level ($N_{AL}$) | Relative Autonomy Distance |
|---|---|---|---|
| A | 2.48 | 3 | 0 |
| B | 2.69 | 1 | 2.01 |

**User defined weights**:

| sUAS | Component Potential ($N_{CP}$) | Autonomy Level ($N_{AL}$) | Relative Autonomy Distance |
|---|---|---|---|
| A | 3.17 | 3 | 2.49 |
| B | 4.66 | 1 | 0 |

When using uniform weights, sUAS B has the largest relative, and is therefore considered the best performing among the two. With the user defined weights, however, the order changes and sUAS A now has a larger relative Autonomy Distance. This example shows how our method can adapt to different feature weight values and thus weight assignment can have a significant effect on the autonomy evaluation of the systems. In case the user has no preference over selected features, we recommend using a uniform weight vector.

## Contextual Autonomy Ranking

Affiliated publications: [Donald et al., 2023]

### Purpose

This test method focuses on the autonomy evaluation of sUAS in a mission-specific context. It allows the user to rank multiple sUAS according to their mission-specific performance by calculating an autonomy score based on various mission, environmental, and operator-related factors. Similar to other autonomy evaluation approaches, we assume a mission can be decomposed to a set of sub-tasks. If required, we can have different levels of sub-tasks in a hierarchical form. For example, a sub-task in level 1 can be decomposed to a set of sub-tasks in level 2.

### Summary of Test Method

In this test, the user selects a sub-task to be evaluated in a specific environment. Several examples have been described in this document including Runtime Endurance in Enclosed Spaces, Takeoff and Land/Perch, Navigation Through Apertures, Navigation Through Corridors, and the Room Clearing test. It should be noted that the proposed framework is not limited to these sets of tests and can be used for any sUAS mission. For each experiment, data should be collected according to three axes of Environmental Complexity (EC), Mission Complexity (MC), and Human Independence (HI). This is similar to the Autonomy Levels for Unmanned Systems (ALFUS) framework [Huang et al., 2007]. These three axes allow the user to categorize various factors of a mission that can affect a system's autonomy efficiently. Here we give a brief explanation for each category.

- Environmental Complexity (EC) axis accounts for the differences in terrains and environments of a mission. In this axis, larger values are a result of increased complexity in the environment of the mission. An example of this is a mission that takes place in an open, brightly lit space as a low level on the EC axis, while the same mission in a space with many obstacles, possible wind drifts, and a lower lighting condition would be considered to represent a larger value in the EC axis.
- Mission Complexity (MC) axis accounts for the different levels of difficulty in movements, actions, or decisions required to complete a mission successfully. In this axis, larger values are a result of increased complexity in the mission. Missions that require more detailed and precise movements are considered more complex. An example of a low MC level is a translational movement between three waypoints, while an increased MC level could include more points and/or different types of movements such as landing or hovering in place.
- Human Independence (HI) axis accounts for the level of independence the sUAS offers to the operator for the successful completion of a mission. This axis also accounts for the types of actions which are able to be completed by the sUAS, and the complexities of those actions. Higher values on this axis indicate sUAS that can complete more complex movements while requiring more independence from the operator. The more difficult the tasks which the sUAS is able to perform results in a larger value along the HI axis. In addition, the larger the portion of the mission which is able to be completed by the sUAS without the operator, will also result in a larger value along the HI axis. This allows for a more complete evaluation along the HI axis in this autonomy space.

The user can add or remove the mission-specific features to each axis as they deem relevant. The following features are relevant examples for subterranean and constrained indoor operations, and can typically be derived from the test methods specified in this document. Note that metrics used in the evaluation of sub-tasks, as described in other sections of this document, can be used as features in this test for comparing different sUAS autonomy levels.

- Aperture/Hallway Cross-Sectional Area: This is the cross-sectional area of the aperture/hallway that is being traversed, in meters squared.
- Ambient Light Level: This is the ambient light level during the test, measured in Lux.
- Verticality of the Hallway: A measure which represents the angle from the horizontal of the hallway, in degrees. This will be between 0 and 90.

- Number of Crashes: Total number of crashes for the sUAS during the specific test.
- Number of Rollovers: Total number of rollovers for the sUAS during the specific test.
- Completion Percentage: Percentage of runs which result in a completion of the test, for a specific test.
- Static Roll Angle: The roll angle, in degrees, of the platform the sUAS is taking off of, or landing on.
- Static Pitch Angle: The pitch angle, in degrees, of the platform the sUAS is taking off of, or landing on.
- Static Vertical Obstruction: The distance from the center of the platform the sUAS is taking off of, or landing on, to a vertical obstruction, in meters.
- Static Horizontal Obstruction: The distance from the center of the platform the sUAS is taking off of, or landing on, to a horizontal obstruction, in meters.
- Coverage Percentage: The percentage of Landolt C's which are identifiable in the Room Clearing test.
- C's Detected: The number of Landolt C's which are identified in the Room Clearing test.
- Duration: The duration of the test, in minutes.
- Obstructions: The number of obstructions in the path of the sUAS.

## Apparatus and Artifacts

The apparatus and artifacts required to complete this evaluation depends on the ones required for data collection in the sub-task in hand (e.g., split-cylinder fiducials for the Mapping test methods). For the tests described in this project, the contextual autonomy evaluation process relies on the designated apparatus and artifacts described throughout this document for each test. For any test not included in this document, the requirements should be determined by the user.

## Equipment

If the data is being used from previously conducted experiments for the target sub-task, then no additional equipment is needed. If the test data is not available, the user is required to use relevant equipment and metrics (e.g., a timer to record test duration in Field Readiness: Runtime Endurance test method). To evaluate the contextual autonomy of the systems using this method, in this document, we used the collected test data from other sections.

## Metrics

We define four contextual autonomy metrics:

- sUAS MC Autonomy Score: This score represents the autonomy score of a sUAS along the MC axis by combining various features that affect mission complexity. This score is in range [0,1]. A score of 0 relates to systems that cannot efficiently handle missions with low levels of complexity. Whereas, a score of 1 represents a system that can handle complex missions.
- sUAS EC Autonomy Score: This score represents the autonomy score of a sUAS along the EC axis by combining various features that affect environmental complexity. This score is in range [0,1]. A score of 0 relates to systems that cannot efficiently perform in environments with low levels of complexity. Whereas, a score of 1 represents a system that can perform well in complex environments.
- sUAS HI Autonomy Score: This score represents the autonomy score of a sUAS along the HI axis by combining various features that affect human independence. This score is in range [0,1]. A score of 0 relates to systems that cannot operate independently. Whereas, a score of 1 represents a system that can operate with minimum assistance from a user.
- sUAS Combined Contextual Autonomy Score: This score is the combination of other three scores and represents the combined contextual autonomy score of a sUAS. This score is in range [0,1]. Different values in this range represent systems according to their performance depending on mission complexity, environmental complexity, and system's human independence level and a unified score.

## Procedure

1. Specify a sub-task by decomposing a mission into simpler sub-tasks if required.
2. For a given test, select features and collect data (e.g., the data and metrics collected in other sections of this document).
3. For each feature, create a set of levels that can be used to represent different intensities or levels of the feature. For example, for the ambient light feature, we select three levels and call them *low*, *medium*, and *high*.
4. Define a range for each feature that includes all possible values that feature can take. This is a design step and the range can be different for different features. However, using the range [0, 1] is a good choice for most features given that their actual range can be mapped to [0, 1] without loss of generality. For example, we can select three membership functions and name them low, medium, high. These should cover the general testing range, however outliers may need to be adjusted to fit into this range. The lower bound of the range is the point at which it is certain that any value worse than this is just as bad in terms of the performance. (e.g. in our test data, 3 or more crashes.) The upper bound of the range is the point at which any value better than this, is just as good in terms of the performance (e.g. Ambient light of 750 Lux or greater).
5. Define one membership function for each level in your range. A membership function represents the degree of truth for a defined level in the whole range. For example, we select three triangular membership functions each of which can be defined using a three point tuple (a, b, c) that represent the first, middle, and last point of the triangle, respectively. Each membership function should be designed by selecting the three values a, b, and c. For instance, see the below image.

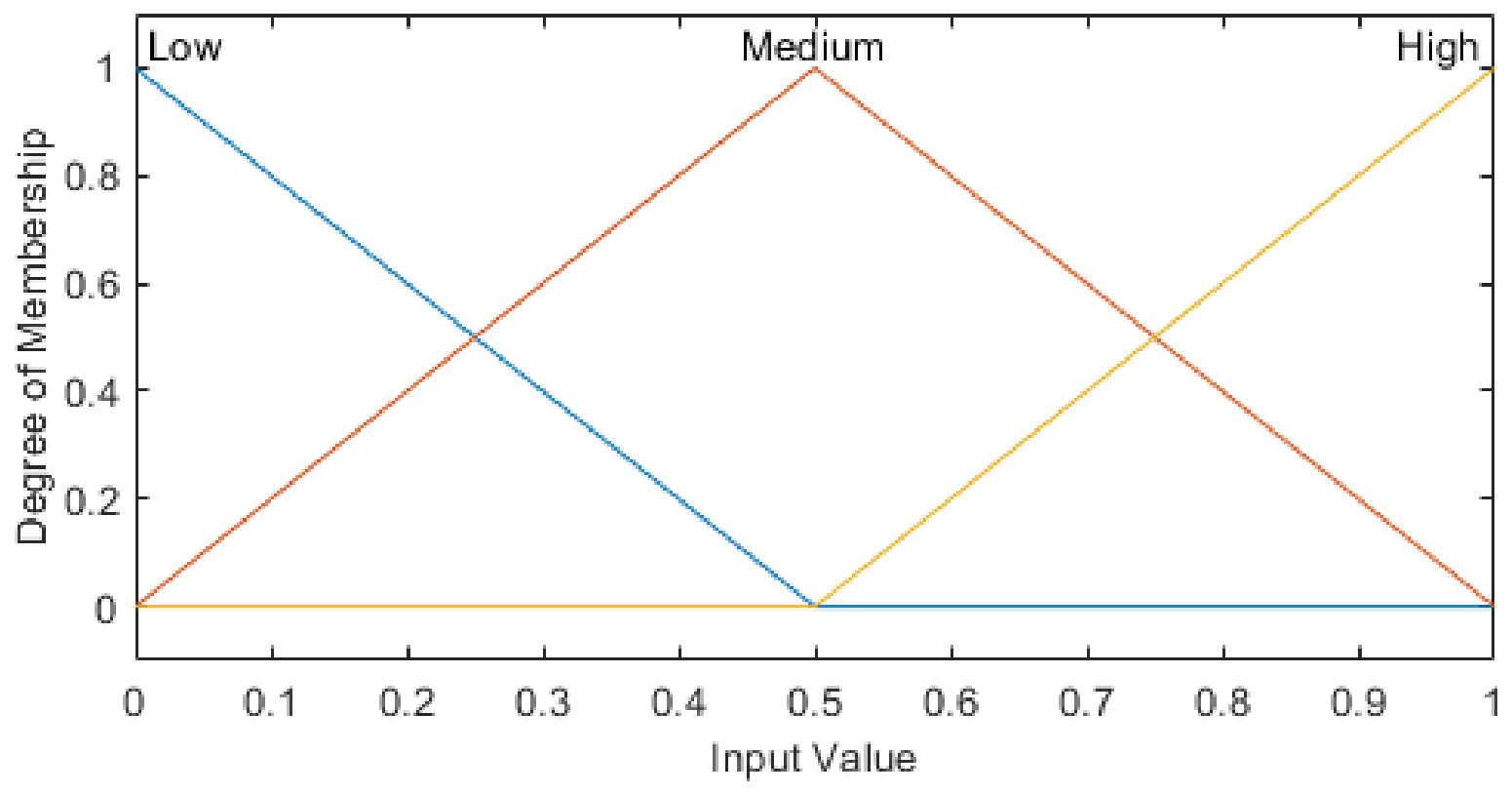

   Note that the user must define the number of the levels and the type of the membership functions. For all the experiments, we have selected three triangular membership functions. The user can tune the placement of the membership functions over the range of the feature, however, a uniform placement of membership functions over the range would suffice in most cases.

6. The user must then define a set of constant membership functions for the output of the system (i.e., metrics). Those membership functions are constants selected in range [0, 1]. Our systems have five output values equally spaced in the range [0,1]. Although, the range can be anything and is not limited to [0,1]. An example can be seen in the following figure.

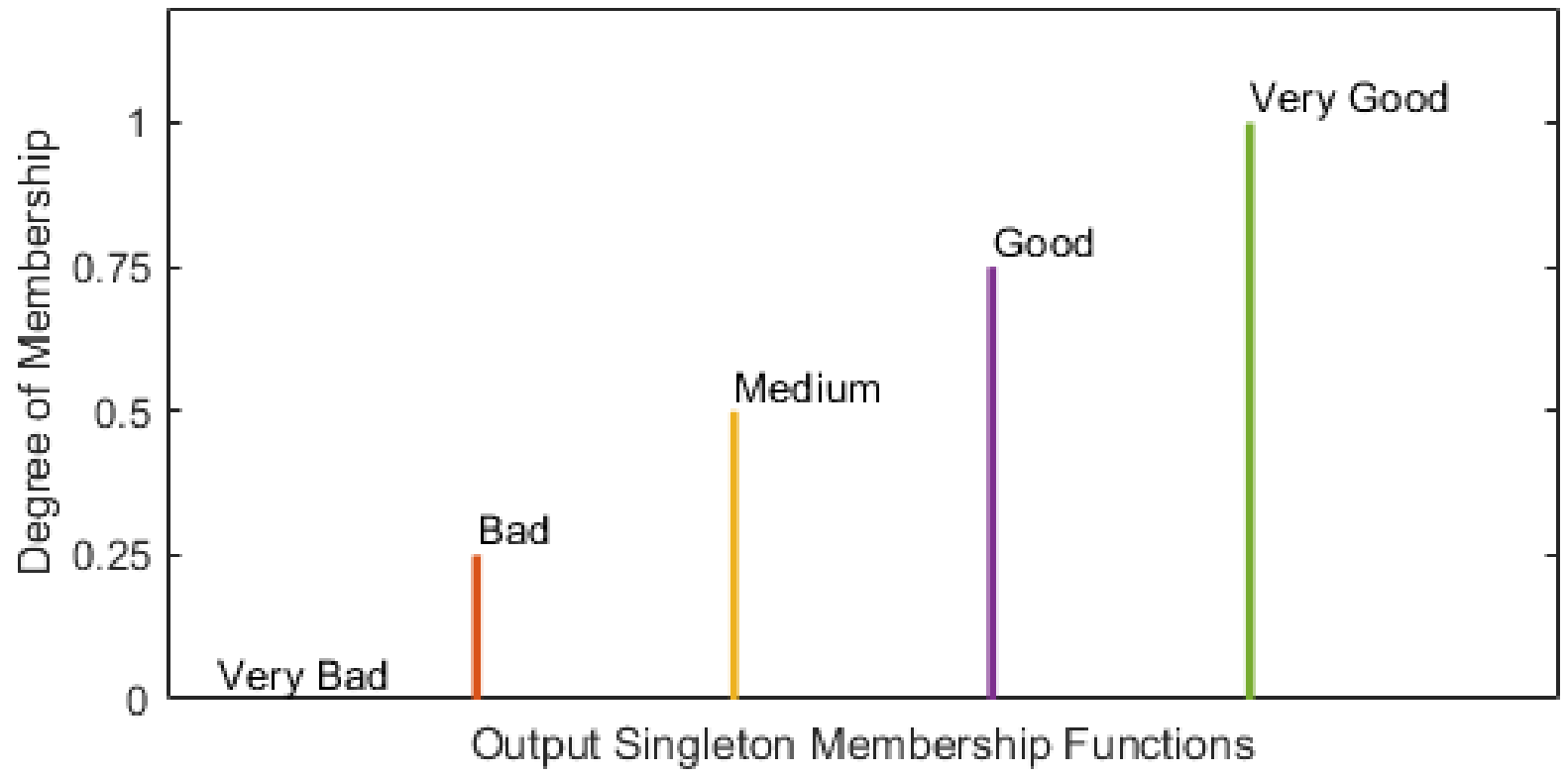

7. When the membership functions for all the features and the outputs are defined, we should generate a linguistic rule-base. An example of a linguistic rule is: “IF number of crashes is low AND ambient light level is low THEN output is very good”. This rule combines two features: number of crashes and ambient light level over the low and low membership function defined for the features, respectively, relating them to the membership function defined for the output of the system. The rule-base which determines how the system evaluates and combines inputs, should be intuitive and include multiple rules per input. However, it does not require to be complete (i.e., there is one rule for all the combinations of inputs, membership functions, and outputs). The following table shows an example of a rule-base for our combined contextual autonomy fuzzy system.

| | | Mission Complexity Axis | | |
|---|---|---|---|---|
| | | Low | Medium | High |
| Environment Complexity Axis | Low | Very Bad | Bad | Medium |
| | Medium | Bad | Medium | Good |
| | High | Medium | Good | Very Good |

8. The input and output membership functions, defined ranges, and the designed rule-base can be used to generate a Fuzzy Inference System (FIS), for instance using the provided software or in MATLAB. A figure showcasing the overall structure of our cFIS can be found below:

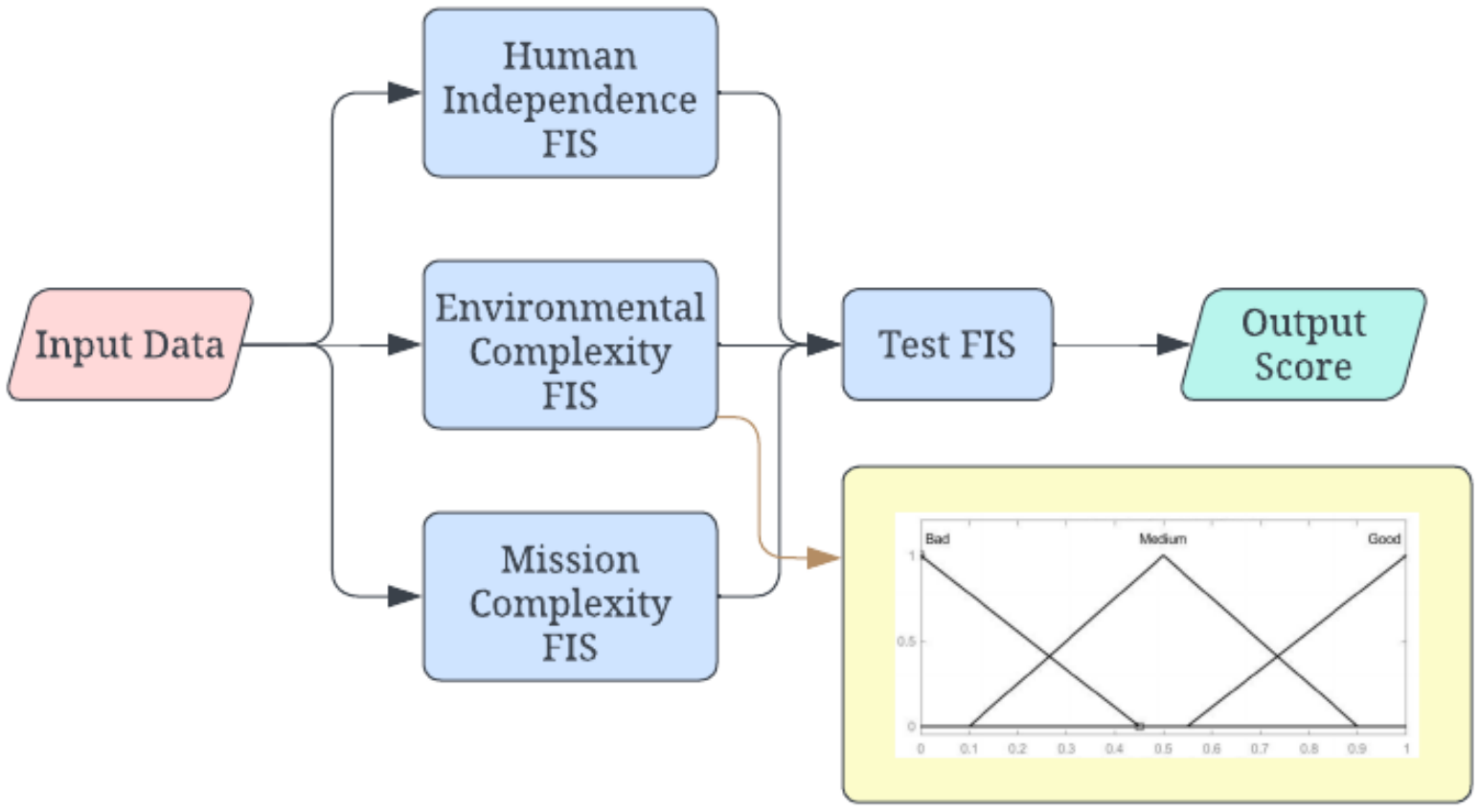

This structure calculates the three scores and then combines them to get the combined contextual autonomy score.

9. Given that this score was calculated for the specified sub-task, the user can go one level further to calculate an overall score for the system over different sub-tasks in a mission by combining the sub-task autonomy scores. We call this overall or predictive mission score. Calculating this overall score requires a weight value for each sub-task indicating its importance during the mission. Weights must add to 1.
10. The predictive score for each sUAS can then be calculated utilizing the weighted multiple equation in the non-contextual evaluation (Equation 1).

## Example Data

**Data acquisition:**

The data used in this portion of the evaluation is simply data from other tests, and thus to collect this data, it requires the other tests to be completed first. Sometimes, the data used is not necessarily tabulated along with other metrics for each test, but inherent to the test design itself. One example of this is the different environmental factors, such as the cross-sectional area of an aperture/corridor, the verticality of a corridor, or the ambient light level. These need to be measured and accounted for in each test used, however some of these values are intrinsic to the test being performed, and are defined. An example of this is the Takeoff and Land/Perch test, where the different conditions are inherent to the test, such as the lateral obstructions.

To calculate the results from each test, it is easiest to input the FIS into a program, either MATLAB or Python, alongside the data points for each sUAS. Once this is done, the specified input data can be input into the correct FIS representing the axis the data is associated with. Then, the outputs of the three axes can be input into the combinational FIS, and output scores will then be calculated. In the example data, there are some gaps, due to different sUAS being unavailable for certain tests for various reasons. As a result, we utilize the percentage of the possible score achieved, as this allows us to compare results between systems which have more or less testing performed. Ideally, there would be no gaps in data, and as a result the raw scores can be utilized, which will allow for a better representation of each system.

**Designed FIS:**

In this section, a single cFIS will be explained, including each of the three FIS included, and associated inputs and outputs. This cFIS represents the Takeoff and Land/Perch test evaluation, and can be seen in the figure below:

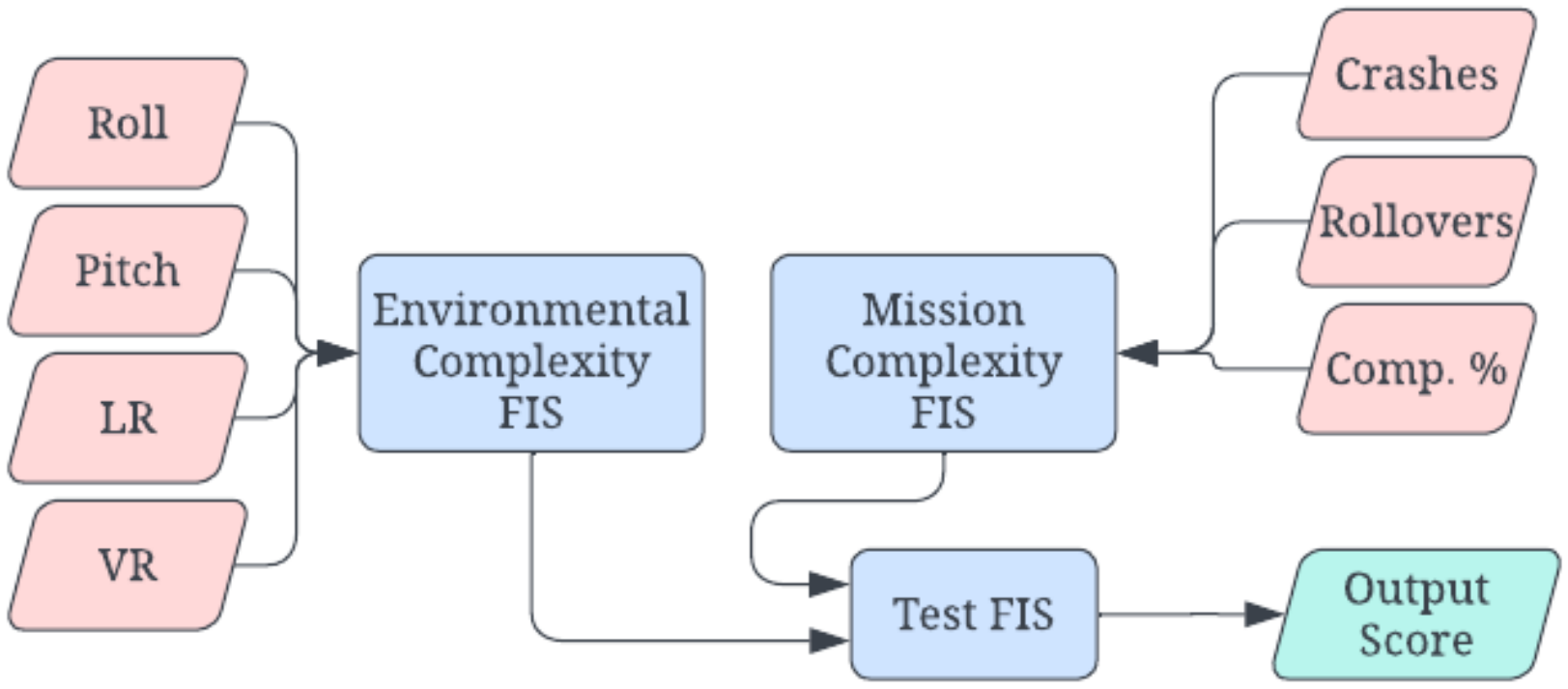

In this figure, there are two initial FIS, which represent the Environmental Complexity and Mission Complexity axes. The Human Independence axis is not taken into account for this test, as each sUAS utilized in this test was flown with complete teleoperation. For the MC FIS, there are three inputs, Crashes, Rollovers, and Completion Percentage. For each of these, the following membership functions are used: Crashes: {low: [0, 0, 1.25], medium: [0.5, 1.5, 2.5], high: [0.5, 1.5, 2.5]}, Rollovers: {low: [0, 0, 1.25], medium: [0.5, 1.5, 2.5], high: [0.5, 1.5, 2.5]}, Completion Percentage: {low: [0, 0, 0.55], medium: [0.15, 0.6, 0.92], high: [0.7, 1, ]}. In terms of the Crashes and Rollovers, the two identical sets of membership functions form the way they do because each value is inherently an integer, and the number of crashes which is always considered low is 0 in this case, the number that is always considered high is 3 in this case. The data which is not as clearly defined is in the middle, where 1 and 2 are both more likely to be considered medium, however 1 can sometimes be considered low, and 2 can sometimes be considered high. The completion percentage is skewed towards the higher end, because what is considered in the middle of the performance is closer to the higher end, and a high completion percentage is typically viewed as only the higher end of the range. For this FIS, the following ruleset is utilized to calculate the output. For each row, the convection follows “IF Crashes and Completion Percentage and Rollover then Output”.

| Crashes | Completion Percentage | Rollover | Output |
|---|---|---|---|
| Low | High | Low | Very Good |
| Not Low | High | Not Low | Good |
| Many | Not High | High | Very Bad |
| Not Low | Medium | Not Low | Bad |
| Low | Low | Low | Bad |
| Not Low | Low | Not Low | Very Bad |
| Low | Medium | Low | Good |
| Low | High | Not Low | Good |
| Not Low | High | Low | Good |
| Not Many | High | High | Medium |

Next, the same will be explained for the EC FIS. As shown in the figure, this FIS contains four inputs, however due to the nature of the test, this FIS actually is more simplistic than the previous one. These inputs are as follows: platform pitch angle, platform roll angle, platform lateral restriction, and platform vertical restriction. Next, each variable has the following membership functions associated with them: Roll: {low: [0, 0, 4.17}, medium:[0.83, 5, 9.12], high:[5.83, 10, 10]}, Pitch: {low: [0, 0, 4.17}, medium:[0.83, 5, 9.12], high:[5.83, 10, 10]}, Lateral Obstruction: {low: [1.2, 1.2, 2.4}, medium:[1.4, 2.4, 3.4], high:[2.6, 3.6, 3.6]}, Vertical Obstruction: {low: [0.6, 0.6, 1.1}, medium:[0.7, 1.8, 1.7], high:[1.3, 1.8, 1.8]}. This test is unique, in which a version is performed in which each variable is varied independently. Due to this, there is a classification of "low", "medium", or "high" difficulty for each variable inherent to the test. This FIS allows us to accommodate the classification of an environment when more than one variable is varied. The ruleset we use is as follows:

| **Roll Angle** | **Pitch Angle** | **Lateral Obstruction** | **Vertical Obstruction** | **Output** |
|---|---|---|---|---|
| Low | Low | Low | Low | Very Bad |
| Medium | Medium | Medium | Medium | Medium |
| High | High | High | High | Very Good |

Finally, the outputs from each FIS are input to the combinational FIS, which is described in the procedure, along with the example ruleset, which is the ruleset for the combinational FIS.

Found below is an evaluation of several sUAS, in each test. These values are a percentage of the possible scores for each sUAS, as mentioned above.

| **sUAS** | **Navigation: Through Corridors** | **Navigation: Through Apertures** | **Takeoff** | **Landing** | **Runtime Endurance: Indoor Movement** | **Room Clearing** | **Predictive Score** |
|---|---|---|---|---|---|---|---|
| sUAS A | 0.90 | 1.0 | 0.71 | 0.87 | 0.76 | 0.73 | 0.82 |
| sUAS B | 1.0 | 1.0 | 1.0 | 1.0 | 0.5 | 0.76 | 0.85 |
| sUAS C | 0.84 | 1.0 | 1.0 | 0.87 | - | - | 0.92 |
| sUAS D | 0.83 | 0.83 | 1.0 | 1.0 | 0.5 | 0.79 | 0.80 |
| sUAS E | - | - | 0.75 | 0.97 | 0.65 | 0.75 | 0.77 |
| sUAS F | - | - | 0.99 | 0.91 | - | - | 0.95 |
| sUAS G | 0.80 | 1.0 | 0.82 | 0.89 | - | 0.85 | 0.87 |

For the calculation of the predictive score, we used equal weight values for each test. If a test has not been completed, it is given a weight of 0 and is not included in the calculation.

# Trust

There are numerous factors which will determine whether or not a human user will trust and utilize sUAS for performing a mission. Performing missions in subterranean and constrained indoor environments is risky and dangerous, which may add to these factors or vary the importance of some of the factors of human trust in the system. These include system-related factors (e.g., performance, appearance, behaviors), human-related factors (i.e., operator-specific preferences, characteristics), and task-related factors (i.e., task type, environment). The trust test methods measure the impact of factors that may have more considerable effects on the formation or loss of trust between a human and sUAS in a subterranean and constrained indoor environment, and can be used for assessing trustability of different sUAS.

## Characterizing Factors of Trust

### Purpose

The purpose of this test method is to assess and compare the effects of several factors related to sUAS performance, appearance, noise profile, task/interaction type, and the environment on human trust in the sUAS, as well as determine the impact of each factor on human-sUAS trust.

### Summary of Test Method

This test method consists of recording videos of scripted sUAS operations and conducting surveys of participants to provide subjective feedback regarding their trust in the system. A set of relevant sUAS features that could impact trust are selected (e.g., performance, appearance, interaction type) along at least two conditions for each feature (e.g., appearance: with or without exposed propellers). A relevant task and environment description is also generated to match the use case. The videos are then scripted such that the task can be performed equivalently when sUAS are used that possess each set of relevant feature conditions. A series of videos are recorded, one for each set of feature conditions, showing both an external view of the operator commanding the sUAS and a camera feed from the onboard the sUAS. The videos should follow a similar structure in terms of timing, resolution, and camera angles such that they are similar enough to the others in the series aside from the feature condition that is active. The videos can be edited as needed in order to adjust their feature conditions (e.g., to make a sUAS appear to generate less noise than it actually does, replace the audio with that of a quieter sUAS or by adjusting the volume). Examples of videos are linked later in this section; screenshots of example videos can be seen in Figure 1.

Participants read a description of a scenario involving sUAS performing an operationally relevant task (e.g., performing a mapping mission in the ruins of a subway line) in which there might be some obstacles, hazards, and other features of interest. In the scenario, the feature conditions that will be evaluated are described as options for how to perform the task with the sUAS. For example, if the impact of interaction type on trust is to be evaluated, then the scenario will describe that the sUAS can either be teleoperated remotely or while co-located with the sUAS. Participants then watch a video of the scenario showing the sUAS performing the task with one of the feature conditions active. The selection of which feature condition to show a participant is selected randomly from the set of generated videos, but the number of responses sought for each feature condition should be equivalent to balance the results.

It should be noted that a human subjects research protocol will likely be required in order to recruit and run participants. All materials to run this test method including the videos and questionnaires should be reviewed by an institutional review board (IRB) and/or human research protections office (HRPO) as needed.

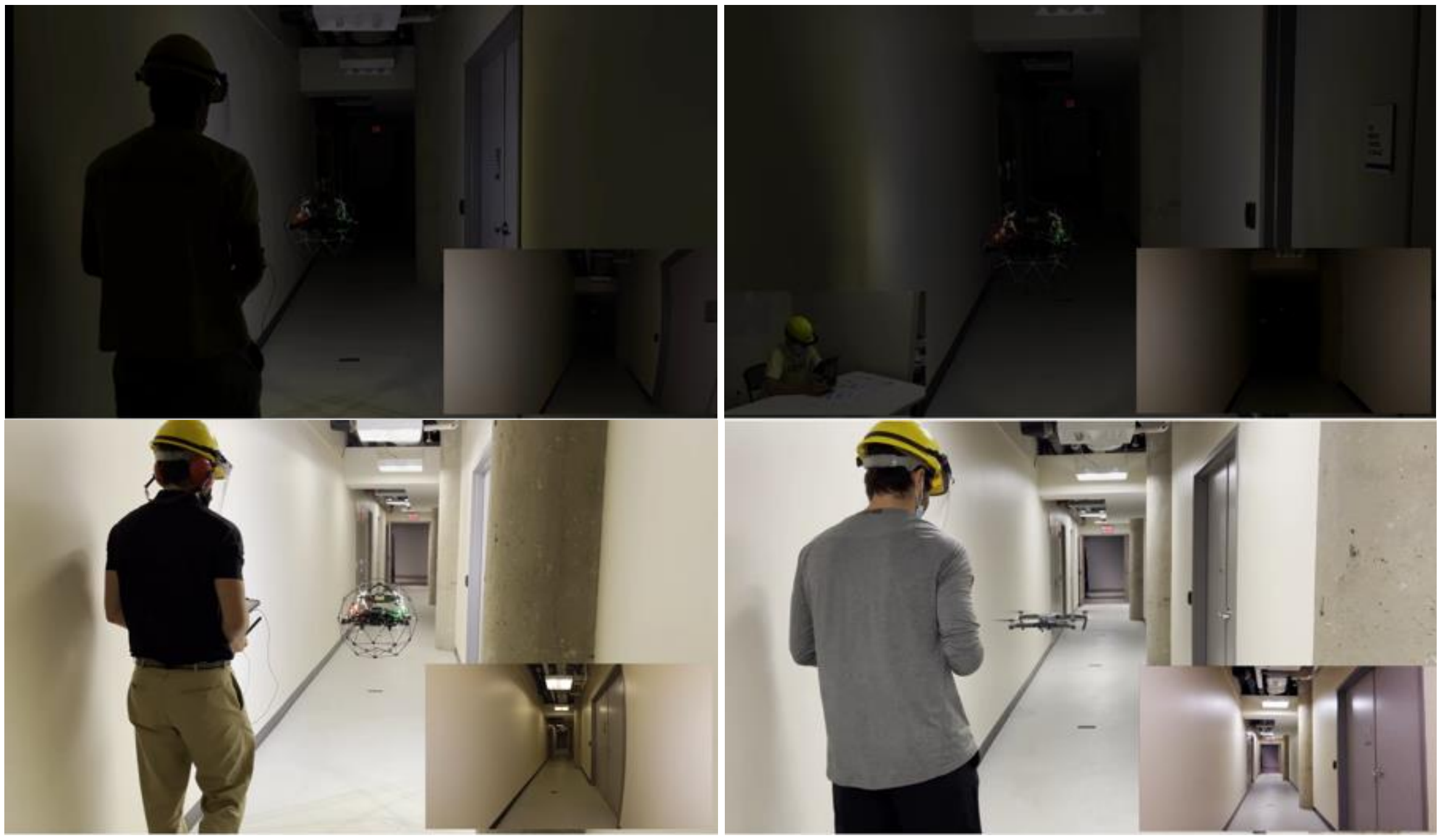

*Figure 1. Screenshot of example videos shown to participants with an external view showing the operator and a picture-in-picture view of the sUAS camera feed. Top: Example Feature Characterization videos showing dark environments flying sUAS A when co-located (left) and remote (right). Bottom: Example System Comparison videos showing bright environment co-located operation flying sUAS A 2 (left) and sUAS B (right).*

Two types of tests are specified that investigate the effects of different factors on human trust in a sUAS as an assistant in a cooperative task: Feature Characterization and System Comparison. Figure 2 shows an outline of each test. Participants can be recruited to perform this test in-person or online (e.g., Amazon Mechanical Turk, Prolific), noting that online questionnaires can yield higher response rates. It is also recommended that manipulation check questions are added that can be used to validate whether a participant's response is valid. An example of this is asking the participant a question about what they saw/heard in the video they were shown such that, if answered correctly, implies they were paying close enough attention.

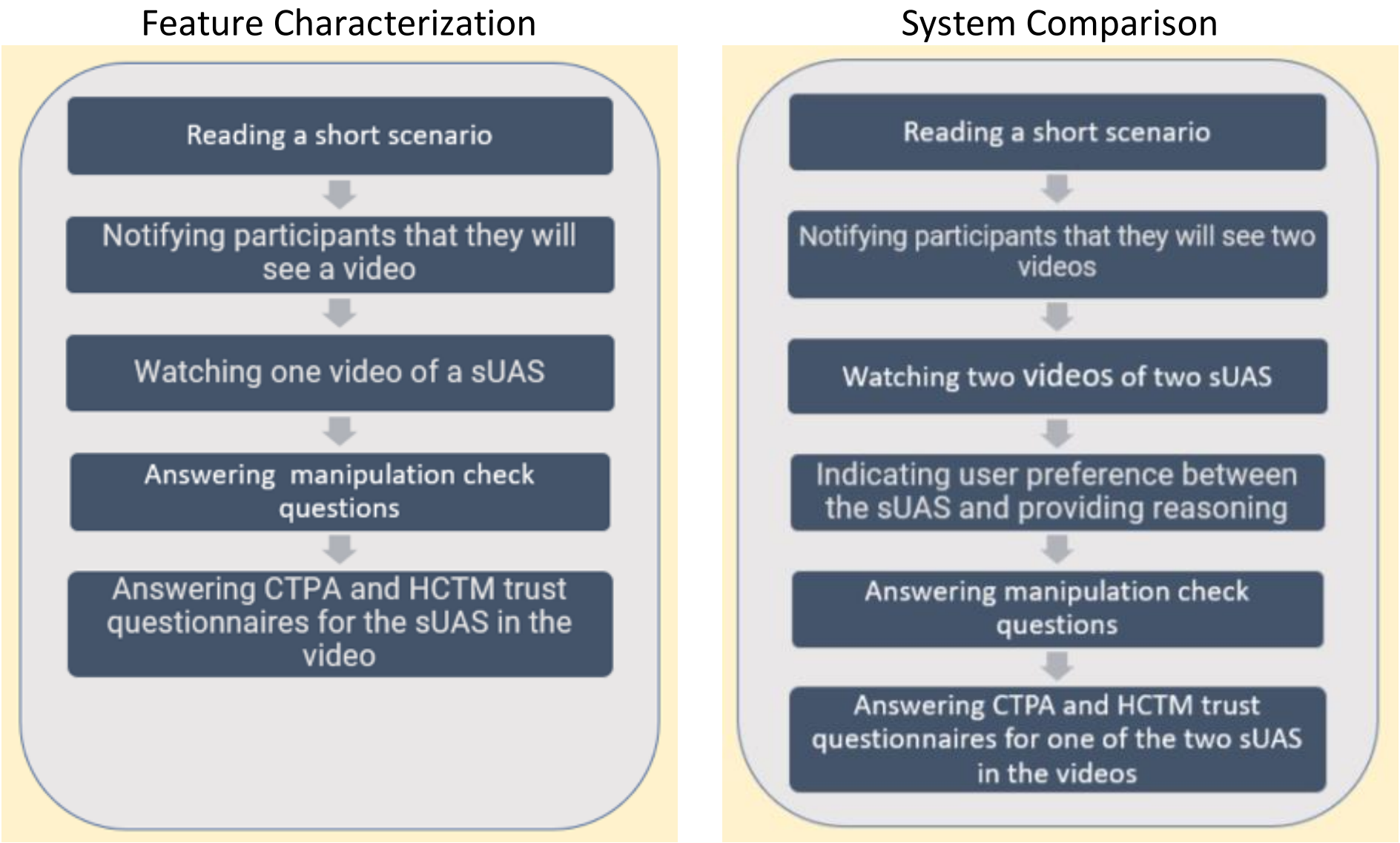

*Figure 2. The procedures of the two types of tests used in this test method.*

Feature Characterization: Each participant watches a single video/feature condition. After watching the video, participants respond to questionnaires to rate their trust in the sUAS shown in the video. Below are example videos of feature conditions for a navigation task through an indoor office (note: these videos also feature text or audio stating the length of the hallway and the number of rooms in the environment, of which the user is queried about as manipulation checks):

| **Lighting** | **Interaction type** | **Noise** | **Appearance** | **Illuminators** | **Video** |
|---|---|---|---|---|---|
| Dark | Co-located | High | Protective cage | On | https://youtu.be/i0jdArx_bbw |
| Dark | Remote | High | Protective cage | On | https://youtu.be/eq028gvP7bM |
| Lighted | Co-located | High | Protective cage | N/A | https://youtu.be/j6TPDhefXcg |
| Dark | Co-located | Low | Protective cage | On | https://youtu.be/L7ZhB3ccda8 |
| Lighted | Co-located | Low | Exposed propellers | N/A | https://youtu.be/sVieSEru9e0 |
| Dark | Co-located | Low | Exposed propellers | Off | https://youtu.be/JQ-hDmHGik0 |

System Comparison: Each participant watches two videos (randomly ordered, but balanced across participants) with one or more feature conditions active, each video using a different sUAS platform to perform the task. After watching the videos, participants first answer a questionnaire to indicate their preference of which system they would prefer to have perform the task, provide a reason why, and then respond to questionnaires to rate their trust in one of the two sUAS (randomly selected, but balanced across participants) shown in the video. Below are example videos of feature conditions for a navigation task through an indoor office (note: these videos also feature text or audio stating the length of the hallway and the number of rooms in the environment, of which the user is queried about as manipulation checks):

<table>
<tr><th>Lighting</th><th>Interaction type</th><th>System</th><th>Noise</th><th>Appearance</th><th>Illuminators</th><th>Video</th></tr>
<tr><td rowspan="2">Light</td><td rowspan="2">Co-located</td><td>sUAS A</td><td>High</td><td>Protective cage</td><td>N/A</td><td>https://youtu.be/9F0LaH4lOHo</td></tr>
<tr><td>sUAS B</td><td>Low</td><td>Exposed propellers</td><td>N/A</td><td>https://youtu.be/b6H12nRdRNI</td></tr>
<tr><td rowspan="2">Dark</td><td rowspan="2">Co-located</td><td>sUAS A</td><td>High</td><td>Protective cage</td><td>On</td><td>https://youtu.be/7m9p3LO9or8</td></tr>
<tr><td>sUAS B</td><td>Low</td><td>Exposed propellers</td><td>Off</td><td>https://youtu.be/eASF3gq1FR0</td></tr>
</table>

## Apparatus and Artifacts

Environmental and task elements may need to be fabricated in order to generate videos with the selected feature conditions, but otherwise no specific apparatus or artifacts are required.

## Equipment

One or more video cameras are required in order to record sUAS and operator activities to generate the test videos shown to participants. The participants must have access to a computer for watching the videos and responding to the questionnaires.

## Metrics

- Manipulation check: The participant answers one or more questions about what they saw/heard in the video they were shown. If answered correctly, their responses to the questionnaires can be considered valid, but if not answered incorrectly, their responses to the questionnaires may not be valid.
- Checklist for Trust between People and Automation (CTPA): [Jian et al., 2000] The user indicates their level of agreement with a series of statements using a 7-point Likert scale (1 = not at all, 7 = extremely). The CTPA used for this test method consists of the following statements:
  - I am wary of the system.
  - The system's actions will have a harmful or injurious outcome.
  - I am confident in the system.
  - The system provides security.
  - The system has integrity.
  - The system is dependable.
  - The system is reliable.
  - I can trust the system.
  - I am familiar with the system.
- Human-Computer Trust Measure (HCTM): [Pinto et al., 2020] The user indicates their level of agreement with a series of statements using a 7-point Likert scale (1 = not at all, 7 = extremely). The HCTM used for this test method consists of the following statements:
  - I believe that there could be negative consequences when using the system.
  - I feel I must be cautious when using the system.
  - It is risky to interact with the system.
  - I believe that the system will act in my best interest.
  - I believe that the system will do its best to help me if I need help.
  - I believe that the system is interested in understanding my needs and preferences.
  - I think that the system is competent and effective in performing the mapping mission.
  - I think that the system performs its role as an assistant for the mapping mission very well.
  - I believe that the system has all the functionalities I would expect from an assistant for performing the mapping mission.
  - If I use the system, I think I would be able to depend on it completely.
  - I can always rely on the system for performing the mapping mission.
  - I can trust the information presented to me by the system.
- User preference: The participant chooses which of the systems they prefer and provide reasoning for their choice (e.g., possession of a cage, less disruptive noise, etc.).

## Procedure

After recording and editing the videos needed to run the test method, the survey method that will provide the participant with instructions, show the videos, and provide them with questionnaires to fill out must be designed. This could be done manually using pen and paper with a computer screen to display the videos, or through an online survey tool such as Qualtrics.

**Data collection**

1. Participants are first asked to review and sign a form providing their consent to participate (following IRB and HRPO requirements).
2. Participants are asked to read a scenario about performing a task with sUAS, the description of which shall include the feature conditions selected for the experiment.
3. If the Feature Characterization test is being conducted:
   a. Participants are instructed that they will next see one video of the sUAS performing the mission with one or more of the feature conditions active.
   b. Participants watch one of the videos.
   c. Participants are asked manipulation check questions.

d. Participants are provided the CTPA and HCTM questionnaires to indicate their trust in the sUAS to perform the task described in the scenario.
   e. After collecting a sufficient number of responses from participants, calculate the metrics in order to determine which features impact trust of the sUAS.
4. If the System Comparison test is being conducted:
   a. Participants are instructed that they will next see two videos of different sUAS performing the mission with one or more of the feature conditions active.
   b. Participants watch both of the videos.
   c. Participants are asked manipulation check questions.
   d. Participants are asked which system they prefer and to provide reasoning for their selection.
   e. Participants are provided the CTPA and HCTM questionnaires to indicate their trust in one of the sUAS (randomly selected) to perform the task described in the scenario.
   f. After collecting a sufficient number of responses from participants, calculate the metrics in order to determine which sUAS is trusted more than the other.

**Analysis**

1. Remove data points belonging to participants who failed to accurately respond to the manipulation check.
2. Towards removing outliers, calculate the first quartile (Q1, the middle value between the minimum and the median) and third quartile (Q3, the middle value between the median and the maximum) of the data for each item in the questionnaire, and the range (R) by subtracting Q1 from Q3 (i.e., R=Q3-Q1).
3. Outliers are then calculated as data points that fall outside of the range between Q1-1.5*R and Q3+1.5*R. It is recommended to not remove more than 10% of the data from the experiment (e.g., in a study with 30 participants, to not remove more than 3 outliers, else the participant pool may need to be increased)
4. After removing outliers, statistical significance tests are conducted on the resulting data to compare the two conditions (i.e., two features or two sUAS, depending on the type of evaluation). The Mann-Whitney statistical test is conducted to determine if there is a significant difference between the control data and the data from each condition (e.g., without a protective cage vs. with a protective cage), run pairwise among the results of consecutive questions or items of the questionnaires in the control data and the data from each condition.

## Example Data

- Environment characterization: Dark
- Performance Data: System Comparison

| Human-Computer Trust Measure (HCTM) | sUAS A Mean Score | sUAS B Mean Score | T-Test | Mann-Whitney Test |
|---|---|---|---|---|
| 1. I believe that there could be negative consequences when using the drone | 3.06 | 4.24 | S=-2.73 p=0.008 | U=259.5 p=0.005 |
| 2. I feel I must be cautious when using the drone | 3.86 | 5.13 | S=-2.183 p=0.003 | U=253.5 p=0.002 |
| 3. It is risky to interact with the drone | 3.48 | 4.10 | S=-1.20 p=0.23 | U=345 p=0.11 |
| 4. I believe that the drone will act in my best interest | 5.75 | 4.37 | S=4.13 p=0.0001 | U=140 p=0.0001 |
| 5. I believe that the drone will do its best to help me if I need help | 5.13 | 4.62 | S=1.22 p=0.22 | U=354 p=0.14 |
| 6. I believe that the drone is interested in understanding my needs and preferences | 4.62 | 3.41 | S=2.84 p=0.003 | U=249 p=0.002 |
| 7. I think that the drone is competent and effective in mapping mission | 5.81 | 4.48 | S=4.43 p=0.000004 | U=137 p=0.000006 |
| 8. I think that the drone performs its role as tool for mapping very well | 5.34 | 4.42 | S=2.24 p=0.02 | U=273 p=0.009 |
| 9. I believe that the drone has all the functionalities I would expect from an assistant for mapping | 5.82 | 3.93 | S=5.33 p=0.0000001 | U=137 p=0.0000001 |
| 10. If I use the drone, I think I would be able to depend on it completely | 5.59 | 3.65 | S=4.7 p=0.000001 | U=157 p=0.000001 |
| 11. I can always rely on the drone for performing mapping task | 5.37 | 3.75 | S=3.71 p=0.0004 | U=188 p=0.0001 |
| 12. I can trust the information presented to me by the drone | 5.73 | 4.72 | S=2.87 p=0.005 | U=228.5 p=0.004 |

| Checklist for Trust between People and Automation (CTPA) | sUAS A Mean Score | sUAS B Mean Score | T-Test | Mann-Whitney Test |
|---|---|---|---|---|
| 1. I am wary of the system. | 3.57 | 4.37 | S=-1.29 p=0.09 | U=337.5 p=0.05 |
| 2. The system's actions will have a harmful or injurious outcome. | 3.79 | 4.13 | S=-0.51 p=0.32 | U=405 p=0.21 |
| 3. I am confident in the system. | 5.85 | 4.0 | S=4.83 p=0.0001 | U=155 p=0.000007 |
| 4. The system provides security. | 5.87 | 4.03 | S=3.26 p=0.000002 | U=132 p=0.000001 |
| 5. The system has integrity. | 5.13 | 4.03 | S=2.81 p=0.006 | U=275.5 p=0.001 |
| 6. The system is dependable. | 5.84 | 4.55 | S=3.98 p=0.0002 | U=173 p=0.0001 |
| 7. The system is reliable. | 5.20 | 4.07 | S=2.21 p=0.03 | U=298.5 p=0.02 |
| 8. I can trust the system. | 5.82 | 4.51 | S=3.62 p=0.0006 | U=209 p=0.0006 |
| 9. I am familiar with the system. | 4.79 | 4.17 | S=1.54 p=0.12 | U=333 p=0.08 |

# Situation Awareness (SA)

These tests aim to evaluate whether sUAS provide the proper situation awareness (SA) for an operator during operations in subterranean and constrained indoor environments. The methods detailed in this category utilize quantitative SA assessment methods that can be utilized for evaluating SA of an individual (e.g., an operator after performing a mission) or for broader evaluation of the SA afforded by one sUAS compared to another. The latter allows for a comparison between multiple systems in order to characterize the type of SA they afford according to a series of mission-relevant characteristics.

## Interface-Afforded Attention Allocation

Affiliated publications: [Choi et al., 2022]

### Purpose

This test method is used to evaluate the attention allocation proportion to various situation elements (SEs), which are afforded by the sUAS interfaces and the defined mission objectives.

### Summary of Test Method

When controlling a sUAS, the operator relies on the OCU display to recognize the situation. According to the Attention Allocation Model [Xu et al., 2013], it is possible to quantify the attention allocation proportion ($f_i$) for situation elements (SEs) in the surrounding environment. This test method uses the SEEV model [Wickens et al., 2008; Xu et al., 2013] to quantify $f_i$. Then, by comparing multiple platforms using the SA Survey Comparison test method using videos of sUAS operations, $f_i$ can be confirmed via the correct rate of SAGAT, and OSA provided by platforms by platforms can be compared.

The typical OCU display screen of sUAS consists of images taken from the sUAS's camera with interface elements such as altitude, heading, etc., which can be determined as SEs. In order to calculate SA by applying the SEEV model, each SE can be divided into the four SEEV parameters (Salience, Effort, Expectancy, and Value) to quantify the attention allocation proportion ($f_i$). The quantitative value is the weighted sum or multiplication of four parameters. Salience is weighted according to the color, size, and type of the SE. Effort is weighted according to the movement of the operator's eye to see the SE. Expectancy refers to the weight of the event frequency or changeability of SE. Value refers to the importance according to the mission of the task.

4 Landolt Cs (Orange/Red/Green/Blue), 2 images (Radioactive/Oxygen), and 4 indicators (Altitude/Heading/Front Distance to Surface/System Battery) were identified as SEs from a series of operationally relevant scenarios (ORSs) were designed in order to mimic specific conditions in the environment, elements of interest, and mission tasks for sUAS operations in subterranean and constrained indoor spaces.

| **Situation Elements** |
|---|
| Landolt Cs (Orange/Red/Green/Blue)<br>Images (Radioactive/Oxygen)<br>Altitude<br>Heading<br>Front Distance<br>sUAS battery |

*Table 1. Situation elements (SEs) relevant to sUAS operations in subterranean and constrained indoor environments.*

SEs can be quantified by substituting numbers corresponding to information of SEs about Salience/Effort/Expectancy/Value into the SEEV model. All systems must be filmed from the same distance between the target and the systems with the target on the center of the screen of the controller (see Figure 1, right). Based on the result of the attention allocation proportion ($f_i$), we set out to compare $f_i$ among different platforms (Depending on the platform, there are cases where there is no SE(s) among indicator SEs) .

All systems must be filmed from the same distance between the target by placing a laser pointer and the systems with the target on the center of the screen of the controller (see Figure 1, right). Based on the result of the attention allocation proportion ($f_i$), we devised a cross-platform SA evaluation method. We set out to compare OSA with $f_i$ among different platforms.

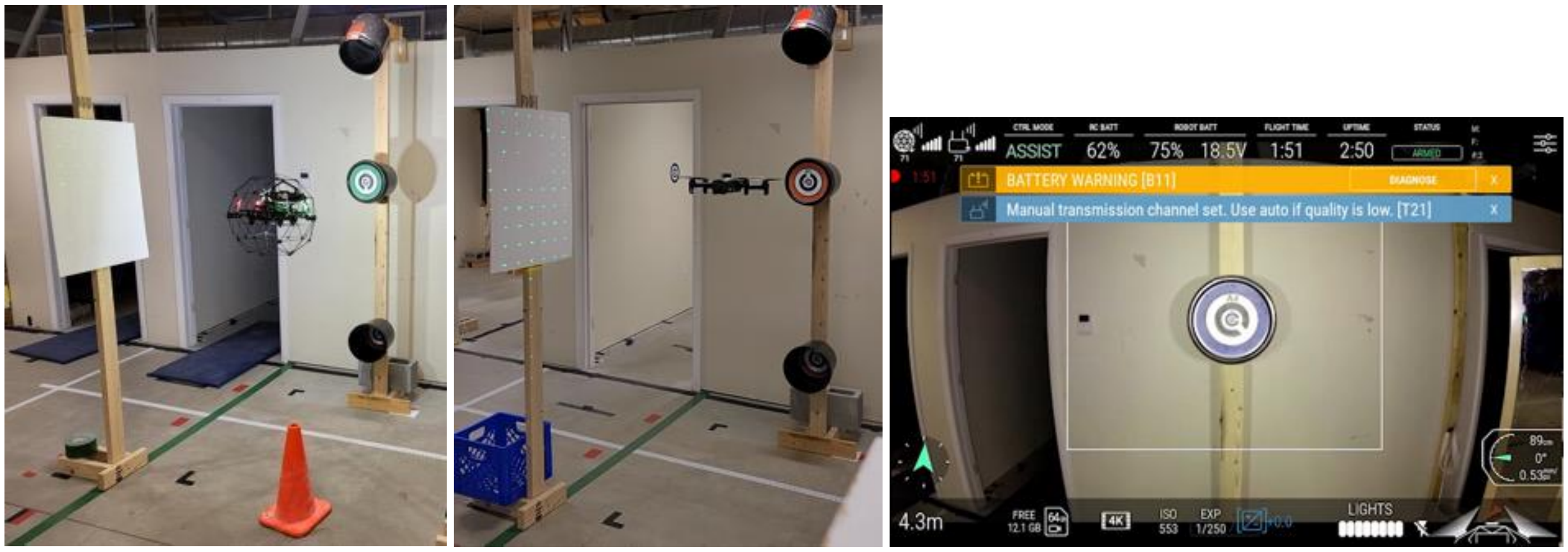

*Figure 1. Scenes of obtaining information of SEs (left: sUAS A, and center: sUAS B). Right: View of acuity target centered in the sUAS camera view.*

## Apparatus and Artifacts

Environmental and task elements may need to be fabricated in order to generate videos with the relevant SEs, which are obscure image items and acuity indicators called a Landolt C were used.

## Equipment

One or more video cameras are required in order to record sUAS activity and the OCU screen to generate the test videos shown to participants. Rather than using a camera to record the OCU display screen, it may contain functionality to record the screen directly. A laser pointer  is used to measure the position of sUAS. The participants must have access to a computer for watching the videos and responding to the questionnaires.

## Metrics

- Attention allocation proportion ($f_i$): The quantitative value is the multiplication of four parameters of SE (Salience($Sa_i$), Effort($E_i$), Expectancy($\beta$), and Value($V_i$ )).

$$f_i = A_i / \sum_{i=1}^{n} A_i, \quad A_i = \beta Sa_i {E_i}^{-1} V_i$$

Where, $A_i$ indicates the attention resources of SE.

## Procedure

1. Measuring the attention allocation proportion ($f_i$) for a cross-platform SA test method
   a. Prepare a target structure and place the rubber cone at a distance of 1 m between the sUAS and the target (e.g., 1m).

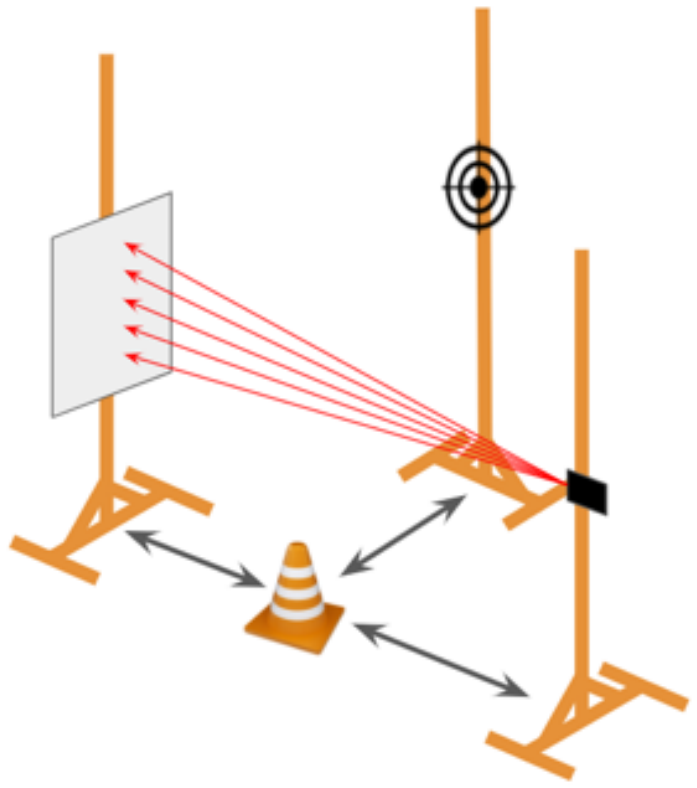

*Figure 2. Scenes of measuring fi*

b. Place a laser pointer structure for measuring the position of the sUAS at a distance of 1 m from the rubber cone in the vertical direction on both sides.
c. Fly a sUAS over the cone and adjust the height so that the target is centered on the screen of the controller and the distance (1m) between the sUAS and the target using the laser point structure.
d. Record a video for 5 secs. (Write the direction of the sUAS, height between the sUAS and ground, distance between the sUAS and the target in the experimental condition)
e. Using the attention allocation calculation model from the SEEV model, calculate the attention allocation proportion ($f_i$).

## Example Data

- Environment characterization: Lighted
- Performance data:

| SE | Name | sUAS A | sUAS B |
|---|---|---|---|
| 1 | Landolt C (Red) | 0.586 | 0.600 |
| 2 | Landolt C (Orange) | 0.644 | 0.642 |
| 3 | Landolt C (Green) | 0.706 | 0.661 |
| 4 | Landolt C (Blue) | 0.751 | 0.659 |
| 5 | Image (Oxygen) | 0.685 | 0.578 |
| 6 | Image (Radioactive) | 0.694 | 0.592 |
| 7 | Altitude | 0.407 | 0.371 |
| 8 | Heading | 0.373 | 0.332 |
| 9 | Front Distance | 0.395 | - |
| 10 | sUAS Battery | 0.505 | 0.349 |
| Total | | 5.746 | 4.784 |

*Table 2. Example of Comparison between attention allocation portions according to the sUAS platforms.*

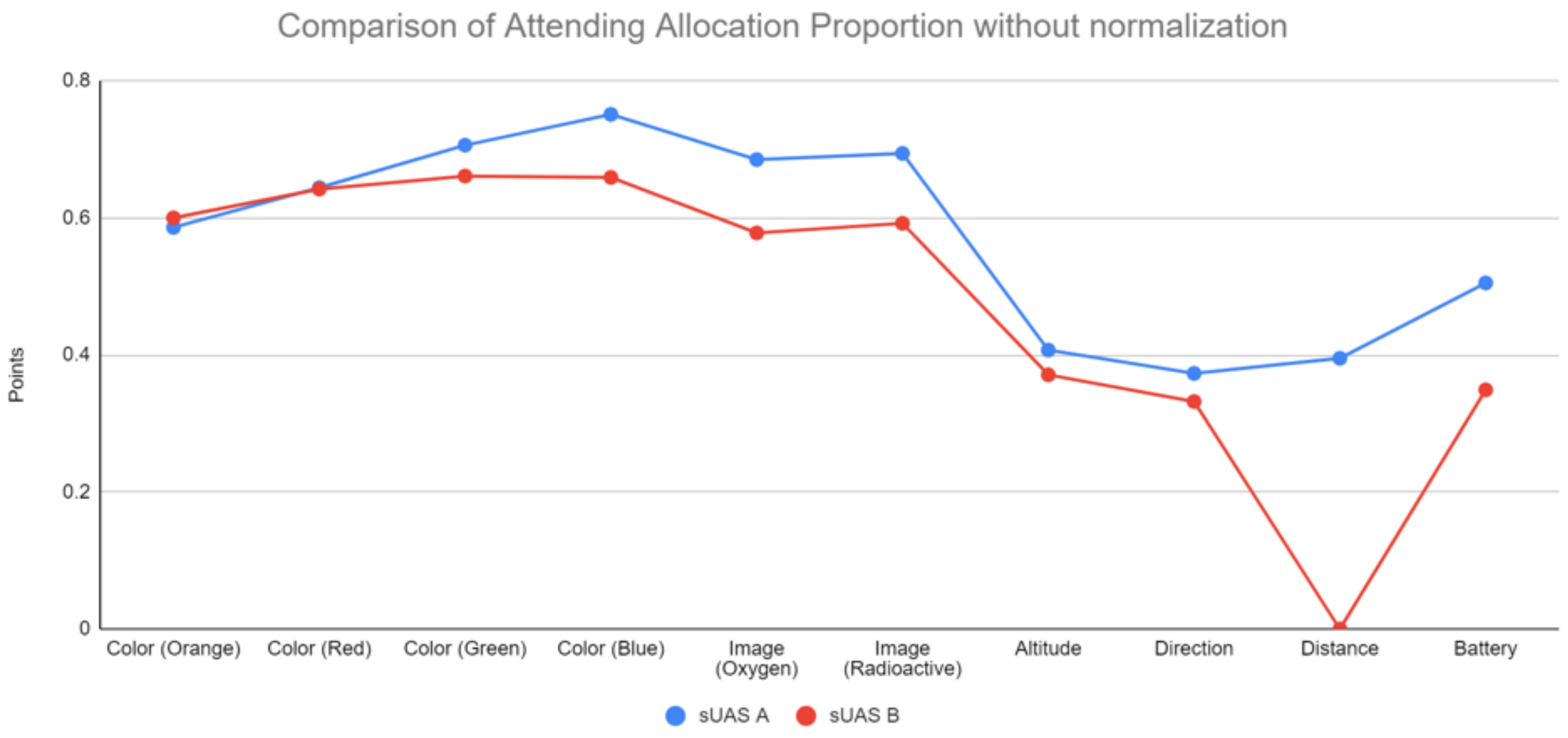

*Figure 3. Example of Comparison for each SE between attention allocation portions according to the sUAS platforms.*

## Situation Awareness (SA) Survey Comparison

Affiliated publications: [Choi et al., 2021]

### Purpose

This test method is used to evaluate operator SA afforded by two or more sUAS platforms in comparison to each other using two quantitative models of SA for evaluation.

### Summary of Test Method

As mentioned in the "Interface-Afforded Attention Allocation" SA test method, Landolt Cs, Images, and indicator SEs were identified as SEs from the ORS. By applying the SEEV model, the attention allocation proportion ($f_i$) and the probability of attending of SE (P(SE)) could be quantified.

According to the flight missions (Aviate/Navigate/Hazard detection), SEs are divided into required SEs and desire SEs based on the highest importance perceived by pilots as shown in Table 1. Therefore, the difference in abilities according to the missions can be emphasized by assigning weight required SEs and desired SEs. Through an experiment with a participant, each perception level (p(SE)), which is measured whether the participant is aware of the SE (i.e., undetected, detected, or comprehended), is obtained. The perception levels of SEs can be expressed as a metric of SA. Existing models and assessment methods, including the MIDAS (Man-machine Integration Design and Analysis)-based SA model and Attention Allocation Model [Xu et al., 2013], are adopted and refined in order to make them more suitable for sUAS operations in subterranean and constrained indoor environments. The metric goes through the theoretical models and OSA is obtained.

| Situation Element Type | Aviate | Navigate | Hazard |
|---|---|---|---|
| Required | Altitude<br>Heading<br>Front Distance<br>sUAS battery | Altitude<br>Heading<br>Front Distance<br>Images | Landolt Cs<br>Images |
| Desired | Images<br>Landolt Cs | Landolt Cs | Altitude<br>Heading<br>Front Distance |

*Table 1. Situation elements (SEs) relevant to sUAS operations in subterranean and constrained indoor environments.*

The operator's ability to identify situation elements (SEs) is evaluated according to the defined mission objectives, which include the use of visual acuity targets. The sUAS flies according to the defined flight path per each ORS, and if targets appear during flight, it stays in front of the target for 5 seconds. Videos of sUAS operations are then recorded performing these tasks with the various SEs in play. See Figures 1 and 2 for an overview of an example ORS.

Participants can be recruited to perform this test in-person or online (e.g., Amazon Mechanical Turk, Prolific), noting that online questionnaires can yield higher response rates. Participants are first briefed about the task and mission objective (i.e., what to look for, etc.) accompanied by a sample video. After this, participants watch an evaluation video of a sUAS flight with a specific focus to observe and comprehend the SEs of the operational environment. It should be noted that a human subjects research protocol will likely be required in order to recruit and run participants. All materials to run this test method including the videos and questionnaires should be reviewed by an institutional review board (IRB) and/or human research protections office (HRPO) as needed. Videos of sUAS operations are shown to participants who are then surveyed regarding certain characteristics of the operation in order to evaluate the SA afforded by the sUAS during the operation. The videos can be either of real or simulated sUAS activities.

Example videos of 8 parts of an ORS are linked below:

| | | |
|---|---|---|
| https://vimeo.com/687537430 | https://vimeo.com/587632877 | https://vimeo.com/687473527 |
| https://vimeo.com/687473657 | https://vimeo.com/687473736 | https://vimeo.com/687473834 |
| https://vimeo.com/687473879 | https://vimeo.com/687474050 | https://vimeo.com/687473943 |

At certain points while watching the video, the video will pause and questions from the SAGAT questionnaire [Salmon et al., 2009; Sulistyawati et al., 2011] will pop up for the participant to answer. The participants' answers are evaluated for accuracy to determine a corresponding score, and SA values are calculated using quantitative assessment methods. See Table 2 for examples of questions that can be used in this test method crossed with their related SEs and SA level.

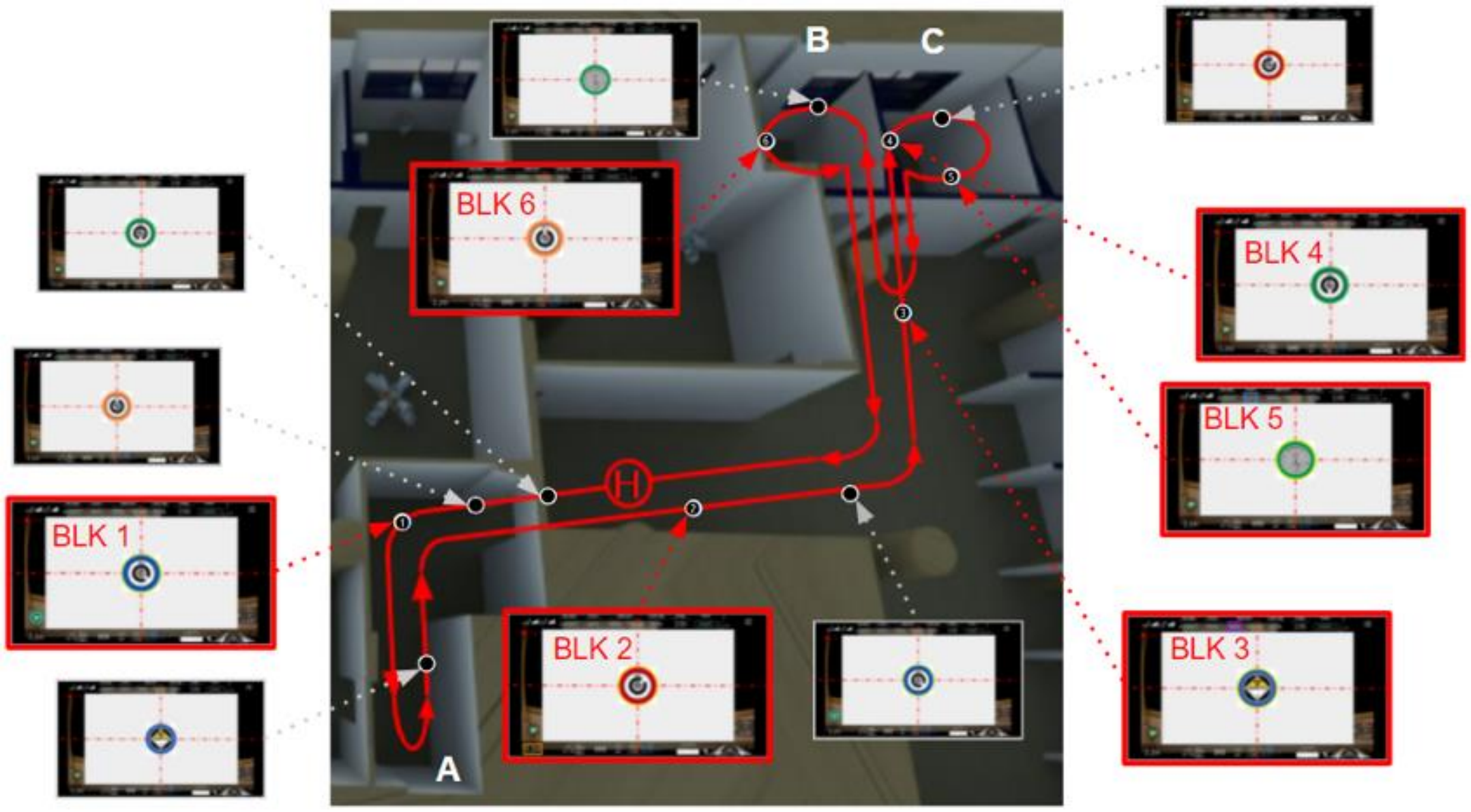

*Figure 1. Example of an operationally relevant scenario (ORS) used in this test method.*

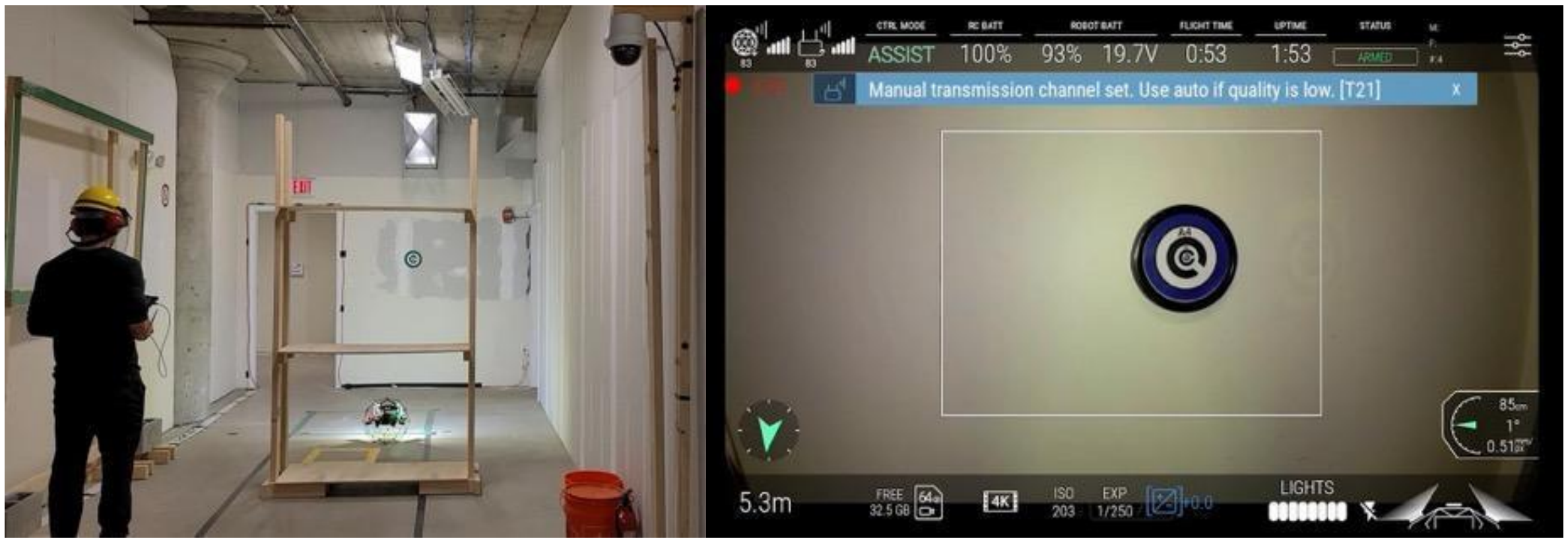

*Figure 2. Operator flying sUAS in the ORS (left) and a screen recorded in front of a target (right).*

| SE | SA Level | Question |
|---|---|---|
| Landolt Cs | Level 1 | Which color is the ring around the Landolt C when the video paused? |
| Images | Level 1 | What obscure image item was visible before the video stopped? |
| Altitude (Height) | Level 1 | What was the altitude of the drone while inspecting the last Landolt C indicator? |
| Heading | Level 1 | Which direction was the drone facing when the video paused? |
| Front Distance | Level 1 | What is the front distance to the surface before the video stopped? |
| sUAS battery | Level 1 | What percent of the drone's battery is left in the last second? |
| Understanding | Level 2 | How many doorway(s) did this room have? |
| Understanding | Level 2 | In which direction is the exit of the room you came from in reference to your current heading? |
| Understanding | Level 2 | How many Landolt Cs does this room have? |
| Understanding | Level 2 | How many rooms did you enter? |
| Understanding | Level 2 | How many doorways did you notice that you entered, or have entered in one of the previous videos? |

*Table 2. Examples SAGAT questions for use in this test method and their associated situation element (SE) and SA level.*

Cross-platform Operator SA assessment is introduced for this SA test method. A participant takes two experiments of two platforms, in which were randomly assigned the order of two platforms. To prevent the participant from becoming familiar with the experiments, by changing the order of the rooms (A,B, and C type) as shown in Figure 3 and artifacts (a, and b type) as shown in Figure 4, a total of 6 ORSs can be constructed as shown in Table 3.

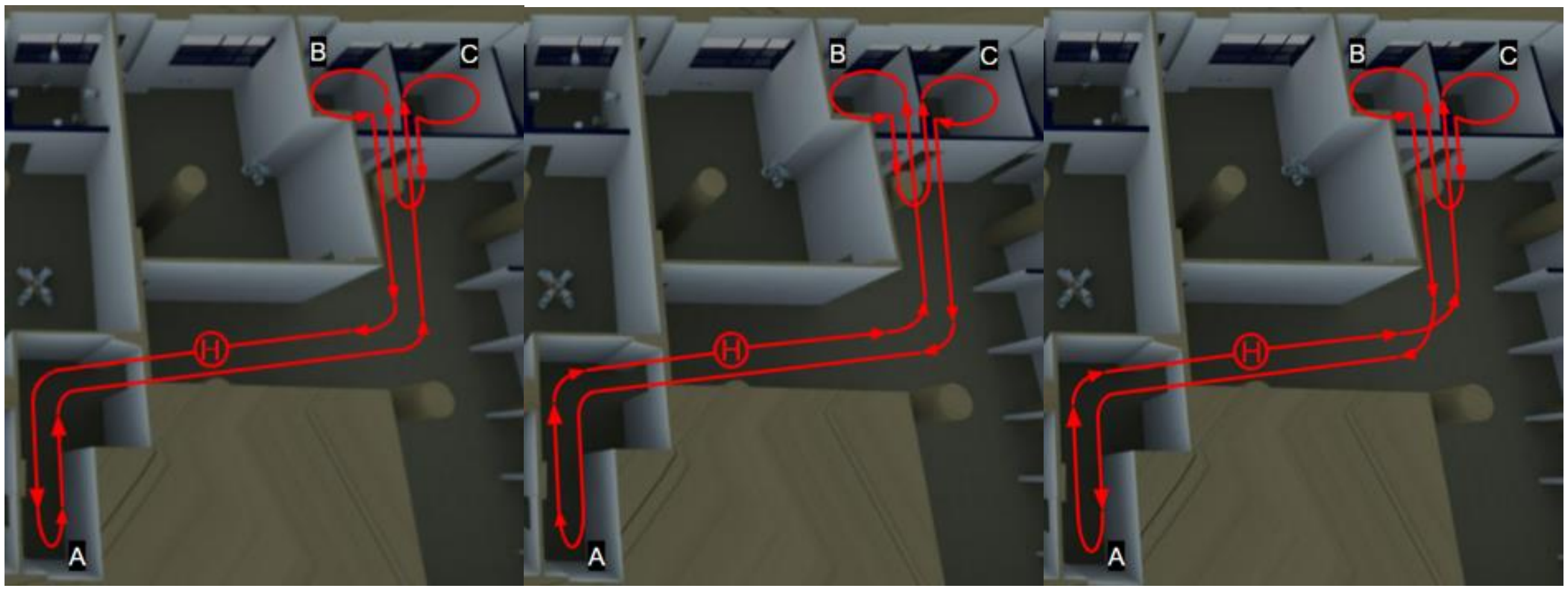

*(a) Type A (A→C→B)* *(b) Type B (B→C→A)* *(c) Type C (C→B→A)*

*Figure 2. Examples of an operationally relevant scenario (ORS) with the type of the order of rooms*

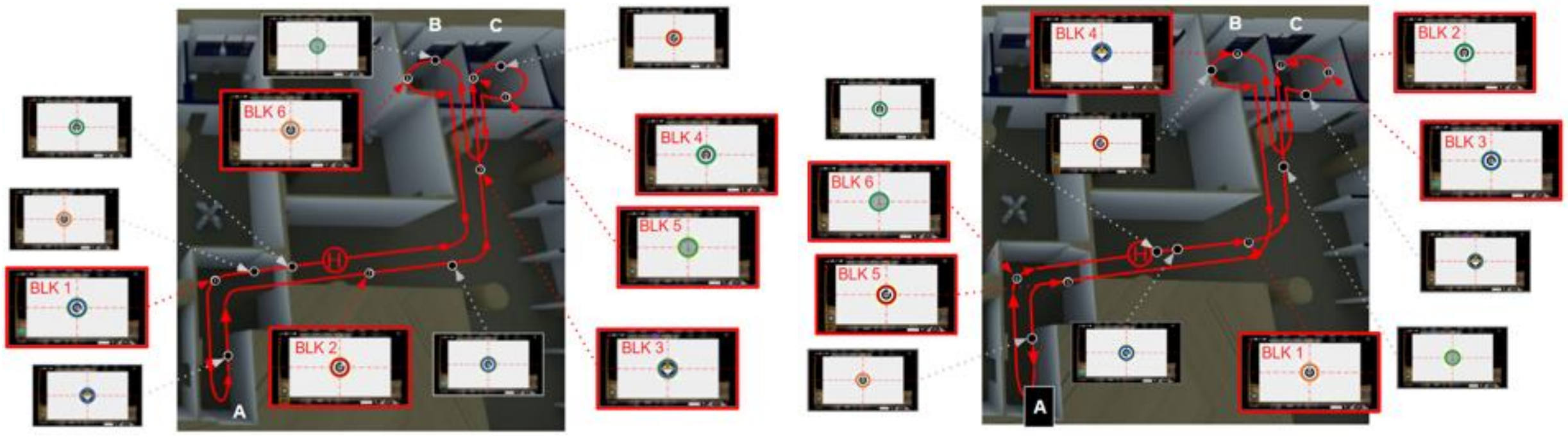

*Figure 3. Examples of an operationally relevant scenario (ORS) with the type of artifacts. (left: a type, and right: b type)*

To compare a platform with others, there are cases where there is no SE(s) among indicator SEs. Therefore, virtual proportion (VP) was developed to calculate the attention allocation proportion (or the probability of attending) of SE, is not on the screen of the controller.  VP (See Table 4) can be calculated through linear interpolation by comparing the correct rate of other SE.

| **ORS** | **Type of the order of rooms** | **Type of artifacts** | **Note** |
|---|---|---|---|
| ORS 1 | A | a | |
| ORS 2 | B | a | Add 1 door in room B |
| ORS 3 | C | a | Add 1 door in room C |
| ORS 4 | A | b | |
| ORS 5 | B | b | Add 1 door in room B |
| ORS 6 | C | b | Add 1 door in room C |

*Table 3. Comparison of 6 ORSs*

Examples of calculated attention allocation proportion ($f_i$) for sUAS in a test course with visual acuity targets can be seen in Table 1.

| SE | Name | sUAS A | sUAS B |
|---|---|---|---|
| 1 | Landolt C (Red) | 0.116 | 0.128 |
| 2 | Landolt C (Orange) | 0.125 | 0.138 |
| 3 | Landolt C (Green) | 0.135 | 0.142 |
| 4 | Landolt C (Blue) | 0.143 | 0.127 |
| 5 | Image (Oxygen) | 0.112 | 0.106 |
| 6 | Image (Radioactive) | 0.114 | 0.124 |
| 7 | Altitude | 0.054 | 0.054 |
| 8 | Heading | 0.051 | 0.052 |
| 9 | Front Distance | 0.051 | 0.053* |
| 10 | sUAS Battery | 0.099 | 0.076 |
| Total | | 1.000 | 1.000 |

*Table 4. Example of Comparison between attention allocation portions according to the sUAS platforms. * = Virtual Proportion Value*

## Apparatus and Artifacts

Environmental and task elements may need to be fabricated in order to generate videos with the relevant SEs, such as visual acuity targets consisting of Landolt Cs and hazmat symbols.

## Equipment

One or more video cameras are required in order to record sUAS activity and the OCU screen to generate the test videos shown to participants. Rather than using a camera to record the OCU display screen, it may contain functionality to record the screen directly. The participants must have access to a computer for watching the videos and responding to the questionnaires.

## Metrics

- The Probability of attending of SE (P(SE)): The quantitative value from the SEEV model is the weighted sum of four parameters (<u>S</u>alience, <u>Ef</u>fort, <u>Ex</u>pectancy, and <u>V</u>alue).

  $$P(SE) = sS - efEF + exEX + vV$$

  Where, coefficients in the uppercase describe the properties of SEs, while those in the lower case describe the weight assigned to those properties.

- <u>Awareness of SE</u>: The output of the quantitative SA assessment method provides a measure of SA per the SE being evaluated. First, the probability of attending (P(SE)), which is the sum of SE information of SE, and the Attention Allocation Proportion ($f_i$), which is the product of information of SE, are quantified by the visual attention model. Then, the perception level (p(SE)) of the corresponding SE is obtained through the perception model, and the SA can be calculated through the applied SA models. Refer to [Hooey et al., 2011; Xu et al., 2013] for more thorough information regarding how to perform these calculations.

## Procedure

After recording and editing the videos needed to run the test method, the survey method that will provide the participant with instructions, show the videos, and provide them with questionnaires to fill out must be designed. This could be done manually using pen and paper with a computer screen to display the videos, or through an online survey tool such as Qualtrics.

1. Participants are first asked to review and sign a form providing their consent to participate (following IRB and HRPO requirements).
2. Participants are provided instructions regarding what they will be asked to do during the study:
    a. Watch videos of sUAS operations from the perspective of the operator through the OCU,
    b. Observe the environment as seen by the sUAS, and
    c. Answer questions regarding the operations occurring, which will be prompted when the video pauses.
3. By substituting the corresponding score from the participants' answers into the quantitative SA assessment method, SA measures per SE can be obtained.

## Example Data

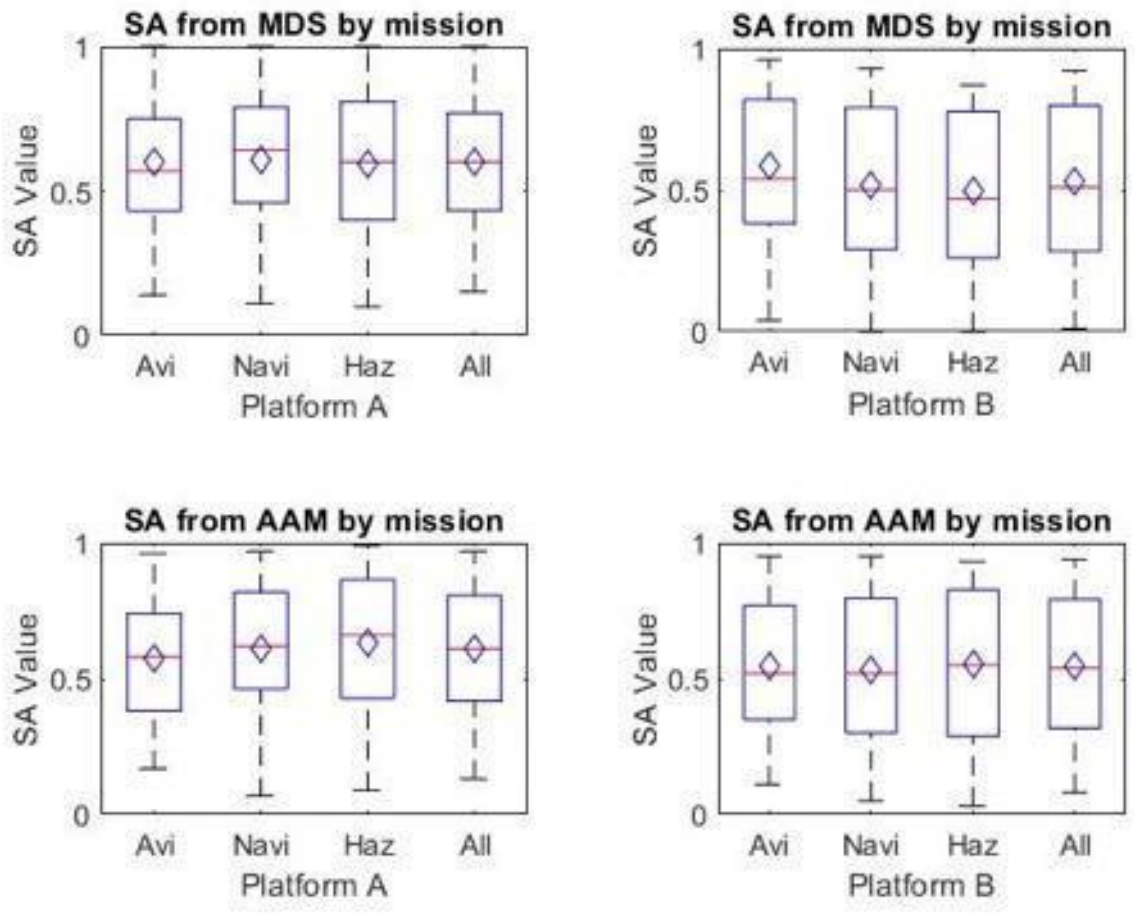

| Mission | sUAS A | | sUAS B | |
|---|---|---|---|---|
| | MDS | AAM | MDS | AAM |
| | $\mu, \sigma$ | $\mu, \sigma$ | $\mu, \sigma$ | $\mu, \sigma$ |
| Aviate | 0.22, 0.60 | 0.22, 0.59 | 0.27, 0.59 | 0.26, 0.55 |
| Navigate | 0.23, 0.61 | 0.24, 0.61 | 0.29, 0.51 | 0.28, 0.54 |
| Hazard | 0.26, 0.59 | 0.25, 0.63 | 0.28, 0.49 | 0.28, 0.55 |
| Overall | 0.23, 0.60 | 0.23, 0.61 | 0.27, 0.53 | 0.27, 0.54 |

The OSA calculated for participants of the experiment are tabulated in terms of mean and standard deviation, denoted by $\mu$ and $\sigma$, respectively. The results are grouped based on the mission and OSA values are calculated using both MIDAS-based SA model and Attention Allocation Model.